\documentclass[10pt,twocolumn,letterpaper]{article}

\usepackage{cvpr}              % To produce the CAMERA-READY version
\usepackage[dvipsnames]{xcolor}

\usepackage[T1]{fontenc}

\definecolor{cvprblue}{rgb}{0.21,0.49,0.74}
\usepackage[pagebackref,breaklinks,colorlinks,citecolor=cvprblue,linkcolor=cvprblue,menucolor=cvprblue,urlcolor=cvprblue]{hyperref}
\usepackage{wrapfig}
\usepackage{adjustbox}
\usepackage{multicol}

\usepackage{titletoc}
\usepackage{chngcntr}
\usepackage{algorithm}
\usepackage{algorithm}
\usepackage{algpseudocode}

\usepackage[accsupp]{axessibility} % Improves PDF readability for those with disabilities.
\usepackage{comment}

\usepackage[accsupp]{axessibility} % Improves PDF readability for those with visual impairments.

\def\paperID{7623} % *** Enter the Paper ID here
\def\confName{CVPR}
\def\confYear{2024}

\newcommand{\ggF}{\color{SeaGreen}}
\newcommand{\yyF}{\color{GreenYellow}}
\newcommand{\rrF}{\color{WildStrawberry}}

\title{Systematic comparison of semi-supervised and self-supervised learning \\ for medical image classification}

\author{%
\begin{tabular}{cccc} 
Zhe Huang\thanks{Authors Zhe Huang and Ruijie Jiang contributed equally to this work.} & 
Ruijie Jiang\footnotemark[1] & 
Shuchin Aeron & 
Michael C. Hughes
\end{tabular} \\
\begin{tabular}{cc} 
Tufts University, School of Engineering
\end{tabular} \\
{\tt\small \{Zhe.Huang, Ruijie.Jiang, Shuchin.Aeron, Michael.Hughes\}@tufts.edu}
}

\begin{document}
\maketitle

\begin{abstract}

In typical medical image classification problems, labeled data is scarce while unlabeled data is more available. Semi-supervised learning and self-supervised learning are two different research directions that can improve accuracy by learning from extra unlabeled data. 
Recent methods from both directions have reported significant gains on traditional benchmarks. Yet past benchmarks do not focus on medical tasks and rarely compare self- and semi- methods together on an equal footing.
Furthermore, past benchmarks often handle hyperparameter tuning suboptimally. First, they may not tune hyperparameters at all, leading to underfitting. Second, when tuning does occur, it often unrealistically uses a labeled validation set that is much larger than the training set.
Therefore currently published rankings might not always corroborate with their practical utility
%Both cases raise questions about the reliability of previously published rankings of methods. 
This study contributes a systematic evaluation of self- and semi- methods with a unified experimental protocol intended to guide a practitioner with scarce overall labeled data and a limited compute budget.
We answer two key questions:
\textbf{Can hyperparameter tuning be effective with realistic-sized validation sets?}
If so, when all methods are tuned well, \textbf{which self- or semi-supervised methods achieve the best accuracy}?
%over a supervised training on the labeled-set only?
%\textbf{is hyperparameter tuning still worthwhile with realistic-sized validation sets?} 
Our study compares 13 representative semi- and self-supervised methods to strong labeled-set-only baselines on 4 medical datasets.
%with more than 20000 total GPU hours devoted to the study. 
From 20000+ GPU hours of computation, we provide valuable best practices to resource-constrained practitioners: hyperparameter tuning is effective, and the semi-supervised method known as MixMatch delivers the most reliable gains across 4 datasets.

\end{abstract}

\makeatletter\def\Hy@Warning#1{}\makeatother\let\thefootnote\relax\footnotetext{Code: \href{https://github.com/tufts-ml/SSL-vs-SSL-benchmark}{github.com/tufts-ml/SSL-vs-SSL-benchmark} [MIT license].}

% \let\thefootnote\relax\footnotetext{Open-source code [MIT license]: \url{https://anonymous.4open.science/r/SSL-vs-SSL-benchmark-48B0/README.md}}
% \url{https://github.com/tufts-ml/SSL-vs-SSL-benchmark}

\section{INTRODUCTION}
Deep neural networks can deliver exceptional performance on classification tasks when trained with vast labeled datasets. 
However, in medical imaging applications assembling a large dataset with appropriate label can be prohibitively costly due to manual effort required by a human expert. 
In contrast, images alone, without labels, are often readily available in health records databases. In recent years, significant research has focused on developing methods that leverage both the large unlabeled set and a small labeled set to enhance image classifier training, aiming to surpass models trained solely on labeled data.

%HZ: GOAL OF THE PROJECT 1: Direct comparison of Semi - Self
Two ways of leveraging unlabeled data are particularly popular: semi-supervised and self-supervised learning, both often abbreviated as SSL.
Recent efforts in \emph{semi-supervised} learning~\citep{zhu2005semi,van2020survey} usually train deep classifiers \emph{jointly}~\citep{sohn2020fixmatch,berthelot2019mixmatch} using an objective with two loss terms, one favoring labeled-set accuracy and the other favoring label-consistency or label-smoothness.
Alternatively, \emph{self-supervised} methods~\citep{qi2020small} take a two-stage approach, first training deep representations on the unlabeled set, then fine-tuning a classifier on the labeled set. Exemplars are numerous~\citep{chen2020simple, he2020momentum, chen2020improved, caron2020unsupervised, chen2021exploring}. 
Despite the remarkable progress reported in each direction, these two paradigms have been largely developed independently~\citep{chen2022semi}. 
A direct comparison of the two paradigms has been notably absent.
A practitioner building a medical image classifier from limited labeled data may ask, ``\textbf{Which recent semi- or self-supervised methods are likely to be most effective?}''

%HZ: GOAL OF THE PROJECT 2: Directly correspond ot Oliver et al's claim about 'Realistic' validation set size. 
% Further, seminal work by \citet{oliver2018realistic} expressed concern about the ability to properly tune hyperparameters for semi-supervised learning in practice when available labeled data is limited, saying ``Extensive hyperparameter tuning may be somewhat futile due to an excessively small collection
% of held-out data to measure performance on''. 

Performance can be quite sensitive to hyperparameters for both semi-SL~\citep{su2021realistic,sohn2020fixmatch} and self-SL~\citep{wagner2022importance}.
We argue that any careful comparison must consider \emph{tuning} hyperparameters of all methods in a fair fashion.
However, recent work has not handled this well.
First, as reviewed in Table~\ref{tab:related_work_comparison}, recent benchmarks for both semi- and self- paradigms often omit \emph{any} tuning of hyperparameters (see the Fungi semi-SL experiments of~\citet{su2021realistic} or \citet{ericsson2021well}'s self-SL benchmark).
This practice of using off-the-shelf defaults likely leads to under-performing given a new task, and may impact some methods more than others.
An even bigger issue with prevailing semi-SL practice was originally raised by 
\citet{oliver2018realistic}: many published papers perform hyperparameter tuning on a labeled validation set that is larger (sometimes much larger) than the limited labeled train set. 
Such experimental settings are \emph{unrealistic}. SSL is intended for practitioners without abundant available labeled data. Practitioners that need to make the most of 1000 available labeled images will not elect to put more images in validation than training, and thus are not helped by benchmarks that do.
Unfortunately, five years later after \citep{oliver2018realistic} we find this issue is still widespread (see Tab.~\ref{tab:related_work_comparison}). For example, \citet{wang2022usb}'s semi-SL results on TissueMNIST tune on a validation set over 50x larger than the labeled train set. 

%They concern that in real applications with realistic-sized small validation set, hyperparameter tuning might be futile. 
% \citet{oliver2018realistic} concern that hyperparameter tuning in realistic semi-supervised setting might be ``futile'' due to the fact that in real application of semi-supervised learning, the validation set would be much smaller than what has been used in general practice in academic papers.
% concern about the ability to properly tune hyperparameters for semi-supervised learning in practice when available labeled data is limited, saying ``Extensive hyperparameter tuning may be somewhat futile due to an excessively small collection
% of held-out data to measure performance on''. 

\citet{oliver2018realistic} further cast some doubt on whether effective tuning is possible with small labeled sets, saying ``Extensive hyperparameter tuning may be somewhat futile due to an excessively small collection
of held-out data to measure performance on''. 
There thus remains a pressing question for resource-constrained practitioners: ``\textbf{Given limited available labeled data and limited compute, is hyperparameter tuning worthwhile?}''

% This study develops an open-source benchmark intended to help practitioners leverage unlabeled data to tackle new medical image classification challenges where only modest-sized labeled datasets exist.
This study makes progress toward answering these questions by delivering a comprehensive comparison of semi- and self-supervised learning algorithms under a resource-constrained scenario that matches how SSL might be used on real-world medical tasks.
Our goal is to enable practitioners to tackle new image classification challenges where only modest-sized labeled datasets exist.
We target settings with roughly 2-10 class labels of interest, where each class has 30-1000 available labeled images for all model development (including training and validation).
We select representative tasks across 4 datasets that span low-resolution (28x28) to moderate-resolution (112x112 and 384x384).
On each task, we run careful experiments with representative recent methods from both semi- and self-supervised learning to clarify what gains are possible with unlabeled data and how to achieve them.
%To train each method, we assume access to one high-end GPU (nVidia A100) for a fixed time budget.

We emphasize realism throughout, in 4 distinct ways: (1) using validation set sizes that are \emph{never bigger than the train set}, avoiding the unrealistically-large validation sets common in past work (Tab.~\ref{tab:related_work_comparison}); (2) performing hyperparameter search using the \emph{same protocol}, \emph{same compute budget} (fixed number of hours)*\footnote{*\citet{oliver2018realistic} give each method 1000 trials of a cloud hyperparameter tuning service. However, training speed can differ substantially across SSL methods. We argue a fixed wallclock time budget is more fair.} and \emph{same hardware} (one NVIDIA A100 GPU) for all algorithms for fair comparison; (3) respecting natural \emph{class imbalance}; (4) profiling \emph{performance over time}, to inform labs with smaller runtime budgets.

%, avoiding the unfortunately common practice in other published benchmarks of validation sets that are larger, often far larger, than the train set (see Table 1). 

In summary, the contributions of this study are:

\begin{enumerate}[leftmargin=*,nosep]
\item We provide a \textbf{systematic comparison of semi- and self-supervised methods} on equal footing to connect two research directions that have been heretofore separate.
\item We adopt a \textbf{realistic experimental protocol} designed to consider the same constraints on available labels and runtime that SSL practitioners face, avoiding the unrealistic aspects of past benchmarks, especially the use of far too large validation sets. 

\item We show that \textbf{hyperparameter tuning is viable with a \emph{realistic-sized} validation set}, and in many cases \textbf{\emph{necessary}} to do well at a new classification task with new data.
\end{enumerate}
Ultimately, we hope this study guides practitioners with limited data toward successful deployment of semi- and self-supervised methods on real problems.

\label{intro}

\newcommand{\WWW}{\hspace{2mm}}
\setlength{\tabcolsep}{.13cm}
\begin{table*}[!t]
\begin{tabular}{l l l r r l}
Benchmark & Methods & Unrealistic Experiments & Labeled train size & Labeled val.~size & Acc vs.~Time?	
\\ \hline
Realistic eval. SSL {\footnotesize [a]}
        & Semi
        & CIFAR-10 {\small (Tab. 1-2, Fig. 2)}
        % & CIFAR-10
	& 4000
        & 5000
        & \WWW  no
	
\\
        &
        & SVHN {\small (Tab. 1-2, Fig. 3-4)}
        % & SVHN
	& 1000
        & 7325
        &
	
\\ \hline
Fine-grained SSL {\footnotesize [b]}
        & Semi \& 1 Self$^*$
	& Semi-Aves
	& 5959
        & 8000
        & \WWW  no
\\

        &  
	& Semi-Fungi
	& 4141
        & None
        &
	
% \\
% TorchSSL {\tiny \citep{zhang2021flexmatch}}
% 	& NO
% 	& Not used\todo{DISCUSS}
% 	& NO
% 	& Semi
\\ \hline
USB {\footnotesize [c]}
        & Semi
	& TissueMNIST
	& 80/400
        & 23640$^{\dagger}$
        & \WWW  no
        
\\

        & 
	& Semi-Aves
	& 5959
        & None
        &
	
\\ \hline
Self benchmark  {\footnotesize [d]}
        & Self 
	& ImageNet
	& 1.28 mil.
        & None
        & \WWW  no
\\ \hline
SSL-vs-SSL~\emph{(ours)}
        & Semi \& Self 
	& no 
    & 400-1660
    & %235-600
    \small{no bigger than train}
    & \WWW  yes
% \\
% \textbf{SSL-vs-SSL} (ours) 
% 	& realistic
% 	& YES
% 	& 6 Semi \& 5 Self

\end{tabular}

\caption{\textbf{Comparison of related benchmarks of semi-supervised and self-supervised learning.} 
Past works either do no hyperparameter tuning at all (val.~size = None), or use an unrealistically large validation set (defined as \emph{larger than the labeled train set}) in some or all experiments.
Further, each work almost exclusively looks at methods from one paradigm, either semi- or self- (*: [b] includes one self-supervised method, MOCO). 
In contrast, in this paper we benchmark 6 semi- and 7 self-supervised algorithms 
with hyperparameter tuning on realistic validation sets.
\emph{Acc vs. Time?} indicates whether the work analyzes performance over training time.
Number marked $\dagger$ confirmed via  \href{https://github.com/microsoft/Semi-supervised-learning/issues/82}{GitHub comment by USB authors}.
Citations: a: \citet{oliver2018realistic}, b: \citet{su2021realistic} 
     c: \citet{wang2022usb} d: \citet{ericsson2021well}.
    }
\vspace{-0.1cm}
\label{tab:related_work_comparison}
\end{table*}

\section{BACKGROUND AND METHODS}
% In superivsed learning, we are given a training dataset of input-target pairs $(x,y) \in D^L$. 
%In this benchmark, we compared 5 semi-supervised, 5 self-supervised learning algorithms and 3 strong supervised learning baselines. Here, we briefly describe the algorithms we used. More details can be found in App~\ref{App_Algorithms_Details}.

\paragraph{Unified Problem Formulation.} Following the recent survey by~\citet{chen2022semi}, we adopt a unified perspective for supervised, semi-supervised and self-supervised image classification with limited available labeled data.
For model development (including training and hyperparameter selection), we assume there are two available datasets.
First, a small labeled dataset $\mathcal{L}$ of feature-label pairs $(x, y)$, where each image is represented by a $D$-dimensional feature vector $x \in \mathbb{R}^D$ and its corresponding class label takes one of $C$ possible values: $y \in \{1,2, \ldots C\}$.
Second, an unlabeled set $\mathcal{U}$ containing only feature vectors.
Typically, we assume the unlabeled set is much larger: $|\mathcal{U}| \gg |\mathcal{L}|$.

Given labeled set $\mathcal{L}$ and unlabeled set $\mathcal{U}$, we wish to train a neural network that can map each input $x$ to a probability vector in the $C$-dimensional simplex  $\Delta^C$ representing a distribution over $C$ class labels.
Let $f_v(\cdot) : \mathbb{R}^D \rightarrow \mathbb{R}^F$ denote a backbone neural network with parameters $v$ producing an $F$-dimensional embedding given any input image. Let $g_w(\cdot) : \mathbb{R}^F \rightarrow \Delta^C$ denote a final linear-softmax classification layer with parameters $w$.
The following unified objective can capture all three learning paradigms:
%\begin{equation}
\begin{align}
\label{eq:objective_unified}
    v^*, w^* \gets \arg\min_{v, w} ~~
    &
    \textstyle \sum_{x,y \in \mathcal{L}}  \lambda^{L} \ell^{L}( y,  g_w( f_v(x)) ) \\ \notag
    & + 
    \textstyle \sum_{x \in \mathcal{U}} \lambda^{U} \ell^{U}( x, f_v, g_w).
\end{align}
%\end{equation}
Here, $\ell^L$ represents a labeled-set loss (e.g. multi-class cross-entropy), and $\ell^U$ represents a unlabeled-set loss. $\lambda^L, \lambda^U \geq 0$ are weights for the corresponding loss terms. The design of $\ell^U$ is usually what differs substantially across methods. For instance, setting $\ell^U$ to be cross-entropy computed with pseudo-labels generated from classifier $g$ recovers PseudoLabel \citep{lee2013pseudo}; a temperature-scaled instance-similarity contrastive loss recovers SimCLR~\citep{chen2020simple}.

% \begin{align}
%     v^*, w^* \gets \arg\min_{v, w} ~~
%     \textstyle \sum_{x,y \in \mathcal{L}} \lambda^{L} \ell^{L}( y,  g_w( f_v(x)) )
%     \\+ 
%     \textstyle \sum_{x \in \mathcal{U}} \lambda^{U} \ell^{U}( x, f_v, g_w)
% \label{eq:objective_unified}
% \end{align}

All three learning paradigms that we study optimize Eq.~\eqref{eq:objective_unified}, yet differ in the number of phases and in how to set the scalar weights $\lambda^L, \lambda^U$ on each loss term.
Supervised learning ignores the unlabeled term throughout training ($\lambda^U = 0$), thus learning parameters using only the labeled set.
Semi-supervised methods include both terms in one end-to-end training, keeping both $\lambda^U > 0$ and $\lambda^L > 0$.

Self-supervised learning has two phases.
In phase 1 (``pretraining''), the labeled term is omitted ($\lambda^L = 0, \lambda^U = 1$) and the focus of learning is an effective representation layer $f_v$ (classifier $g_w$ is not included in this phase).
In phase 2 (``fine-tuning''), we focus on the labeled term and omit the unlabeled term ($\lambda^L = 1, \lambda^U = 0$). We fix the representation parameter $v$ and fine-tune the classifier $w$.
%\todo{TODO are both v and w fine-tuned, or just classifier w?}

%We assume a labeled set $D^L = \{x_i, y_i\}_{i=1}^{N_{L}}$ with size $N_{L}$ and each sample $x_i$ belongs to one of the $K$ class labels $y_i \in \{1, 2, ..., K\}$ and a large unlabeled set 
%$D^U = \{x_j\}_{j=1}^{N_{U}}$ with only feature samples $x_j$ and no label information for each sample, our goal is to train a neural network that given a new sample $x_{*}$ assigns the correct label $y_{*}$. The loss function for training this neural network can be expressed as 
%\begin{align} 
%L = \lambda_l L_{sup} + \lambda_u L_{unsup}
%\end{align}
%Supervised learning uses only $D^L$ throughout. For the semi-supervised learning algorithms we focus on in this study, we have $\lambda_{sup} \neq 0$ and $\lambda_{unsup} \neq 0$. Whereas for self-supervised learning, $\lambda_{sup}=0$ during pretraining and $\lambda_{unsup}=0$ during finetuning. 

We now identify the 16 methods we will evaluate:

\textbf{Supervised methods.}
The goal of leveraging unlabeled data $\mathcal{U}$ is to obtain better performance than what we could obtain using only the labeled set $\mathcal{L}$.
Therefore, we naturally compare to 3 high-quality supervised baselines that use only the labeled set.
First, ``Sup'' denotes a classifier trained with supervised loss $\ell^L$ set to multi-class cross-entropy. 
Second, ``MixUp'' trains with cross entropy with the addition of mixup data augmentation~\citep{zhang2017mixup}.
Finally, ``SupCon'' pursues a supervised contrastive learning loss for $\ell^L$~\citep{khosla2020supervised}.
%The latter two represent modern strong supervised baselines \emph{without} any unlabeled data.

\textbf{Semi-supervised methods.}
%Semi-supervised learning uses both a modest-sized labeled set and a (much larger) unlabeled set to jointly train a classifier. 
%The key idea is that the unlabeled data provides important clues to inform underlying data distribution. 
%We chose to concentrate on deep neural net methods that utilize both labeled and unlabeled data \emph{simultaneously} during  training, as in Eq.~\eqref{eq:objective_unified}. %Specifically, we focus on techniques that make only minor adjustments to the standard objectives of discriminative neural networks, primarily by appending an additional loss term that draws upon the unlabeled data. 
We compare 6 semi-supervised methods that train deep classifiers
on both labeled and unlabeled data \emph{simultaneously} as in Eq.~\eqref{eq:objective_unified}. 
To represent the state-of-the-art of semi-supervised image classification, we select Pseudo Label (``PseudoL'')~\citep{lee2013pseudo}, Mean Teacher (``MeanTch'')~\citep{tarvainen2017mean}, MixMatch~\citep{berthelot2019mixmatch}, FixMatch~\citep{sohn2020fixmatch}, FlexMatch~\citep{zhang2021flexmatch} and CoMatch~\citep{li2021comatch}. These choices cover a reasonably wide spectrum of unlabeled loss design strategy, year of publication,  and computation cost.
CoMatch represents a recent trend of combining semi- and self-supervision.
See App \ref{app:semi_supervised_method_details} for a wider literature review.

\textbf{Self-supervised methods.}
We compare 7 self-supervised algorithms: SimCLR~\citep{chen2020simple}, MOCO (v2)~\citep{he2020momentum,chen2020improved}, SwAV~\citep{caron2020unsupervised}, BYOL~\citep{grill2020bootstrap}, SimSiam~\citep{chen2021exploring}, DINO~\citep{caron2021emerging} and Barlow Twins (``BarlowTw'') ~\citep{zbontar2021barlow}.
These algorithms epitomize the field of self-supervised learning as of this writing. SimCLR, MOCO (v2), and SwAV are based on contrastive learning, which learns effective representations when provided with similar and dissimilar samples. BYOL was designed to circumvent the need for dissimilar samples. SimSiam is a simple yet effective Siamese representation learning method. DINO uses a teacher-student network architecture for representation distillation. Barlow Twins preserves information and reduces redundancy by favoring a close-to-identity cross-correlation matrix between augmented pairs of data.
See App \ref{app:self_supervised_method_details} for further review.

%Different from supervised and semi-supervised learning, self-supervised learning aims to learn a representation function, denoted as $f$, that is independent of the subsequent task. In this approach, no natural classifier is learned during the training process. Given the limited number of labeled data, we utilize the learned representation function $f$ and train an additional linear layer, as the classifier for the downstream task.

\label{methods}

\section{RELATED WORK}
%% MCH: moved semi- and self- basic paragraphs to BACKGROUND

% This study emphasizes methods that utilize the labeled set and unlabeled set at the same time during training and modify standard objectives for disciminative neural networks by adding additional loss term using unlabeled data. These methods are the current state-of-the-art for semi-supervised learning on image classification tasks.
% Multiple categories of semi-supervised learning method exists. But we will focus on the deep semi-supervised learning algorithms that utilize the labeled set and unlabeled set at the same time during training since these methods currently represent the state-of-the-art for SSL on image classification datasets.

Table~\ref{tab:related_work_comparison} summarizes key attributes of existing major benchmarks for semi-supervised and self-supervised methods. We now discuss how our study situates in this context.

\textbf{Comparison to Oliver et al.}
\citet{oliver2018realistic} provide an influential benchmark of deep semi-supervised methods. 
% covering several valuable angles,
% such as varying the amount of labeled and unlabeled data and comparing to transfer learning. 
Like our work, they highlight the pressing issue of using unrealistically large validation sets for hyperparameter tuning.
%in common academic practices. 
However, most of their experiments (e.g. all their tables and their figures 2-4) still use this unrealistic setting to be comparable to other works; only one subsection (Sec. 4.6) examines validation sets no larger than the train set.
Instead, \emph{we exclusively use validation set sizes no larger than training set}, mimicking what practitioners would face in real-applications. 
Our definition of a ``realistic'' size for a validation set -- no larger than the labeled train set -- is broader than \citeauthor{oliver2018realistic} 's definition of 10\% of the train set.
We favor our definition because  \citeauthor{oliver2018realistic} find specifically that ``for validation sets of the same size (100\%) as the training set, some differentiation between the approaches is possible.'' In contrast, they found reliable differentiation was not always possible with much smaller validation sets.
%endrevision

Furthermore, the benchmarking results presented by \citeauthor{oliver2018realistic} require each algorithm to complete ``1000 trials of Gaussian Process-based black-box optimization using Google Cloud ML Engine''. This resource-intensive process seems impractical for researchers outside well-funded industrial labs. We use more modest compute budgets and specifically allow each method the same total wallclock runtime for tuning, again obeying practical constraints.
%have limited practical applicability to many domain practitioners, such as clinical researchers seeking to apply semi-/self-supervised methods to medical datasets. The reported results were obtained through  for each algorithm, a resource-intensive process impractical for researchers outside well-funded industrial labs

% while \citeauthor{oliver2018realistic} stressed the importance of fair hyperparameter search, their reported results were based on ``1000 trials of Gaussian Process-based black box optimization using Google Cloud ML engine''. This is impractical for researchers who are not affiliated with highly-resourced industrial labs.

% Their focus on final performance also masks the dynamics of the performance versus computation tradeoff, which renders it less insightful for practitioners with new data at hand trying to seek the best outcome under limited available resources. To address this, we give each algorithm \emph{a fixed computation budget of some hours, and analyse how performance progresses over time.} 

\textbf{Other semi-supervised benchmarks.}
TorchSSL~\citep{zhang2021flexmatch} benchmarked eight popular semi-supervised learning algorithm using a unified codebase. The same group later extended that effort to natural language and audio processing in their \emph{USB} benchmark~\citep{wang2022usb}. 
% They further found that using pre-trained Transformer~\citep{vaswani2017attention,dosovitskiy2020image} on ImageNet-1K~\citep{russakovsky2015imagenet} could be beneficial to semi-supervised learning. However, for new medical imaging datasets, it is non-trivial to find a similar-enough dataset for such a pre-training (more discussion in~\ref{app:PracticalTransferLearning}).
Another recent effort for fine-grained classification by \citet{su2021realistic} evaluates semi-supervised learning on datasets that exhibit class imbalance and contains images from novel classes in the unlabeled set, studying the effect of different initializations and the contents of unlabeled data on the performance of semi-supervised methods. While making significant contributions, these works focused on semi-supervision almost exclusively and did not include multiple self-supervised learning methods, thus leaving open the questions ``\emph{How do the two paradigms compare? Which methods are best overall?}''

\textbf{Prior self-supervised benchmarks.}
%Benchmarking self-supervised learning methods is an important and challenging task that requires a systematic and comprehensive evaluation protocol. 
Several works have proposed different benchmarks for self-supervised methods. % on various tasks and domains. 
\citet{goyal2019scaling} introduced a benchmark that covers 9 different tasks, such as object detection and visual navigation. \citet{ericsson2021well} compared thirteen top self-supervised models on 40 downstream tasks. \citet{da2022solo} presented a library of self-supervised methods that can be easily plugged into different downstream tasks and datasets. However, these works mainly consider other self-SL methods without direct comparison to the semi-SL paradigm. % using the same codebase. 
Moreover, they mostly do not incorporate the use of a validation set for hyperparameter tuning \emph{at all}. This can lead to suboptimal accuracy at test time and complicate comparisons to methods that do tune.

% However, a few aspects of these benchmarks do not cover the realistic needs of practitioners. 
% %An examination of the current literature suggests that 
% First, they do not incorporate the use of a validation set (which can lead to suboptimal accuracy at test time and complicate comparisons to methods that do tune on validation sets).
% Second, they usually consider only the final performance, leaving out the accuracy changes over time. 
%account for the balance between computer resources and performance evaluation and  

\textbf{Self-supervision for medical images.}
Our work is complementary to recent efforts that assess self-supervised pipelines for medical image classification~\citep{aziziRobustDataefficientGeneralization2023,aziziBigSelfSupervisedModels2021}. They focus primarily on how to design multi-stage transfer learning pipelines and do not comprehensively compare many different self-supervised methods or \emph{any} semi-supervised methods. Further, many datasets studied by \citep{aziziRobustDataefficientGeneralization2023} come from proprietary projects conducted at Google.
In contrast, all our data and experiments are open and reproducible by others.

\textbf{Combining semi- and self-supervision.}
Recent research has explored the \emph{fusion} of semi-supervised learning and self-supervised learning ideas~\citep{zhai2019s4l,kim2021selfmatch,li2021comatch,zheng2022simmatch}.
Some recent surveys~\citep{qi2020small,chen2022semi}  compare semi- and self-supervised learning, but they focus on literature review while we offer realistic and comprehensive benchmarking experiments.

\label{related_work}

\section{DATASETS AND TASKS}
\begin{table}[!t]
\begin{tabular}{cc}
\multicolumn{1}{c}{TissueMNIST} &
\multicolumn{1}{c}{PathMNIST} \\
% Table for TissueMNIST and PathMNIST (First Row)
% \begin{minipage}[t]{.22\textwidth}
% \begin{small}
% \setlength{\tabcolsep}{0.3mm}
% \begin{tabular}{r r r r r}
%  & \multicolumn{3}{c}{Labeled} & Unlab. \\
%  & Train & Val & Test & Train \\
% \hline
% total &  400&  400& 47280  &  165066\\ 
% \hline
% \todo{cls2} & 15 & 15 & 1677 & 5851 \\
% cls1 & 19 & 19 & 2233 & 7795 \\
% cls5 & 19 & 19 & 2202 & 7686 \\
% cls4 & 28 & 28 & 3369 & 11761 \\
% cls3 & 37 & 37 & 4402 & 15369 \\
% cls7 & 59 & 59 & 7031 & 24549 \\ 
% cls6 & 95 & 95 & 11201 & 39108 \\
% cls0 & 128 & 128 & 15165 & 52947 \\
% \end{tabular}
% \end{small}
% \end{minipage}
\begin{minipage}[b]{.22\textwidth}
\begin{small}
\setlength{\tabcolsep}{0.3mm}
\begin{tabular}{r r r r r}
 & \multicolumn{3}{c}{Labeled} & Unlab. \\
 & Train & Val & Test & Train \\
\hline
total &  400&  400& 47280  &  165066\\ 
\hline
GE & 15 & 15 & 1677 & 5851 \\
DCT & 19 & 19 & 2233 & 7795 \\
POD & 19 & 19 & 2202 & 7686 \\
LEU & 28 & 28 & 3369 & 11761 \\
IE & 37 & 37 & 4402 & 15369 \\
TAL & 59 & 59 & 7031 & 24549 \\ 
PT & 95 & 95 & 11201 & 39108 \\
CD/CT & 128 & 128 & 15165 & 52947 \\
~ & \\ % added to make tables align
\end{tabular}
\end{small}
\end{minipage}
&
% Table for PathMNIST (First Row)
\begin{minipage}[b]{.22\textwidth}
\begin{small}
\setlength{\tabcolsep}{0.3mm}
\begin{tabular}{rrrrr}
 & \multicolumn{3}{c}{Labeled} & Unlab. \\
 & Train & Val & Test & Train \\
\hline
total &  450&  450& 7180 &  89546\\ 
\hline
NORM & 39 & 39 & 741 & 7847 \\
MUC & 40 & 40 & 1035 & 7966 \\
ADI & 47 & 47 & 1338 & 9319 \\
STR & 47 & 47 & 421 & 9354 \\ 
BACK & 48 & 48 & 847 & 9461 \\
DEB & 52 & 52 & 339 & 10308 \\
LYM & 52 & 52 & 643 & 10349 \\
MUS & 61 & 61 & 592 & 12121 \\
TUM & 64 & 64 & 1233 & 12821 \\ 
\end{tabular}
\end{small}
\end{minipage}
% \begin{minipage}[t]{.22\textwidth}
% \begin{small}
% \setlength{\tabcolsep}{0.3mm}
% \begin{tabular}{rrrrr}
%  & \multicolumn{3}{c}{Labeled} & Unlab. \\
%  & Train & Val & Test & Train \\
% \hline
% total &  450&  450& 7180 &  89546\\ 
% \hline
% cls6 & 39 & 39 & 741 & 7847 \\
% cls4 & 40 & 40 & 1035 & 7966 \\
% cls0 & 47 & 47 & 1338 & 9319 \\
% cls7 & 47 & 47 & 421 & 9354 \\ 
% cls1 & 48 & 48 & 847 & 9461 \\
% cls2 & 52 & 52 & 339 & 10308 \\
% cls3 & 52 & 52 & 643 & 10349 \\
% cls5 & 61 & 61 & 592 & 12121 \\
% cls8 & 64 & 64 & 1233 & 12821 \\ 
% \end{tabular}
% \end{small}
% \end{minipage}
\\
\multicolumn{1}{c}{TMED-2} &
\multicolumn{1}{c}{AIROGS} \\
% Table for TMED2 and AIROGS (Second Row)
\begin{minipage}[t]{.22\textwidth}
\begin{small}
\setlength{\tabcolsep}{0.3mm}
\begin{tabular}{lrrrr}
 & \multicolumn{3}{c}{Labeled} & Unlab. \\
    & Train & Val & Test & Train \\ 
\hline
total & 1660 & 235 & 2019 & 353500 \\ 
\hline
PSAX & 223 & 50 & 342 & - \\
A2C & 325 & 28 & 319 & - \\ 
A4C & 462 & 39 & 423 & - \\
PLAX & 650 & 118 & 935 & - \\
\end{tabular}
\end{small}
\end{minipage}
&
% Table for AIROGS (Second Row)
\begin{minipage}[t]{.22\textwidth}
\begin{small}
\setlength{\tabcolsep}{0.3mm}
\begin{tabular}{lrrrr}
 & \multicolumn{3}{c}{Labeled} & Unlab. \\
    & Train & Val & Test & Train \\ 
\hline
total & 600 & 600 & 6000 & 94242 \\ 
\hline
{\footnotesize Glaucoma} & 60 & 60 & 600 & - \\
{\footnotesize No Glauc.} & 540 & 540 & 5400 & - \\ 
~ \\
~ \\
\end{tabular}
\end{small}
\end{minipage}
\end{tabular}
\caption{
\textbf{Summary statistics of train/validation/test splits} for all datasets in our study. Each table's rows are arranged in ascending order based on the number of per-class labeled train images. Full description of class names can be found in App~\ref{app:class_description}.
}%end caption
\label{tab:dataset_splits}
\end{table}

We study four open-access medical image classification datasets, all with 2D images that are fully deidentified. Table~\ref{tab:dataset_splits} reports statistics for all train/test splits.
The exact splits used in our study can be found in our open-source codebase, documented in App.~\ref{app:code}.

%in this study, all with 2D images. 
Two datasets -- PathMNIST and TissueMNIST -- are selected from the MedMNIST collection~\citep{yang2023medmnist} (criteria in App.~\ref{app:dataset_selection}).
Prior experiments by \citet{yang2023medmnist} suggest their 28x28 resolution is a reasonable choice for rapid prototyping; using larger 224x224 resolution does not yield much more accurate classifiers for these two datasets.

Two other datasets represent more moderate resolutions closely tied to contemporary clinical research: the 112x112 Tufts Medical Echocardiogram Dataset (TMED-2) and the 384x384 AIROGS dataset.
Further details for each dataset are provided in a dedicated paragraph below.

%train/validation/test split counts by class, are provided in .

% \todo{MAYBE DELETE THIS? THE ARGUMENT 'KEEP COST AFFORDABLE' SEEMS CONTRADICT TO OUR 'RAN 5 RANDOMSEED' For each dataset, we ran experiments on \todo{just one} train/validation split to keep costs affordable.} 

For each dataset, our data splitting strategy closely mirrors the conditions of real SSL applications.
% in trying to apply SSL methods to a smaller labeled set and a much larger unlabeled set. 
First, we let labeled training and validation sets contain a natural distribution of classes even if that may be imbalanced. This reflects how data would likely be affordably collected (by random sampling) and avoids artificially balanced training sets that will not match the population an algorithm would encounter in a deployment. 
Second, to be realistic we ensure validation sets are never larger than the available labeled train sets.
This is in contrast to previous benchmarks:  \citet{wang2022usb}'s TissueMNIST hyperparameter search used a validation set of 23,640 images even though the labeled training set contained only 80 or 400 images. 
In practice, such a large validation set would almost certainly be re-allocated to improve training, as noted in \citet{oliver2018realistic}.

\textbf{TissueMNIST} is a lightweight dataset of 28x28 images of human kidney cortex cells, %segmented from 3 reference tissue specimens and 
organized into 8 categories. 
The original dataset is fully-labeled and distributed with predefined train/val/test split of 165,466/23,640/47,280 images with some class imbalance.
We assume a total labeling budget of 800 images, evenly split between training and validation (to ensure hyperparameter tuning can differentiate methods). 
% so that validation-based hyperparameter tuning may be reliable. 
We form a labeled training set of 400 images from the predefined training set, sampling each class by its frequency in the original training set.
%(in many medical applications, this frequency can often be determined using prior knowledge). 
We form a labeled validation set of 400 images sampled from the predefined validation set. 
%Our resulting data splits are in Table~\ref{tab:dataset_splits}. 
For unlabeled set, we keep all remaining images in the original training split, discarding known labels.

\textbf{PathMNIST} is another lightweight dataset of 28x28 images of patches from colorectal cancer histology slides that comprise of 9 tissue types. The original dataset is fully-labeled and distributed with predefined train/val/test split of 89,996/10,004/7,180 images. Class imbalance is less severe than TissueMNIST.
We assume a total labeled data budget of 900 images, again evenly split between training and validation.
Labeled train, labeled valid, and unlabeled sets are sampled via the same procedures as in TissueMNIST. 
For both MedMNIST datasets, we use the predefined test set.

% above.
%We form a labeled training set of 450 images from the predefined training set, sampling each class by its original train set frequency.

% \revision{It may be tempting to worry that 28x28 resolution is too small; however experiments by \citet{yang2023medmnist} suggest that on both Path and Tissue datasets, using larger resolution of 224x224 only results in minimum gain in performance compared to 28x28 (${<}0.01$ gain in AUROC).}

\textbf{TMED-2}~\citep{huang2021new,huang2022tmed} is an open-access dataset of 112x112 2D grayscale images captured from routine echocardiogram scans. Each scan produces dozens of ultrasound images of the heart, captured from multiple acquisition angles (i.e., different anatomic views). In this study, we adopt \citeauthor{huang2022tmed}'s view-classification task: the goal is to classify each image into one of 4 canonical view types (see App.~\ref{app:class_description}). 
%whether an image corresponds to parasternal long axis (PLAX), parasternal short axis (PSAX), apical 2-chamber (A2C) or apical 4-chamber (A4C) view. 
View classification is clinically important: measurements or diagnoses of heart disease can only be made when looking at the right anatomical view~\citep{madani2018fast,wessler2023automated}. 
% assuming if it is 2nd it is the split with id #1 in code
We used the provided train/validation/test split (id \#1), which is naturally imbalanced and matches our criteria for realistic sizing.
We further use the provided large unlabeled set of 353,500 images.
TMED-2's unlabeled set is both \emph{authentic} (no true labels are available at all, unlike other benchmarks that ``forget'' known labels) and \emph{uncurated}~\citep{huang2022fix} (contains images of view types beyond the 4 classes in the labeled task).

\textbf{AIROGS}~\citep{deventeAIROGSArtificialIntelligence2023} is a public dataset released for a recent competition where the binary classification task is to decide whether the patient should be referred for glaucoma or not, given a color fundus image of the retina (eye).
We selected this dataset because it represents an active research challenge even with all available labels. 
Furthermore, because labels for this dataset were acquired at great expense (multiple human annotators graded each of over 100k images), we hope our work helps assess how self-/semi-supervision could reduce the annotation costs of future challenges. 
We selected the 384x384 resolution favored by several high-performing competition entries.
We chose a labeling budget of 1200 total images, evenly split between train and validation; the rare positive class (9 to 1 imbalance) is representative of many screening tasks.
We took the rest of available images as the unlabeled train set.

\label{tasks_and_datasets}

\section{EXPERIMENTAL DESIGN}
\textbf{Performance metric.}
We use \emph{balanced accuracy}~\citep{guyon2015design,grandini2020metrics} 
as our primary performance metric.
For a task with $C\geq 2$ classes, let $y_{1:N}$ denote true labels for $N$ examples in a test set, and $\hat{y}_{1:N}$ denote a classifier's predicted labels. Let $\text{TP}_c$ count \emph{true positives} for class $c$ (number of correctly classified examples whose true label is $c$), and let $N_c$ count the total number of examples with true label $c$. Then we compute balanced accuracy (BA) as a percentage:
%There are class imbalance present in all the benchmarked dataset, which means standard accuracy is less suitable~\citep{huang2021new}.  
\begin{align}
\textsc{BA}(y_{1:N}, \hat{y}_{1:N}) = \frac{1}{C} \sum_{c=1}^{C} \frac{\text{TP}_{c}(y_{1:N}, \hat{y}_{1:N})}{N_{c}(y_{1:N})} \cdot 100\%
\label{eq:balanced_accuracy}
\end{align}
Balanced accuracy is more suitable than standard accuracy for imbalanced problems when each class matters equally.
The expected $\textsc{BA}$ of a uniform random guess is $\frac{100}{C}\%$.

On AIROGS, we also track  metrics recommended by its creators~\citep{deventeAIROGSArtificialIntelligence2023} for clinical utility: AUROC, partial AUROC (${>}$90\% specificity) and sensitivity-at-95\%-specificity.

%\paragraph{BRAINSTORM: Maybe we can also propose some evaluation metrics that take into consideration both the performance and compute?}

\textbf{Architectures.}
 CNN backbones are popular in medical imaging~\citep{esteva2017dermatologist,wessler2023automated,wu2019deep,ghorbani2020deep,ghayvat2023ai,billot2022robust,kushnure2023lim,gaur2023medical, lai2022smoothed, huo2023deep, huang2023detecting, lai2021joint}. We use ResNet-18~\citep{he2016deep} on Tissue and Path, and WideResNet-28-2~\citep{zagoruyko2016wide} on TMED-2. We experiment with both ResNet-18 and 50~\citep{he2016deep} on AIROGS to assess architectural differences. 

% \revision{Other architectures are possible, but preliminary results suggest a larger ResNet-50 did not substantially improve performance on AIROGS.}

% \revision{Other architectures are possible, but preliminary results suggest a larger ResNet-50 did not substantially improve performance on AIROGS.}

\textbf{Training with early stopping.}
For each training phase, we perform minibatch gradient descent on Eq.~\eqref{eq:objective_unified} with Adam \citep{kingma2014adam} optimizer and a cosine learning rate schedule~\citep{sohn2020fixmatch}.  
Each training phase proceeds for up to 200 epochs, where one epoch represents enough minibatch updates to process the equivalent of the entire combined training set $\mathcal{L} \cup \mathcal{U}$.
After every epoch, we record balanced accuracy on the validation set.
If this value plateaus for 20 consecutive epochs, we stop the current training phase early.
%\todo{What if it gets worse?}

%We employ an early stopping criterion: should the validation accuracy plateau for 20 consecutive epochs, we terminate the current training and proceed with the next hyperparameter choice.

\textbf{Hyperparameters.}
% and the performance of Self-supervised learning can also be dramatically impacted by the choice of hyperparameters~\citep{wagner2022importance}
Semi- and self-supervised learning can both be sensitive to hyperparameters~\citep{su2021realistic,wagner2022importance}.
Our evaluations tune both shared variables (e.g. learning rates, weight decay, unlabeled loss weight $\lambda^U$) as well as hyperparameters unique to each algorithm. See App.~\ref{App_Hyper_Details} for details on all hyperparameters for all methods.

\textbf{Unified procedure for training and hyperparameter tuning.}
We formulate a unified procedure mindful of realistic hardware and runtime constraints for a non-industrial-scale lab working on a new medical dataset. 
We assume each algorithm has access to one NVIDIA A100 GPU for a fixed number of hours.
Within the alotted compute budget, for each algorithm we execute a serial random search over hyperparameters, sequentially sampling each new configuration and then training until either an early stopping criteria is satisfied or the maximum epoch is reached.
We track the best-so-far classifier in terms of validation-set performance every epoch.
Algorithm~\ref{alg:hyperparam_tuning} provides pseudocode for the procedure.
Each time a new hyperparameter configuration is needed, we sample each hyperparameter value independently from a distribution designed to cover its common settings in prior literature (App.~\ref{App_Hyper_Details}). 
Our choice of random search on a budget is thought to yield better performance than grid search \citep{bergstra2012random}, fairly expends the same effort to train all methods regardless of their cost-per-epoch or hyperparameter complexity, and does not assume industrial-scale access to 1000 cloud-computing trials as in \citet{oliver2018realistic}.
We find that performance saturates after about 25 hours per 80000 unlabeled examples.
We thus allocate for the total time budget 25 hours for PathMNIST, 50 hours for TissueMNIST, and 100 hours for TMED-2. We allow 100 hours for AIROGS due to its larger resolution. Due to limited resources, we select only a few competitive methods to run on AIROGS based on the results on other datasets.

\textbf{Self-supervised classification phase.}
Self-supervised methods by definition do not utilize label information when training representation layer weights $v$.
To enable a proper comparison, for all self-SL methods our Alg.~\ref{alg:hyperparam_tuning}  introduces an additional classification layer, training it anew after each epoch, each time using a 10 trial random search for an L2-penalty regularization hyperparameter.
%ten times per epoch, 
We retain the best performing weights $w$ on the validation set. We emphasize that $w$ does not impact overall self-supervised training of $v$.

\textbf{Data augmentation.} 
%\textit{Semi-supervised:} 
Random flip and crop are used for all semi-supervised and supervised methods.
MixMatch also utilizes MixUp~\citep{zhang2017mixup}, while RandAugment~\citep{cubuk2020randaugment} is used by FixMatch and FlexMatch. For each self-supervised method as well as SupCon, we apply 
 the same SimCLR augmentation: random flip, crop, color jitter, and grayscale. %Gaussian blur.
%\todo{do we need to say ``including SwAV''? or provide more clarity here?}
%For supervised baselines ``labeled-only'' and ``MixUp'', we use random flip and crop.
%For ``Sup Contrast'', we follow common practice and use the same SimCLR augmentation above.

\textbf{Multiple trials.}
We repeat Alg.~\ref{alg:hyperparam_tuning} for each method across 5 separate trials (distinct random seeds). We record the \emph{mean balanced accuracy} on valid and test sets of these trials every 60 minutes (every 30 min. for Tissue and Path).

% We record the mean balanced accuracy across these trials on the test set every 60 minutes (every 30 min. for MedMNIST images). 
% When helpful, we also report variability across trials via the (min, max) interval of these 5 trials.

 %We use default augmentation choices for each algorithm from common practices. To be specific, random flip and crop are used for all semi-supervised learning algorithms. Additionally, MixUp~\citep{zhang2017mixup} is utilized in MixMatch, and  \textit{Self-Supervised Learning:} we utilize the same augmentation method for different algorithms. Specifically, we adopt the standard SimCLR augmentation method. This entails the use of random flip, crop, color jitter, grayscale, and Gaussian blur for all methods under consideration. As we are comparing the algorithm's ability, we generate two augmentations for each image in each iteration for all methods (including SwAV). \textit{Supervised-learning:} For MixUp and ``labeled-only'' baseline, we use random flip and crop. For supervised contrast, we use the SimCLR augmentation described above.

%\paragraph{Hardware}
%To ensure fair comparison, we use the same hardware, NVIDIA A100 GPU for each algorithm. 

\label{experiments}

\setlength{\tabcolsep}{1mm}
\newcommand{\WW}{0.32}
\begin{figure*}[t]
    \begin{tabular}{c c c}
        \includegraphics[width=\WW\textwidth]{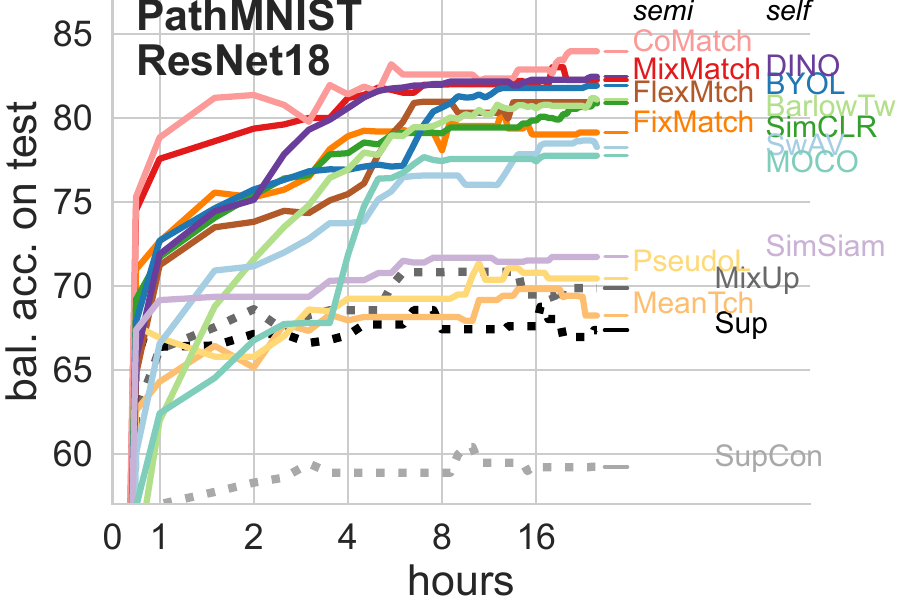}
        &
        \includegraphics[width=\WW\textwidth]{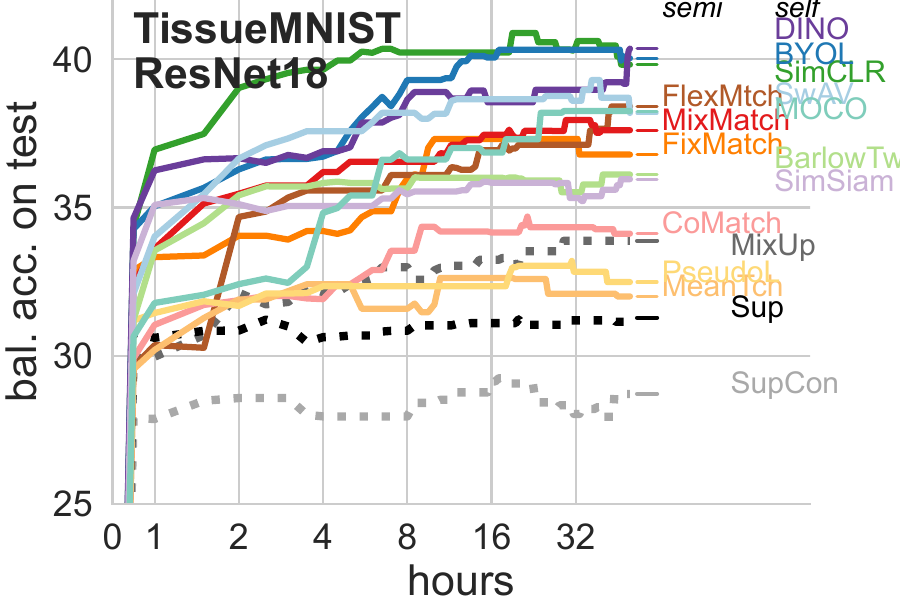}
        &
        \includegraphics[width=\WW\textwidth]{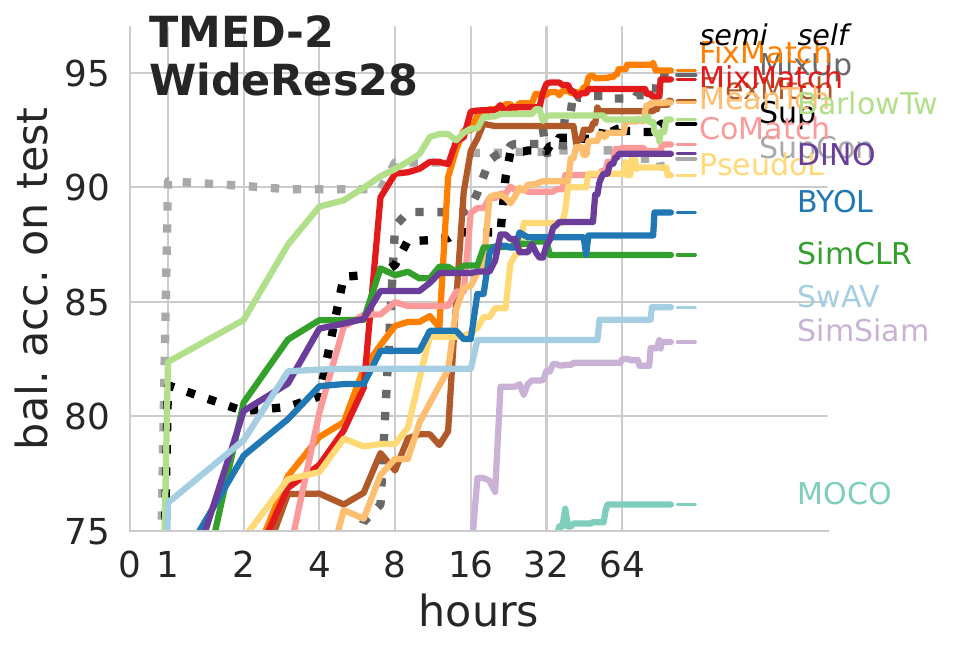}
        \\
        \includegraphics[width=\WW\textwidth]{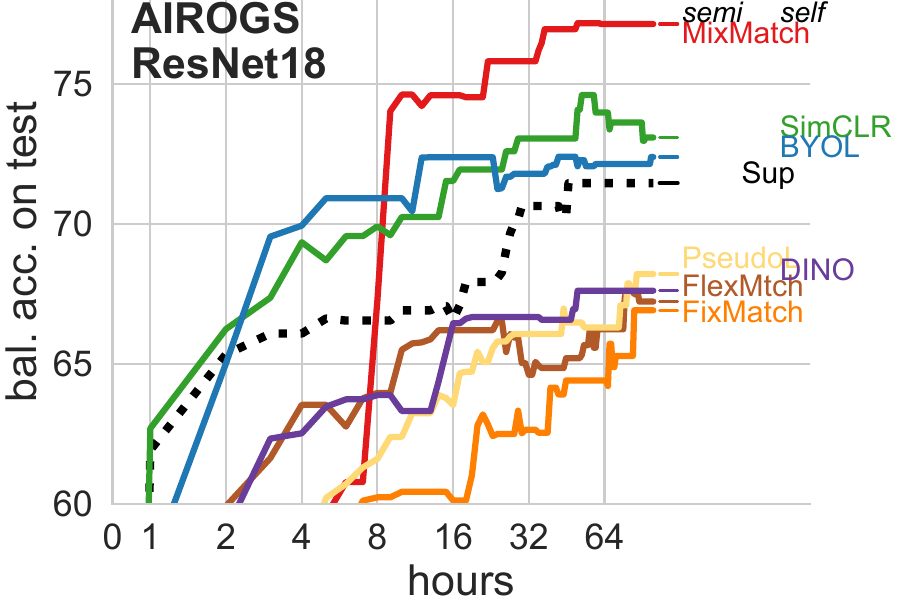}
        &
        \includegraphics[width=\WW\textwidth]{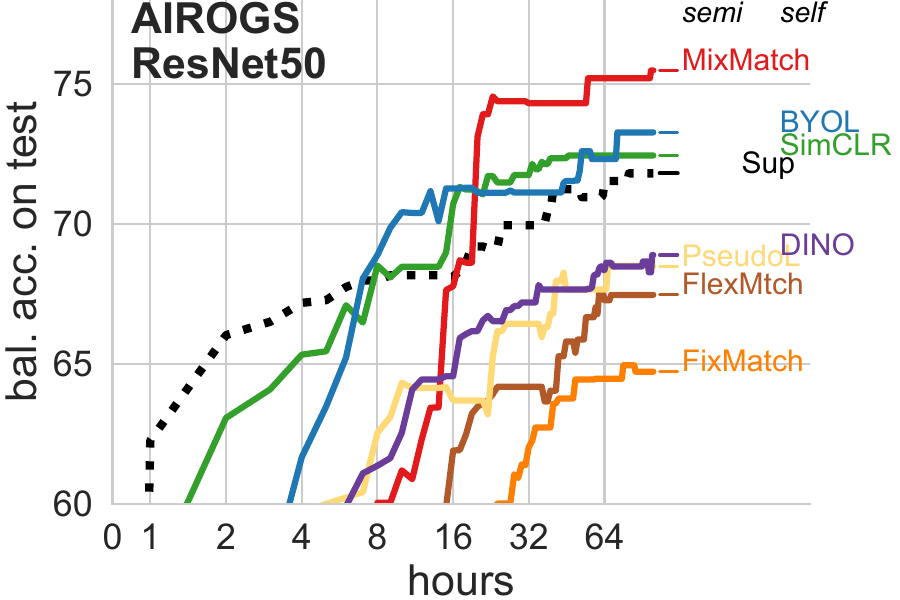}
        &
        \includegraphics[width=\WW\textwidth]{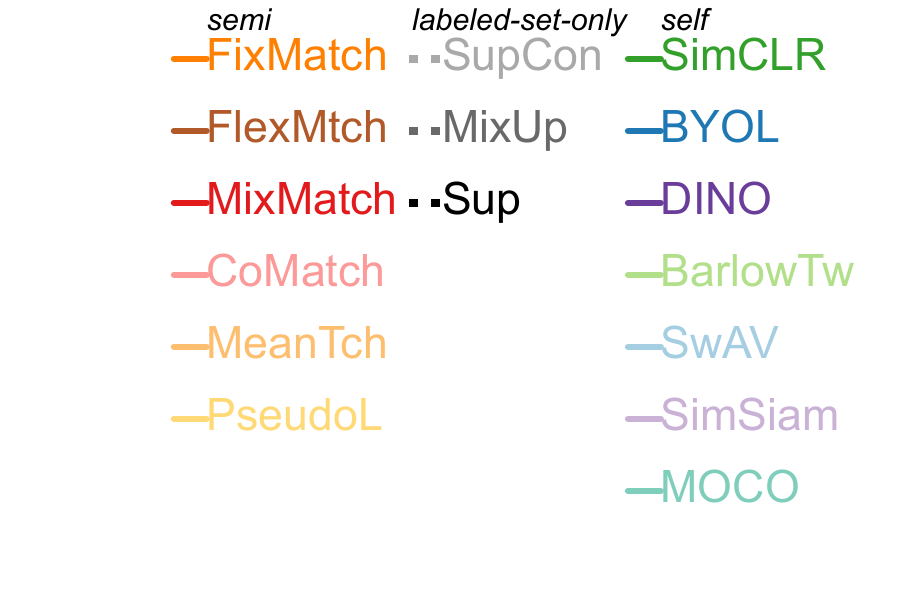}
    \end{tabular}
\caption{\textbf{Balanced accuracy over time profiles of semi- and self-supervised methods} across all 4 datasets. At each time, we report the test set bal.~acc.~of each method (mean over 5 trials of Alg.~\ref{alg:hyperparam_tuning}). Best viewed electronically.
\emph{Top Row:} On these 3 datasets, we compare all 6 semi- and 7 self- methods (see legend in lower right, citations in Sec.~\ref{methods}), to 3 labeled-set-only baselines.
\emph{Bottom Row:} 
On larger AIROGS dataset, we compare selected methods representative of the best in the top row.
Thin lines show final performance to ease comparison.
From these charts, we suggest that our unified training and tuning (Alg.~\ref{alg:hyperparam_tuning}) is effective, as all methods show gains over time. 
%Method rankings vary across datasets and runtime budgets; specialized tuning for each new dataset is recommended.
}%endcaption
\label{fig:test_performance_vs_time}
%On TissueMNIST (a), SimClr (orange) and BYOL (yellow) are dominant from 1 hour onward. On PathMNIST (b), MixMatch (red) and CoMatch (olive) are dominant from 1 hour onward, On TMED-2 (c), Sup Contrast wins for the first 7 hours, then MixMatch overtakes it, eventually joined (not surpassed) by FixMatch and FlexMatch.}    
\end{figure*}

\section{RESULTS \& ANALYSIS}
% \todo{1. MixMatch even better than FixMatch and FlexMatch}
% \todo{2. SimCLR, BYOL always better than other self-supervised}
% \todo{3. Empirically using small validation set for choosing hyperparameter and selecting checkpoint works}
% \todo{4. Self seems have smaller generalization error than semi}
% \todo{HYPOTHESIS: we hypothesize that self is less sensitive to class imbalance, as it achieve better performance on Tissue}
% \todo{5. Taking the off-the-shelf hyperparameter from closely related dataset work reasonably well, while from distant dataset work less well.}
% \todo{6. For dataset with abundant labels per class, supervised learning is very competitive against semi and self.}
% \todo{7. HYPOTHESIS: self is less affected by class imbalance}

%\subsection{Effectiveness of hyperparameter tunning in resource-constraint setting and Comparison between Semi and Self-supervised learning}

In each panel of Fig.~\ref{fig:test_performance_vs_time}, we focus on one dataset and architecture, plotting for each algorithm one line showing test set balanced accuracy (averaged across 5 trials) as a function of time spent running Alg.~\ref{alg:hyperparam_tuning}. Variance across trials is visualized in Fig~\ref{fig:tmedfigures_timepoint}. Extra results on AIROGS showing other performance metrics over time (AUROC, partial AUROC, and sensitivity at 95\% specificity) are found in Fig.~\ref{fig:test_performance_vs_time_AIROGS}. 
%we analysed the test set performance (balanced accuracy,  on AIROGS (averaged across 5 trials). 
We emphasize that following best practice, each checkpoint represents the best hyperparameters as selected on the validation set; we then report that checkpoint's test set performance. These results help us answer two key questions:

\textbf{Is hyperparameter tuning worthwhile for SSL methods when validation set sizes cannot be larger than the train set?} Across Fig.~\ref{fig:test_performance_vs_time} and \ref{fig:test_performance_vs_time_AIROGS}, all 16 algorithms show roughly monotonic improvements in test performance over time, despite using a realistic-sized validation set.
%Final performance is significantly better after hyperparameter tuning for 
The final performance of every algorithm and dataset shows improvement due to our unified training and tuning procedure
(further discussion in App.~\ref{app:Effectivness of Hyperparameter Tuning}). Therefore, we answer: \emph{Yes, realistically-sized validation sets for hyperparameter tuning and checkpoint selection can be effective.}

\textbf{What are the best SSL methods?}. Among the many methods we study in Fig.~\ref{fig:test_performance_vs_time}, no method clearly outperforms others across all datasets.
On Path, CoMatch, MixMatch, DINO, and BYOL perform best. 
On Tissue, self-supervised methods like DINO, BYOL, and SimCLR score well. 
On TMED-2, ultimately only FixMatch and MixMatch are competitive with the MixUp baseline.
On AIROGS, only MixMatch, SimCLR and BYOL beat the Sup baseline.

For each semi- and self- SSL method, Tab.~\ref{tab:gains} reports its relative gain in balanced accuracy over the best labeled-set-only supervised baseline on each dataset. We conclude that \emph{MixMatch represents the best overall choice} as it consistently performs at or near the top across all tasks. It is the only method to never deliver results worse by more than 1 percentage point than the best labeled-set-only baseline (see ``worst'' column of Tab.~\ref{tab:gains}). \emph{This is in stark contrast to USB~\citep{wang2022usb}, where FixMatch and FlexMatch are ranked notably higher that MixMatch.} Such differences stress the importance of choosing proper evaluation protocol tailored to specific needs. 
We do caution that \emph{hyperparameter tuning is strongly recommended for MixMatch to succeed on a new dataset} (see AIROGS result in Tab.~\ref{tab:hyper-strategy-comparison}).

\begin{table}[!t]
\begin{tabular}{l | rrrr | rr}
                  & \multicolumn{6}{c}{Gain in Bal. Acc. over best labeled-set-only} \\
             name &  Path &  Tissue &  TMED &  {\small AIROGS} &  median &  worst \\
\midrule
MixMatch &          12.4 &             3.7 &          -0.2 &             5.7 &          4.7 &        -0.2 \\
\emph{MOCO} &           7.9 &             4.4 &         -18.8 &             n/a &          4.4 &       -18.8 \\
 \emph{SwAV} &           8.4 &             4.3 &          -10.1 &             n/a &          4.3 &        -10.1 \\
\emph{SimCLR} &          11.0 &             5.9 &          -7.8 &             1.6 &          3.8 &        -7.8 \\
 \emph{BYOL} &          12.1 &             6.2 &          -6.0 &             0.9 &          3.5 &        -6.0 \\
\emph{BarlowTw} &          11.2 &             2.2 &          -1.9 &             n/a &          2.2 &        -1.9 \\

\emph{SimSiam} &           1.9 &             2.1 &         -11.6 &             n/a &          1.9 &       -11.6 \\
FlexMtch &          11.1 &             4.5 &          -1.1 &            -4.2 &          1.7 &        -4.2 \\
 \emph{DINO} &          12.6 &             6.5 &          -3.4 &            -3.8 &          1.6 &        -3.8 \\
FixMatch &           9.3 &             2.9 &           0.2 &            -4.5 &          1.6 &        -4.5 \\
CoMatch &          14.1 &             0.2 &          -3.0 &             n/a &          0.2 &        -3.0 \\
%BEST-LBL-SET-ONLY &           0.0 &             0.0 &           0.0 &             0.0 &          0.0 &         0.0 \\
MeanTch &          -1.6 &            -1.9 &          -1.2 &             n/a &         -1.6 &        -1.9 \\
PseudoL &           0.6 &            -1.4 &          -4.4 &            -3.2 &         -2.3 &        -4.4 \\
\hline
ref. val. & 69.8 & 33.9 & 94.9 & 71.5
          \\
\end{tabular}
\caption{
\textbf{Gains from SSL methods over only using labeled set}, across 4 datasets. \emph{Self (italicized)}, Semi (normal).
We report gains in percentage balanced accuracy (higher is better) in the mean final test set performance (averaged over 5 runs of Alg.~\ref{alg:hyperparam_tuning}) over a reference value that represents the best labeled-set-only run among Sup (minimize cross entropy), SupCon, and MixUp.
MixMatch has the highest median gain, and is the only method to never score notably worse ($>$ 1 point) than the best labeled set only run.
}%endcaption
\label{tab:gains}
\end{table}

\textbf{Pretraining vs.~from scratch.} Plots in App.~\ref{app:results} show accuracy-over-time profiles using initial weights pretrained on ImageNet.
%Similarly, we observed that hyperparameter tuning with realistically-sized validation set is effective, demonstrating that the analysis extends to scenarios involving pretrained weights.
Compared to training from scratch, we see only slight gains (less than 3 points of BA) in ultimate test-set accuracy when top methods use pretraining, which aligns with observations in~\citep{raghu2019transfusion} but counters the ``large margin'' gains (sometimes over 10 points) reported by \citet{su2021realistic}.
Pretraining's benefits still include improved convergence speed and reduced variance across trials.

\textbf{ResNet-18 vs. ResNet-50.}
The bottom row of Fig~\ref{fig:test_performance_vs_time} compares ResNet-18 and ResNet-50 architectures on AIROGS.
We broadly find that the method rankings are similar.
More notably, ResNet-50 is not substantially better than ResNet-18 in this task when compared using the same runtime budget. 
Profiles of other performance metrics in App~\ref{fig:test_performance_vs_time_AIROGS} suggest the same conclusions.
%The results align with those obtained for ResNet-18 in Figure~\ref{fig:test_performance_vs_time}, highlighting the generalizability of our findings across different model architectures.

\textbf{Tuning vs.~Transferring from another dataset.} To make the most of limited labeled data, one strategy is to use the entire labeled set for training, reserving no validation set at all. This approach relies on pre-established ``best'' hyperparameters sourced from other datasets and transferred to the target task, obviating the need for data-specific tuning. \citet{su2021realistic} employed this strategy in their benchmark, avoiding tuning due to concerns that small validation sets would yield unreliable selection. 
However, this claim in their work was not supported by experiments.
Here, we directly compare our hyperparameter tuning strategy (Alg.~\ref{alg:hyperparam_tuning}) with \citeauthor{su2021realistic}'s transfer strategy. We train the latter on the combined train + validation sets, using ``best'' hyperparameters from CIFAR-10 and Tissue (details in App.~\ref{App:Hyper_Transfer_strategy}). Tab.~\ref{tab:hyper-strategy-comparison} reports the final test-set balanced accuracy of each strategy on Path, TMED-2 and AIROGS.

%% MCH moved details to supplement
%(100 epochs on PathMNIST and AIROGS, and 80 g epochs on TMED2)
%Training is terminated early if the train loss does not improve over 20 consecutive epochs (details in App.~\ref{App:Hyper_Transfer_strategy}). Empirically, we observe that all models which did not trigger early stopping reached a plateau in training loss.

\begin{table*}[!t]
    \begin{minipage}{\textwidth}
        \centering
        \setlength{\tabcolsep}{1mm}
        \begin{tabular}{l l r r r c | r r r c | r r r }
            && \multicolumn{3}{c}{PathMNIST} &\hspace{2mm} & \multicolumn{3}{c}{TMED-2} &\hspace{2mm} & \multicolumn{3}{c}{AIROGS}\\
            \hline
            \multicolumn{2}{c}{Hyperparam. strategy:}
             & 
                Tune & Best from  & Best from
             && Tune & Best from  & Best from
             && Tune & Best from  & Best from
             \\
             &&  &  Tissue  & CIFAR-10 
             &&  &  Tissue  & CIFAR-10 
             &&  &  Tissue  & CIFAR-10 \\
            \hline
            semi
            %MixMatch
            & MixMatch & \ggF 82.24 & \ggF \textbf{83.87}  & \rrF 73.51 
            && \ggF \textbf{94.71} & \ggF 92.67  & \rrF 74.73 
            && \ggF \textbf{77.14} & \rrF 61.75   & \rrF Diverged 
            \\
            %CoMatch
            & CoMatch & \ggF 83.95 & \rrF 80.51  & \ggF \textbf{86.42} 
            && \ggF 91.87 & \ggF 92.51 & \ggF \textbf{92.67} 
            &&  & &  
            \\
            & FixMatch & \yyF 79.11 & \ggF \textbf{82.64} & \yyF 79.73 
            && \ggF 95.06 & \ggF 95.12 & \ggF \textbf{95.73} 
            && \ggF \textbf{66.92} & \ggF 66.31  & \ggF 65.50 
            \\ 
            & FlexMatch & \ggF \textbf{80.91} & \ggF 78.70 & \ggF 80.23 
            && \ggF 93.77 & \ggF \textbf{93.97} & \ggF 93.70 
            && \rrF 67.23 & \ggF 72.82  & \ggF \textbf{74.52} 
            \\ 
            & Pseudo-Label & \ggF \textbf{70.42} & \yyF  66.12 & \rrF 64.71  
            && \ggF \textbf{90.52} & \ggF 88.78  & \rrF 82.13 
            && \ggF \textbf{68.21} & \yyF 64.43  & \ggF 66.93 \\ 
            
            & Mean-Teacher & \ggF \textbf{68.21} & \rrF 55.90  & \rrF 56.13 
            && \ggF \textbf{93.70} & \ggF 91.21 & \rrF 58.26 
            &&  & &   \\ 
%            \vspace{0.1mm}
%            \\
            \hline
            self
            &SimCLR & \ggF \textbf{81.05} & \ggF 79.86 & \ggF {80.05}
            && \ggF {87.04} & \yyF 84.41 &  \ggF \textbf{87.24}
            && \ggF \textbf{73.01} & \rrF 67.15  & \rrF 67.95
            \\ 
            &BYOL & \ggF \textbf{81.79} & \ggF 80.35  & \ggF {80.16} 
            && \ggF \textbf{88.90} & \ggF {88.41}  & \yyF 85.15
            && \ggF \textbf{72.40} & \ggF {70.13}  & \ggF 70.63
            \\ 
            &SwAV & \yyF 77.90 & \ggF 78.24 & \ggF \textbf{81.30}
            && \ggF \textbf{84.76} & \yyF  82.17 & \yyF  82.20
            &&  &  & 
            \\ 
            &MoCo & \ggF {77.73} & \ggF \textbf{79.95} & \ggF {79.12}
            && \yyF 76.14 & \ggF {78.56} & \ggF \textbf{79.11}
            &&  &   & 
            \\ 
            &SimSiam & \ggF \textbf{71.78} & \yyF 68.62  & \ggF 70.44 
            && \ggF \textbf{83.24} & \ggF 80.93 & \yyF {80.20}
            &&  &  &   \\ 
            &BT & \ggF \textbf{80.85} & \rrF 72.79 & \ggF 80.25 
            && \ggF \textbf{92.96} & \ggF 91.63 & \ggF 90.81 
            &&  &   &  \\ 
            &DINO & \ggF \textbf{82.52} & \rrF 77.50  & \ggF 80.67
            && \ggF \textbf{91.45} &  \rrF 84.80 & \ggF {90.03}
            && \ggF \textbf{67.61} &  \ggF {66.46} & \ggF {67.13} \\ 
        \end{tabular}
    \end{minipage}%
    \\
    \caption{
    \textbf{Hyperparameter strategy evaluation.}
    Table reports ultimate test-set balanced accuracy as percentage (mean over 5 trials, higher is better) on three target datasets: PathMNIST, TMED-2, and AIROGS.
    3 different hyperparameter strategies are compared for each SSL method. Due to computation constraints, we only select a few algorithms to run on AIROGS, explained in Sec.~\ref{experiments}.
    \emph{Tune} strategy: tuning for best hyperparameters via our proposed Alg.~\ref{alg:hyperparam_tuning} for 100 hours (TMED-2 and AIROGS) and 25 hours (PathMNIST), using provided train/validation split.
    \emph{Best from Tissue/CIFAR-10} strategy: 
    just one training phase on the target dataset using the ``best'' hyperparameters selected from the named source data.
    These runs combine train and validation data for fitting following \citet{su2021realistic}, and typically take 1-4/5-30/10-40 hours on Path/TMED-2/AIROGS depending on the algorithm and whether early stopping is triggered.
    To highlight which strategies are most effective, we bold the best strategy within each dataset-specific row, and color cells relative to this bold value as green (within 2.5 of best), yellow (within 5 of best), and red (worse by 5 or more points).
%    ``Tissue Time'' reports compute time required for each ``from Tissue'' result.
    %    ``Tissue Time'' columns show the compute time each algorithm used when transferring from Tissue's selected hyperparameters}
    % We sort each method (rows) by peak performance, from best to worst.  
    %\todo{Add column of runtime for the ``best from Tissue'' runs? then can better account for runtime-accuracy tradeoff.}
    }%endcaption
    \label{tab:hyper-strategy-comparison}
\end{table*}

Our 
hyperparameter tuning strategy is competitive across all datasets: the ``Tune'' columns of Tab.~\ref{tab:hyper-strategy-comparison} have the highest fraction of green cells. 
The alternative transfer strategy is faster and occasionally yield reasonable results. However, there is no guarantee of good performance from hyperparameter transfer, as evidenced by the inferior cases in Tab.~\ref{tab:hyper-strategy-comparison}~highlighted in yellow and red. There is also no consistent distinction between transferring from Tissue (an arguably more related medical dataset) and transferring from CIFAR-10, underscoring the challenge of identifying a ``closely related'' dataset for hyperparameter transfer.

% \revision{TODO HZ: FIX? On the other hand, transfer of best hyperparameters from another medical dataset (``from Tissue'') without using any validation set seems to work reasonably for many algorithms and cases. Transfer of best hyperparameters from less similar datasets, such as the non-medical CIFAR-10, is less preferred and more often lead to worse performance, even extreme case of model not able to learn (MixMatch on AIROGS).}

Overall, we recommend researchers pursuing SSL for a new medical classification task should adopt our hyperparameter tuning strategy. Transferring off-the-shelf hyperparameters has highly variable performance and can even result in divergence (see MixMatch on AIROGS in Tab.~\ref{tab:hyper-strategy-comparison}). Even with reasonable hyperparameters, having no validation set makes checkpoint selection challenging.  Adjacent epochs that exhibit similar low training losses may differ by over 5\% in test accuracy in our experiments. Improving transfer from existing hyperparameters to a new task is an interesting further direction of research.

\label{benchmarking_results}

\section{DISCUSSION \& CONCLUSION}
We have contributed a benchmark that helps practioners quantify what gains in a classification task are possible from the addition of unlabeled data and which methods help achieve them.
We offer a unified approach to training and hyperparameter selection of semi-supervised methods, self-supervised methods, and supervised baselines that is both affordable and realistic for research labs tackling new challenges without industrial-scale resources.
%\todo{IS THIS PRECISE? For other methods beyond MixMatch, we found hyperparameter search on the target set improves performance dramatically.}

%\textbf{Future work.}
%Could compare to newer methods.

\textbf{Limitations.}
%Our findings may not generalize beyond these 4 datasets.
%Our extensive experiments are limited to the datasets we selected. 
We deliberately focused on a modest number of labeled examples (30 - 1000) per class, motivated by projects where substantive effort has already gone into labeled data collection.
Applications in more scarce-label regimes (e.g. zero-shot or few-shot) may need to also consult other benchmarks, as should those looking at hundreds of fine-grained classes. We also did not specifically study situations where the distribution of labeled and unlabeled data significantly differs~\citep{oliver2018realistic,huang2022fix,saito2021openmatch,huang2024interlude}.

To enable thorough experiments, we focused on moderate ResNet architectures that have proven effective in clinical applications~\citep{wu2019deep,huang2023detecting}. As ViT-based methods become popular in medical imaging~\citep{ahmadi2023transformer,huang2024semi}, it would be valuable to understand if similar findings hold with vision transformers~\citep{dosovitskiy2020image}. We leave this to future work. Ideally, our results would be verified across multiple train-validation splits; however the resources needed remain prohibitive.

Our analysis here is limited the specific metrics of balanced accuracy and sensitivity/specificity (for AIROGS). Other clinically useful metrics, such as AUPRC, calibration, or net benefit, may be needed to decide if a classifier is appropriate for deployment~\citep{steyerberg2014towards}.
For medical applications, it is also key to understand fairness across subpopulations 
\citep{celiSourcesBiasArtificial2022} to avoid propagating structural disadvantages.

%we did not extensively study pretraining then semi-supervised.
%other evidence suggests this can be doable (cite UMass paper).
%For medical contexts, custom resolutions and color schemes (cite paper ethan found) might limit success of generic foundation models.

%focused on a modest number of classes. Fine-grained classification into many dozens of classes, especially those with long-tail behavior TODO CITE, may require separate benchmarks.

\textbf{Broader impact.}
All data analyzed here represent fully-deidentified open-access images approved for widespread use by their creators. We think the benefit of promoting these medical tasks to advance ML research outweighs the slight risk of patient reidentification by a bad actor.

\textbf{Outlook.}
Our experiments show that real benefits from the addition of unlabeled data are sometimes possible: our recommended methods see gains of +5 points of balanced accuracy on TissueMNIST, +10 points on PathMNIST, and +5 points on AIROGS against strong labeled-set-only baselines.
In contrast, most tested methods do not add significant gains on TMED-2, perhaps due to that benchmark's \emph{uncurated} nature~\citep{huang2022fix}.
We hope our work enables the research community to convert decades of effort on SSL into improved patient outcomes and better scientific understanding of disease and possible treatments. 
We further hope that our benchmark inspires those that pursue improved methodological contributions to favor realistic evaluation protocols and clinically-relevant datasets.

\label{discussion}

{
    \small
    \bibliographystyle{ieeenat_fullname}
    \bibliography{cvpr_main}

\begin{thebibliography}{90}
\providecommand{\natexlab}[1]{#1}
\providecommand{\url}[1]{\texttt{#1}}
\expandafter\ifx\csname urlstyle\endcsname\relax
  \providecommand{\doi}[1]{doi: #1}\else
  \providecommand{\doi}{doi: \begingroup \urlstyle{rm}\Url}\fi

\bibitem[Ahmadi et~al.(2023)Ahmadi, Tsang, Gu, Tsang, and
  Abolmaesumi]{ahmadi2023transformer}
N Ahmadi, MY Tsang, AN Gu, TSM Tsang, and P Abolmaesumi.
\newblock Transformer-based spatio-temporal analysis for classification of
  aortic stenosis severity from echocardiography cine series.
\newblock \emph{IEEE Transactions on Medical Imaging}, 2023.

\bibitem[Azizi et~al.(2021)Azizi, Mustafa, Ryan, Beaver, Freyberg, Deaton, Loh,
  Karthikesalingam, Kornblith, Chen, Natarajan, and
  Norouzi]{aziziBigSelfSupervisedModels2021}
Shekoofeh Azizi, Basil Mustafa, Fiona Ryan, Zachary Beaver, Jan Freyberg,
  Jonathan Deaton, Aaron Loh, Alan Karthikesalingam, Simon Kornblith, Ting
  Chen, Vivek Natarajan, and Mohammad Norouzi.
\newblock Big {{Self-Supervised Models Advance Medical Image Classification}}.
\newblock In \emph{International {{Conference}} on {{Computer Vision}}
  ({{ICCV}})}. {arXiv}, 2021.

\bibitem[Azizi et~al.(2023)Azizi, Culp, Freyberg, Mustafa, Baur, Kornblith,
  Chen, Tomasev, Mitrovi{\'c}, Strachan, Mahdavi, Wulczyn, Babenko, Walker,
  Loh, Chen, Liu, Bavishi, McKinney, Winkens, Roy, Beaver, Ryan, Krogue,
  Etemadi, Telang, Liu, Peng, Corrado, Webster, Fleet, Hinton, Houlsby,
  Karthikesalingam, Norouzi, and
  Natarajan]{aziziRobustDataefficientGeneralization2023}
Shekoofeh Azizi, Laura Culp, Jan Freyberg, Basil Mustafa, Sebastien Baur, Simon
  Kornblith, Ting Chen, Nenad Tomasev, Jovana Mitrovi{\'c}, Patricia Strachan,
  S.~Sara Mahdavi, Ellery Wulczyn, Boris Babenko, Megan Walker, Aaron Loh,
  Po-Hsuan~Cameron Chen, Yuan Liu, Pinal Bavishi, Scott~Mayer McKinney, Jim
  Winkens, Abhijit~Guha Roy, Zach Beaver, Fiona Ryan, Justin Krogue, Mozziyar
  Etemadi, Umesh Telang, Yun Liu, Lily Peng, Greg~S. Corrado, Dale~R. Webster,
  David Fleet, Geoffrey Hinton, Neil Houlsby, Alan Karthikesalingam, Mohammad
  Norouzi, and Vivek Natarajan.
\newblock Robust and data-efficient generalization of self-supervised machine
  learning for diagnostic imaging.
\newblock \emph{Nature Biomedical Engineering}, 7\penalty0 (6):\penalty0
  756--779, 2023.

\bibitem[Bergstra and Bengio(2012)]{bergstra2012random}
James Bergstra and Yoshua Bengio.
\newblock Random search for hyper-parameter optimization.
\newblock \emph{Journal of machine learning research}, 13\penalty0 (2), 2012.

\bibitem[Berthelot et~al.(2019)Berthelot, Carlini, Goodfellow, Papernot,
  Oliver, and Raffel]{berthelot2019mixmatch}
David Berthelot, Nicholas Carlini, Ian Goodfellow, Nicolas Papernot, Avital
  Oliver, and Colin~A Raffel.
\newblock Mixmatch: A holistic approach to semi-supervised learning.
\newblock \emph{Advances in neural information processing systems}, 32, 2019.

\bibitem[Billot et~al.(2022)Billot, Magdamo, Arnold, Das, and
  Iglesias]{billot2022robust}
Benjamin Billot, Colin Magdamo, Steven~E Arnold, Sudeshna Das, and Juan~Eugenio
  Iglesias.
\newblock Robust segmentation of brain mri in the wild with hierarchical cnns
  and no retraining.
\newblock In \emph{International Conference on Medical Image Computing and
  Computer-Assisted Intervention}, pages 538--548. Springer, 2022.

\bibitem[Blum and Mitchell(1998)]{blum1998combining}
Avrim Blum and Tom Mitchell.
\newblock Combining labeled and unlabeled data with co-training.
\newblock In \emph{Proceedings of the eleventh annual conference on
  Computational learning theory}, pages 92--100, 1998.

\bibitem[Brown et~al.(2020)Brown, Mann, Ryder, Subbiah, Kaplan, Dhariwal,
  Neelakantan, Shyam, Sastry, Askell, et~al.]{brown2020language}
Tom Brown, Benjamin Mann, Nick Ryder, Melanie Subbiah, Jared~D Kaplan, Prafulla
  Dhariwal, Arvind Neelakantan, Pranav Shyam, Girish Sastry, Amanda Askell,
  et~al.
\newblock Language models are few-shot learners.
\newblock \emph{Advances in neural information processing systems},
  33:\penalty0 1877--1901, 2020.

\bibitem[Caron et~al.(2018)Caron, Bojanowski, Joulin, and Douze]{caron2018deep}
Mathilde Caron, Piotr Bojanowski, Armand Joulin, and Matthijs Douze.
\newblock Deep clustering for unsupervised learning of visual features.
\newblock In \emph{Proceedings of the European conference on computer vision
  (ECCV)}, pages 132--149, 2018.

\bibitem[Caron et~al.(2020)Caron, Misra, Mairal, Goyal, Bojanowski, and
  Joulin]{caron2020unsupervised}
Mathilde Caron, Ishan Misra, Julien Mairal, Priya Goyal, Piotr Bojanowski, and
  Armand Joulin.
\newblock Unsupervised learning of visual features by contrasting cluster
  assignments.
\newblock \emph{Advances in neural information processing systems},
  33:\penalty0 9912--9924, 2020.

\bibitem[Caron et~al.(2021)Caron, Touvron, Misra, J{\'e}gou, Mairal,
  Bojanowski, and Joulin]{caron2021emerging}
Mathilde Caron, Hugo Touvron, Ishan Misra, Herv{\'e} J{\'e}gou, Julien Mairal,
  Piotr Bojanowski, and Armand Joulin.
\newblock Emerging properties in self-supervised vision transformers.
\newblock In \emph{Proceedings of the IEEE/CVF international conference on
  computer vision}, pages 9650--9660, 2021.

\bibitem[Cascante-Bonilla et~al.(2021)Cascante-Bonilla, Tan, Qi, and
  Ordonez]{cascante2021curriculum}
Paola Cascante-Bonilla, Fuwen Tan, Yanjun Qi, and Vicente Ordonez.
\newblock Curriculum labeling: Revisiting pseudo-labeling for semi-supervised
  learning.
\newblock In \emph{Proceedings of the AAAI Conference on Artificial
  Intelligence}, 2021.

\bibitem[Celi et~al.(2022)Celi, Cellini, Charpignon, Dee, Dernoncourt, Eber,
  Mitchell, Moukheiber, Schirmer, et~al.]{celiSourcesBiasArtificial2022}
Leo~Anthony Celi, Jacqueline Cellini, Marie-Laure Charpignon,
  Edward~Christopher Dee, Franck Dernoncourt, Rene Eber, William~Greig
  Mitchell, Lama Moukheiber, Julian Schirmer, et~al.
\newblock Sources of bias in artificial intelligence that perpetuate healthcare
  disparities\textemdash{{A}} global review.
\newblock \emph{PLOS Digital Health}, 1\penalty0 (3), 2022.

\bibitem[Chapelle et~al.(2009)Chapelle, Scholkopf, and Zien]{chapelle2009semi}
Olivier Chapelle, Bernhard Scholkopf, and Alexander Zien.
\newblock Semi-supervised learning (chapelle, o. et al., eds.; 2006)[book
  reviews].
\newblock \emph{IEEE Transactions on Neural Networks}, 20\penalty0
  (3):\penalty0 542--542, 2009.

\bibitem[Chen et~al.(2020{\natexlab{a}})Chen, Qin, Qiu, Ouyang, Wang, Chen,
  Tarroni, Bai, and Rueckert]{chen2020realistic}
Chen Chen, Chen Qin, Huaqi Qiu, Cheng Ouyang, Shuo Wang, Liang Chen, Giacomo
  Tarroni, Wenjia Bai, and Daniel Rueckert.
\newblock Realistic adversarial data augmentation for mr image segmentation.
\newblock In \emph{Medical Image Computing and Computer Assisted Intervention
  {MICCAI}}, 2020{\natexlab{a}}.

\bibitem[Chen et~al.(2020{\natexlab{b}})Chen, Radford, Child, Wu, Jun, Luan,
  and Sutskever]{chen2020generative}
Mark Chen, Alec Radford, Rewon Child, Jeffrey Wu, Heewoo Jun, David Luan, and
  Ilya Sutskever.
\newblock Generative pretraining from pixels.
\newblock In \emph{International conference on machine learning}, pages
  1691--1703. PMLR, 2020{\natexlab{b}}.

\bibitem[Chen et~al.(2020{\natexlab{c}})Chen, Kornblith, Norouzi, and
  Hinton]{chen2020simple}
Ting Chen, Simon Kornblith, Mohammad Norouzi, and Geoffrey Hinton.
\newblock A simple framework for contrastive learning of visual
  representations.
\newblock In \emph{International conference on machine learning}, pages
  1597--1607. PMLR, 2020{\natexlab{c}}.

\bibitem[Chen et~al.(2020{\natexlab{d}})Chen, Kornblith, Swersky, Norouzi, and
  Hinton]{chen2020big}
Ting Chen, Simon Kornblith, Kevin Swersky, Mohammad Norouzi, and Geoffrey~E
  Hinton.
\newblock Big self-supervised models are strong semi-supervised learners.
\newblock \emph{Advances in neural information processing systems},
  33:\penalty0 22243--22255, 2020{\natexlab{d}}.

\bibitem[Chen and He(2021)]{chen2021exploring}
Xinlei Chen and Kaiming He.
\newblock Exploring simple siamese representation learning.
\newblock In \emph{Proceedings of the IEEE/CVF conference on computer vision
  and pattern recognition}, pages 15750--15758, 2021.

\bibitem[Chen et~al.(2020{\natexlab{e}})Chen, Fan, Girshick, and
  He]{chen2020improved}
Xinlei Chen, Haoqi Fan, Ross Girshick, and Kaiming He.
\newblock Improved baselines with momentum contrastive learning.
\newblock \emph{arXiv preprint arXiv:2003.04297}, 2020{\natexlab{e}}.

\bibitem[Chen et~al.(2022)Chen, Mancini, Zhu, and Akata]{chen2022semi}
Yanbei Chen, Massimiliano Mancini, Xiatian Zhu, and Zeynep Akata.
\newblock Semi-supervised and unsupervised deep visual learning: A survey.
\newblock \emph{IEEE Transactions on Pattern Analysis and Machine
  Intelligence}, 2022.

\bibitem[Cubuk et~al.(2020)Cubuk, Zoph, Shlens, and Le]{cubuk2020randaugment}
Ekin~D Cubuk, Barret Zoph, Jonathon Shlens, and Quoc~V Le.
\newblock Randaugment: Practical automated data augmentation with a reduced
  search space.
\newblock In \emph{Proceedings of the IEEE/CVF conference on computer vision
  and pattern recognition workshops}, pages 702--703, 2020.

\bibitem[Da~Costa et~al.(2022)Da~Costa, Fini, Nabi, Sebe, and
  Ricci]{da2022solo}
Victor Guilherme~Turrisi Da~Costa, Enrico Fini, Moin Nabi, Nicu Sebe, and Elisa
  Ricci.
\newblock solo-learn: A library of self-supervised methods for visual
  representation learning.
\newblock \emph{J. Mach. Learn. Res.}, 23\penalty0 (56):\penalty0 1--6, 2022.

\bibitem[{de Vente} et~al.(2023){de Vente}, Vermeer, Jaccard, Wang, Sun,
  Khader, Truhn, Aimyshev, Zhanibekuly, Le, Galdran, Ballester, Carneiro, G, S,
  Puthussery, Liu, Yang, Kondo, Kasai, Wang, Durvasula, Heras, Zapata,
  Ara{\'u}jo, Aresta, Bogunovi{\'c}, Arikan, Lee, Cho, Choi, Qayyum, Razzak,
  {van Ginneken}, Lemij, and
  S{\'a}nchez]{deventeAIROGSArtificialIntelligence2023}
Coen {de Vente}, Koenraad~A. Vermeer, Nicolas Jaccard, He Wang, Hongyi Sun,
  Firas Khader, Daniel Truhn, Temirgali Aimyshev, Yerkebulan Zhanibekuly,
  Tien-Dung Le, Adrian Galdran, Miguel {\'A}ngel~Gonz{\'a}lez Ballester,
  Gustavo Carneiro, Devika~R. G, Hrishikesh~P. S, Densen Puthussery, Hong Liu,
  Zekang Yang, Satoshi Kondo, Satoshi Kasai, Edward Wang, Ashritha Durvasula,
  J{\'o}nathan Heras, Miguel~{\'A}ngel Zapata, Teresa Ara{\'u}jo, Guilherme
  Aresta, Hrvoje Bogunovi{\'c}, Mustafa Arikan, Yeong~Chan Lee, Hyun~Bin Cho,
  Yoon~Ho Choi, Abdul Qayyum, Imran Razzak, Bram {van Ginneken}, Hans~G. Lemij,
  and Clara~I. S{\'a}nchez.
\newblock {{AIROGS}}: {{Artificial Intelligence}} for {{RObust Glaucoma
  Screening Challenge}}, 2023.

\bibitem[Devlin et~al.(2018)Devlin, Chang, Lee, and Toutanova]{devlin2018bert}
Jacob Devlin, Ming-Wei Chang, Kenton Lee, and Kristina Toutanova.
\newblock Bert: Pre-training of deep bidirectional transformers for language
  understanding.
\newblock \emph{arXiv preprint arXiv:1810.04805}, 2018.

\bibitem[Dosovitskiy et~al.(2020)Dosovitskiy, Beyer, Kolesnikov, Weissenborn,
  Zhai, Unterthiner, Dehghani, Minderer, Heigold, Gelly,
  et~al.]{dosovitskiy2020image}
Alexey Dosovitskiy, Lucas Beyer, Alexander Kolesnikov, Dirk Weissenborn,
  Xiaohua Zhai, Thomas Unterthiner, Mostafa Dehghani, Matthias Minderer, Georg
  Heigold, Sylvain Gelly, et~al.
\newblock An image is worth 16x16 words: Transformers for image recognition at
  scale.
\newblock \emph{arXiv preprint arXiv:2010.11929}, 2020.

\bibitem[Ericsson et~al.(2021)Ericsson, Gouk, and Hospedales]{ericsson2021well}
Linus Ericsson, Henry Gouk, and Timothy~M Hospedales.
\newblock How well do self-supervised models transfer?
\newblock In \emph{Proceedings of the IEEE/CVF Conference on Computer Vision
  and Pattern Recognition}, pages 5414--5423, 2021.

\bibitem[Esteva et~al.(2017)Esteva, Kuprel, Novoa, Ko, Swetter, Blau, and
  Thrun]{esteva2017dermatologist}
Andre Esteva, Brett Kuprel, Roberto~A Novoa, Justin Ko, Susan~M Swetter,
  Helen~M Blau, and Sebastian Thrun.
\newblock Dermatologist-level classification of skin cancer with deep neural
  networks.
\newblock \emph{nature}, 542\penalty0 (7639):\penalty0 115--118, 2017.

\bibitem[Gaur et~al.(2023)Gaur, Bhatia, Jhanjhi, Muhammad, and
  Masud]{gaur2023medical}
Loveleen Gaur, Ujwal Bhatia, NZ Jhanjhi, Ghulam Muhammad, and Mehedi Masud.
\newblock Medical image-based detection of covid-19 using deep convolution
  neural networks.
\newblock \emph{Multimedia systems}, 29\penalty0 (3):\penalty0 1729--1738,
  2023.

\bibitem[Ghayvat et~al.(2023)Ghayvat, Awais, Bashir, Pandya, Zuhair, Rashid,
  and Nebhen]{ghayvat2023ai}
Hemant Ghayvat, Muhammad Awais, AK Bashir, Sharnil Pandya, Mohd Zuhair, Mamoon
  Rashid, and Jamel Nebhen.
\newblock Ai-enabled radiologist in the loop: novel ai-based framework to
  augment radiologist performance for covid-19 chest ct medical image
  annotation and classification from pneumonia.
\newblock \emph{Neural Computing and Applications}, 35\penalty0 (20):\penalty0
  14591--14609, 2023.

\bibitem[Ghorbani et~al.(2020)Ghorbani, Ouyang, Abid, He, Chen, Harrington,
  Liang, Ashley, and Zou]{ghorbani2020deep}
Amirata Ghorbani, David Ouyang, Abubakar Abid, Bryan He, Jonathan~H Chen,
  Robert~A Harrington, David~H Liang, Euan~A Ashley, and James~Y Zou.
\newblock Deep learning interpretation of echocardiograms.
\newblock \emph{NPJ digital medicine}, 3\penalty0 (1):\penalty0 10, 2020.

\bibitem[Goyal et~al.(2019)Goyal, Mahajan, Gupta, and Misra]{goyal2019scaling}
Priya Goyal, Dhruv Mahajan, Abhinav Gupta, and Ishan Misra.
\newblock Scaling and benchmarking self-supervised visual representation
  learning.
\newblock In \emph{Proceedings of the ieee/cvf International Conference on
  computer vision}, pages 6391--6400, 2019.

\bibitem[Grandini et~al.(2020)Grandini, Bagli, and Visani]{grandini2020metrics}
Margherita Grandini, Enrico Bagli, and Giorgio Visani.
\newblock Metrics for multi-class classification: an overview.
\newblock \emph{arXiv preprint arXiv:2008.05756}, 2020.

\bibitem[Grill et~al.(2020)Grill, Strub, Altch{\'e}, Tallec, Richemond,
  Buchatskaya, Doersch, Avila~Pires, Guo, Gheshlaghi~Azar,
  et~al.]{grill2020bootstrap}
Jean-Bastien Grill, Florian Strub, Florent Altch{\'e}, Corentin Tallec, Pierre
  Richemond, Elena Buchatskaya, Carl Doersch, Bernardo Avila~Pires, Zhaohan
  Guo, Mohammad Gheshlaghi~Azar, et~al.
\newblock Bootstrap your own latent-a new approach to self-supervised learning.
\newblock \emph{Advances in neural information processing systems},
  33:\penalty0 21271--21284, 2020.

\bibitem[Guyon et~al.(2015)Guyon, Bennett, Cawley, Escalante, Escalera, Ho,
  Maci{\`a}, Ray, Saeed, Statnikov, et~al.]{guyon2015design}
Isabelle Guyon, Kristin Bennett, Gavin Cawley, Hugo~Jair Escalante, Sergio
  Escalera, Tin~Kam Ho, N{\'u}ria Maci{\`a}, Bisakha Ray, Mehreen Saeed,
  Alexander Statnikov, et~al.
\newblock Design of the 2015 {C}ha{L}earn {AutoML} challenge.
\newblock In \emph{2015 International Joint Conference on Neural Networks
  (IJCNN)}, pages 1--8. IEEE, 2015.

\bibitem[He et~al.(2016)He, Zhang, Ren, and Sun]{he2016deep}
Kaiming He, Xiangyu Zhang, Shaoqing Ren, and Jian Sun.
\newblock Deep residual learning for image recognition.
\newblock In \emph{Proceedings of the IEEE conference on computer vision and
  pattern recognition}, pages 770--778, 2016.

\bibitem[He et~al.(2020)He, Fan, Wu, Xie, and Girshick]{he2020momentum}
Kaiming He, Haoqi Fan, Yuxin Wu, Saining Xie, and Ross Girshick.
\newblock Momentum contrast for unsupervised visual representation learning.
\newblock In \emph{Proceedings of the IEEE/CVF conference on computer vision
  and pattern recognition}, pages 9729--9738, 2020.

\bibitem[Hoeffding(1994)]{hoeffding1994probability}
Wassily Hoeffding.
\newblock Probability inequalities for sums of bounded random variables.
\newblock \emph{The collected works of Wassily Hoeffding}, pages 409--426,
  1994.

\bibitem[Huang et~al.(2021)Huang, Long, Wessler, and Hughes]{huang2021new}
Zhe Huang, Gary Long, Benjamin Wessler, and Michael~C Hughes.
\newblock A new semi-supervised learning benchmark for classifying view and
  diagnosing aortic stenosis from echocardiograms.
\newblock In \emph{Proceedings of the Machine Learning for Healthcare
  Conference}. PMLR, 2021.

\bibitem[Huang et~al.(2022)Huang, Long, Wessler, and Hughes]{huang2022tmed}
Zhe Huang, Gary Long, Benjamin~S Wessler, and Michael~C Hughes.
\newblock {TMED} 2: A dataset for semi-supervised classification of
  echocardiograms.
\newblock In \emph{DataPerf: Benchmarking Data for Data-Centric AI Workshop},
  2022.

\bibitem[Huang et~al.(2023{\natexlab{a}})Huang, Sidhom, Wessler, and
  Hughes]{huang2022fix}
Zhe Huang, Mary-Joy Sidhom, Benjamin~S Wessler, and Michael~C Hughes.
\newblock Fix-a-step: Semi-supervised learning from uncurated unlabeled data.
\newblock In \emph{Proceedings of The 26th International Conference on
  Artificial Intelligence and Statistics (AISTATS)}, 2023{\natexlab{a}}.

\bibitem[Huang et~al.(2023{\natexlab{b}})Huang, Wessler, and
  Hughes]{huang2023detecting}
Zhe Huang, Benjamin~S Wessler, and Michael~C Hughes.
\newblock Detecting heart disease from multi-view ultrasound images via
  supervised attention multiple instance learning.
\newblock In \emph{Machine Learning for Healthcare Conference}, pages 285--307.
  PMLR, 2023{\natexlab{b}}.

\bibitem[Huang et~al.(2024{\natexlab{a}})Huang, Yu, Wessler, and
  Hughes]{huang2024semi}
Zhe Huang, Xiaowei Yu, Benjamin~S Wessler, and Michael~C Hughes.
\newblock Semi-supervised multimodal multi-instance learning for aortic
  stenosis diagnosis.
\newblock \emph{arXiv preprint arXiv:2403.06024}, 2024{\natexlab{a}}.

\bibitem[Huang et~al.(2024{\natexlab{b}})Huang, Yu, Zhu, and
  Hughes]{huang2024interlude}
Zhe Huang, Xiaowei Yu, Dajiang Zhu, and Michael~C Hughes.
\newblock Interlude: Interactions between labeled and unlabeled data to enhance
  semi-supervised learning.
\newblock \emph{arXiv preprint arXiv:2403.10658}, 2024{\natexlab{b}}.

\bibitem[Huo et~al.(2023)Huo, Xie, Fang, Wang, Liu, Duan, Zhang, Wang, Xue,
  Liu, et~al.]{huo2023deep}
Tongtong Huo, Yi Xie, Ying Fang, Ziyi Wang, Pengran Liu, Yuyu Duan, Jiayao
  Zhang, Honglin Wang, Mingdi Xue, Songxiang Liu, et~al.
\newblock Deep learning-based algorithm improves radiologists’ performance in
  lung cancer bone metastases detection on computed tomography.
\newblock \emph{Frontiers in Oncology}, 13:\penalty0 1125637, 2023.

\bibitem[Iscen et~al.(2019)Iscen, Tolias, Avrithis, and Chum]{iscen2019label}
Ahmet Iscen, Giorgos Tolias, Yannis Avrithis, and Ondrej Chum.
\newblock Label propagation for deep semi-supervised learning.
\newblock In \emph{Proceedings of the IEEE/CVF conference on computer vision
  and pattern recognition}, pages 5070--5079, 2019.

\bibitem[Jing and Tian(2020)]{jing2020self}
Longlong Jing and Yingli Tian.
\newblock Self-supervised visual feature learning with deep neural networks: A
  survey.
\newblock \emph{IEEE transactions on pattern analysis and machine
  intelligence}, 43\penalty0 (11):\penalty0 4037--4058, 2020.

\bibitem[Kather et~al.(2019)Kather, Krisam, Charoentong, Luedde, Herpel, Weis,
  Gaiser, Marx, Valous, Ferber, et~al.]{kather2019predicting}
Jakob~Nikolas Kather, Johannes Krisam, Pornpimol Charoentong, Tom Luedde,
  Esther Herpel, Cleo-Aron Weis, Timo Gaiser, Alexander Marx, Nektarios~A
  Valous, Dyke Ferber, et~al.
\newblock Predicting survival from colorectal cancer histology slides using
  deep learning: A retrospective multicenter study.
\newblock \emph{PLoS medicine}, 16\penalty0 (1):\penalty0 e1002730, 2019.

\bibitem[Khosla et~al.(2020)Khosla, Teterwak, Wang, Sarna, Tian, Isola,
  Maschinot, Liu, and Krishnan]{khosla2020supervised}
Prannay Khosla, Piotr Teterwak, Chen Wang, Aaron Sarna, Yonglong Tian, Phillip
  Isola, Aaron Maschinot, Ce Liu, and Dilip Krishnan.
\newblock Supervised contrastive learning.
\newblock \emph{Advances in neural information processing systems},
  33:\penalty0 18661--18673, 2020.

\bibitem[Kim et~al.(2021)Kim, Choo, Kwon, Joe, Min, and Gwon]{kim2021selfmatch}
Byoungjip Kim, Jinho Choo, Yeong-Dae Kwon, Seongho Joe, Seungjai Min, and
  Youngjune Gwon.
\newblock Selfmatch: Combining contrastive self-supervision and consistency for
  semi-supervised learning.
\newblock \emph{arXiv preprint arXiv:2101.06480}, 2021.

\bibitem[Kingma and Ba(2014)]{kingma2014adam}
Diederik~P Kingma and Jimmy Ba.
\newblock Adam: A method for stochastic optimization.
\newblock \emph{arXiv preprint arXiv:1412.6980}, 2014.

\bibitem[Kingma et~al.(2014)Kingma, Mohamed, Jimenez~Rezende, and
  Welling]{kingma2014semi}
Durk~P Kingma, Shakir Mohamed, Danilo Jimenez~Rezende, and Max Welling.
\newblock Semi-supervised learning with deep generative models.
\newblock \emph{Advances in neural information processing systems}, 27, 2014.

\bibitem[Kumar et~al.(2017)Kumar, Sattigeri, and Fletcher]{kumar2017semi}
Abhishek Kumar, Prasanna Sattigeri, and Tom Fletcher.
\newblock Semi-supervised learning with gans: Manifold invariance with improved
  inference.
\newblock \emph{Advances in neural information processing systems}, 30, 2017.

\bibitem[Kushnure et~al.(2023)Kushnure, Tyagi, and Talbar]{kushnure2023lim}
Devidas~T Kushnure, Shweta Tyagi, and Sanjay~N Talbar.
\newblock Lim-net: Lightweight multi-level multiscale network with deep
  residual learning for automatic liver segmentation in ct images.
\newblock \emph{Biomedical Signal Processing and Control}, 80:\penalty0 104305,
  2023.

\bibitem[Lai et~al.(2021)Lai, Wang, Oliveira, Dugger, Cheung, and
  Chuah]{lai2021joint}
Zhengfeng Lai, Chao Wang, Luca~Cerny Oliveira, Brittany~N Dugger, Sen-Ching
  Cheung, and Chen-Nee Chuah.
\newblock Joint semi-supervised and active learning for segmentation of
  gigapixel pathology images with cost-effective labeling.
\newblock In \emph{Proceedings of the IEEE/CVF International Conference on
  Computer Vision}, pages 591--600, 2021.

\bibitem[Lai et~al.(2022)Lai, Wang, Gunawan, Cheung, and
  Chuah]{lai2022smoothed}
Zhengfeng Lai, Chao Wang, Henrry Gunawan, Sen-Ching~S Cheung, and Chen-Nee
  Chuah.
\newblock Smoothed adaptive weighting for imbalanced semi-supervised learning:
  Improve reliability against unknown distribution data.
\newblock In \emph{International Conference on Machine Learning}, pages
  11828--11843. PMLR, 2022.

\bibitem[Laine and Aila(2016)]{laine2016temporal}
Samuli Laine and Timo Aila.
\newblock Temporal ensembling for semi-supervised learning.
\newblock \emph{arXiv preprint arXiv:1610.02242}, 2016.

\bibitem[Lee(2013)]{lee2013pseudo}
Dong-Hyun Lee.
\newblock Pseudo-label: The simple and efficient semi-supervised learning
  method for deep neural networks.
\newblock In \emph{Workshop on challenges in representation learning at ICML},
  2013.

\bibitem[Li et~al.(2021)Li, Xiong, and Hoi]{li2021comatch}
Junnan Li, Caiming Xiong, and Steven~CH Hoi.
\newblock Comatch: Semi-supervised learning with contrastive graph
  regularization.
\newblock In \emph{Proceedings of the IEEE/CVF international conference on
  computer vision}, pages 9475--9484, 2021.

\bibitem[Madani et~al.(2018)Madani, Arnaout, Mofrad, and
  Arnaout]{madani2018fast}
Ali Madani, Ramy Arnaout, Mohammad Mofrad, and Rima Arnaout.
\newblock Fast and accurate view classification of echocardiograms using deep
  learning.
\newblock \emph{NPJ digital medicine}, 1\penalty0 (1):\penalty0 6, 2018.

\bibitem[Min et~al.(2020)Min, Chen, Xie, Zha, and Zhang]{min2020mutually}
Shaobo Min, Xuejin Chen, Hongtao Xie, Zheng-Jun Zha, and Yongdong Zhang.
\newblock A mutually attentive co-training framework for semi-supervised
  recognition.
\newblock \emph{IEEE Transactions on Multimedia}, 23:\penalty0 899--910, 2020.

\bibitem[Oliver et~al.(2018)Oliver, Odena, Raffel, Cubuk, and
  Goodfellow]{oliver2018realistic}
Avital Oliver, Augustus Odena, Colin~A Raffel, Ekin~Dogus Cubuk, and Ian
  Goodfellow.
\newblock Realistic evaluation of deep semi-supervised learning algorithms.
\newblock \emph{Advances in neural information processing systems}, 31, 2018.

\bibitem[Oord et~al.(2018)Oord, Li, and Vinyals]{oord2018representation}
Aaron van~den Oord, Yazhe Li, and Oriol Vinyals.
\newblock Representation learning with contrastive predictive coding.
\newblock \emph{arXiv preprint arXiv:1807.03748}, 2018.

\bibitem[Pedregosa et~al.(2011)Pedregosa, Varoquaux, Gramfort, Michel, Thirion,
  Grisel, Blondel, Prettenhofer, Weiss, Dubourg, et~al.]{pedregosa2011scikit}
Fabian Pedregosa, Ga{\"e}l Varoquaux, Alexandre Gramfort, Vincent Michel,
  Bertrand Thirion, Olivier Grisel, Mathieu Blondel, Peter Prettenhofer, Ron
  Weiss, Vincent Dubourg, et~al.
\newblock Scikit-learn: Machine learning in python.
\newblock \emph{the Journal of machine Learning research}, 12:\penalty0
  2825--2830, 2011.

\bibitem[Qi and Luo(2020)]{qi2020small}
Guo-Jun Qi and Jiebo Luo.
\newblock Small data challenges in big data era: A survey of recent progress on
  unsupervised and semi-supervised methods.
\newblock \emph{IEEE Transactions on Pattern Analysis and Machine
  Intelligence}, 44\penalty0 (4):\penalty0 2168--2187, 2020.

\bibitem[Raghu et~al.(2019)Raghu, Zhang, Kleinberg, and
  Bengio]{raghu2019transfusion}
Maithra Raghu, Chiyuan Zhang, Jon Kleinberg, and Samy Bengio.
\newblock Transfusion: Understanding transfer learning for medical imaging.
\newblock \emph{Advances in neural information processing systems}, 32, 2019.

\bibitem[Saito et~al.(2021)Saito, Kim, and Saenko]{saito2021openmatch}
Kuniaki Saito, Donghyun Kim, and Kate Saenko.
\newblock Openmatch: Open-set consistency regularization for semi-supervised
  learning with outliers.
\newblock \emph{arXiv preprint arXiv:2105.14148}, 2021.

\bibitem[Shekoofeh et~al.(2021)Shekoofeh, Basil, Fiona, Zachary, Jan, Jonathan,
  Aaron, Alan, Simon, Ting, et~al.]{shekoofeh2021big}
Azizi Shekoofeh, Mustafa Basil, Ryan Fiona, Beaver Zachary, Freyberg Jan,
  Deaton Jonathan, Loh Aaron, Karthikesalingam Alan, Kornblith Simon, Chen
  Ting, et~al.
\newblock Big self-supervised models advance medical image classification.
\newblock \emph{arXiv preprint arXiv:2101.05224}, 2021.

\bibitem[Sohn et~al.(2020)Sohn, Berthelot, Carlini, Zhang, Zhang, Raffel,
  Cubuk, Kurakin, and Li]{sohn2020fixmatch}
Kihyuk Sohn, David Berthelot, Nicholas Carlini, Zizhao Zhang, Han Zhang,
  Colin~A Raffel, Ekin~Dogus Cubuk, Alexey Kurakin, and Chun-Liang Li.
\newblock Fixmatch: Simplifying semi-supervised learning with consistency and
  confidence.
\newblock \emph{Advances in neural information processing systems},
  33:\penalty0 596--608, 2020.

\bibitem[Steyerberg and Vergouwe(2014)]{steyerberg2014towards}
Ewout~W Steyerberg and Yvonne Vergouwe.
\newblock Towards better clinical prediction models: seven steps for
  development and an abcd for validation.
\newblock \emph{European Heart Journal}, 35\penalty0 (29), 2014.

\bibitem[Su et~al.(2021)Su, Cheng, and Maji]{su2021realistic}
Jong-Chyi Su, Zezhou Cheng, and Subhransu Maji.
\newblock A realistic evaluation of semi-supervised learning for fine-grained
  classification.
\newblock In \emph{Proceedings of the IEEE/CVF Conference on Computer Vision
  and Pattern Recognition}, pages 12966--12975, 2021.

\bibitem[Suzuki(2020)]{suzuki2020consistency}
Teppei Suzuki.
\newblock Consistency regularization for semi-supervised learning with pytorch.
\newblock \url{https://github.com/perrying/pytorch-consistency-regularization},
  2020.

\bibitem[Tarvainen and Valpola(2017)]{tarvainen2017mean}
Antti Tarvainen and Harri Valpola.
\newblock Mean teachers are better role models: Weight-averaged consistency
  targets improve semi-supervised deep learning results.
\newblock \emph{Advances in neural information processing systems}, 30, 2017.

\bibitem[Van~Engelen and Hoos(2020)]{van2020survey}
Jesper~E Van~Engelen and Holger~H Hoos.
\newblock A survey on semi-supervised learning.
\newblock \emph{Machine learning}, 109\penalty0 (2):\penalty0 373--440, 2020.

\bibitem[Wagner et~al.(2022)Wagner, Ferreira, Stoll, Schirrmeister, M{\"u}ller,
  and Hutter]{wagner2022importance}
Diane Wagner, Fabio Ferreira, Danny Stoll, Robin~Tibor Schirrmeister, Samuel
  M{\"u}ller, and Frank Hutter.
\newblock On the importance of hyperparameters and data augmentation for
  self-supervised learning.
\newblock \emph{arXiv preprint arXiv:2207.07875}, 2022.

\bibitem[Wang et~al.(2022)Wang, Chen, Fan, Sun, Tao, Hou, Wang, Yang, Zhou,
  Guo, et~al.]{wang2022usb}
Yidong Wang, Hao Chen, Yue Fan, Wang Sun, Ran Tao, Wenxin Hou, Renjie Wang,
  Linyi Yang, Zhi Zhou, Lan-Zhe Guo, et~al.
\newblock Usb: A unified semi-supervised learning benchmark for classification.
\newblock \emph{Advances in Neural Information Processing Systems},
  35:\penalty0 3938--3961, 2022.

\bibitem[Wessler et~al.(2023)Wessler, Huang, Long~Jr, Pacifici, Prashar,
  Karmiy, Sandler, Sokol, Sokol, Dehn, et~al.]{wessler2023automated}
Benjamin~S Wessler, Zhe Huang, Gary~M Long~Jr, Stefano Pacifici, Nishant
  Prashar, Samuel Karmiy, Roman~A Sandler, Joseph~Z Sokol, Daniel~B Sokol,
  Monica~M Dehn, et~al.
\newblock Automated detection of aortic stenosis using machine learning.
\newblock \emph{Journal of the American Society of Echocardiography},
  36\penalty0 (4):\penalty0 411--420, 2023.

\bibitem[Wu et~al.(2019)Wu, Phang, Park, Shen, Huang, Zorin, Jastrz{\k{e}}bski,
  F{\'e}vry, Katsnelson, Kim, et~al.]{wu2019deep}
Nan Wu, Jason Phang, Jungkyu Park, Yiqiu Shen, Zhe Huang, Masha Zorin,
  Stanis{\l}aw Jastrz{\k{e}}bski, Thibault F{\'e}vry, Joe Katsnelson, Eric Kim,
  et~al.
\newblock Deep neural networks improve radiologists’ performance in breast
  cancer screening.
\newblock \emph{IEEE transactions on medical imaging}, 39\penalty0
  (4):\penalty0 1184--1194, 2019.

\bibitem[Yang et~al.(2021)Yang, Shi, and Ni]{yang2021medmnist}
Jiancheng Yang, Rui Shi, and Bingbing Ni.
\newblock Medmnist classification decathlon: A lightweight automl benchmark for
  medical image analysis.
\newblock In \emph{2021 IEEE 18th International Symposium on Biomedical Imaging
  (ISBI)}, pages 191--195. IEEE, 2021.

\bibitem[Yang et~al.(2023)Yang, Shi, Wei, Liu, Zhao, Ke, Pfister, and
  Ni]{yang2023medmnist}
Jiancheng Yang, Rui Shi, Donglai Wei, Zequan Liu, Lin Zhao, Bilian Ke,
  Hanspeter Pfister, and Bingbing Ni.
\newblock Medmnist v2-a large-scale lightweight benchmark for 2d and 3d
  biomedical image classification.
\newblock \emph{Scientific Data}, 10\penalty0 (1):\penalty0 41, 2023.

\bibitem[Zagoruyko and Komodakis(2016)]{zagoruyko2016wide}
Sergey Zagoruyko and Nikos Komodakis.
\newblock Wide residual networks.
\newblock In \emph{Proceedings of the British Machine Vision Conference
  (BMVC)}, 2016.

\bibitem[Zbontar et~al.(2021)Zbontar, Jing, Misra, LeCun, and
  Deny]{zbontar2021barlow}
Jure Zbontar, Li Jing, Ishan Misra, Yann LeCun, and St{\'e}phane Deny.
\newblock Barlow twins: Self-supervised learning via redundancy reduction.
\newblock In \emph{International Conference on Machine Learning}, pages
  12310--12320. PMLR, 2021.

\bibitem[Zhai et~al.(2019)Zhai, Oliver, Kolesnikov, and Beyer]{zhai2019s4l}
Xiaohua Zhai, Avital Oliver, Alexander Kolesnikov, and Lucas Beyer.
\newblock S4l: Self-supervised semi-supervised learning.
\newblock In \emph{Proceedings of the IEEE/CVF international conference on
  computer vision}, pages 1476--1485, 2019.

\bibitem[Zhang et~al.(2021)Zhang, Wang, Hou, Wu, Wang, Okumura, and
  Shinozaki]{zhang2021flexmatch}
Bowen Zhang, Yidong Wang, Wenxin Hou, Hao Wu, Jindong Wang, Manabu Okumura, and
  Takahiro Shinozaki.
\newblock Flexmatch: Boosting semi-supervised learning with curriculum pseudo
  labeling.
\newblock \emph{Advances in Neural Information Processing Systems},
  34:\penalty0 18408--18419, 2021.

\bibitem[Zhang et~al.(2017)Zhang, Cisse, Dauphin, and
  Lopez-Paz]{zhang2017mixup}
Hongyi Zhang, Moustapha Cisse, Yann~N Dauphin, and David Lopez-Paz.
\newblock mixup: Beyond empirical risk minimization.
\newblock \emph{arXiv preprint arXiv:1710.09412}, 2017.

\bibitem[Zhang et~al.(2022)Zhang, Zhu, Hallinan, Zhang, Makmur, Cai, and
  Ooi]{zhang2022boostmis}
Wenqiao Zhang, Lei Zhu, James Hallinan, Shengyu Zhang, Andrew Makmur, Qingpeng
  Cai, and Beng~Chin Ooi.
\newblock Boost{MIS}: Boosting medical image semi-supervised learning with
  adaptive pseudo labeling and informative active annotation.
\newblock In \emph{Proceedings of the IEEE/CVF {C}onference on {C}omputer
  {V}ision and {P}attern {R}ecognition ({CVPR})}, 2022.

\bibitem[Zheng et~al.(2022)Zheng, You, Huang, Wang, Qian, and
  Xu]{zheng2022simmatch}
Mingkai Zheng, Shan You, Lang Huang, Fei Wang, Chen Qian, and Chang Xu.
\newblock Simmatch: Semi-supervised learning with similarity matching.
\newblock In \emph{Proceedings of the IEEE/CVF Conference on Computer Vision
  and Pattern Recognition}, pages 14471--14481, 2022.

\bibitem[Zhu(2005)]{zhu2005semi}
Xiaojin Zhu.
\newblock Semi-{{Supervised Learning Literature Survey}}.
\newblock Technical Report 1530, {Department of Computer Science, University of
  Wisconsin Madison.}, 2005.

\bibitem[Zhu et~al.(2003)Zhu, Ghahramani, and Lafferty]{zhu2003semi}
Xiaojin Zhu, Zoubin Ghahramani, and John~D Lafferty.
\newblock Semi-supervised learning using gaussian fields and harmonic
  functions.
\newblock In \emph{Proceedings of the 20th International conference on Machine
  learning (ICML-03)}, pages 912--919, 2003.

\bibitem[Zhuang et~al.(2020)Zhuang, Qi, Duan, Xi, Zhu, Zhu, Xiong, and
  He]{zhuang2020comprehensive}
Fuzhen Zhuang, Zhiyuan Qi, Keyu Duan, Dongbo Xi, Yongchun Zhu, Hengshu Zhu, Hui
  Xiong, and Qing He.
\newblock A comprehensive survey on transfer learning.
\newblock \emph{Proceedings of the IEEE}, 109\penalty0 (1):\penalty0 43--76,
  2020.

\end{thebibliography}
    %\bibliography{refs_manual}
    %\bibliography{refs_from_zotero}
}


\clearpage
\appendix

\onecolumn
%% Config Table-of-Contents to track the sections of the appendix
\startcontents[sections]
\counterwithin{table}{section}
\setcounter{table}{0}
\counterwithin{figure}{section}
\setcounter{figure}{0}
\counterwithin{algorithm}{section}
\setcounter{algorithm}{0}
%% Use ONE counter for all figs and tables to give unique identifiers in supplement
\makeatletter 
\let\c@table\c@figure
\let\c@lstlisting\c@figure
\let\c@algorithm\c@figure
\makeatother

\section*{Acknowledgments}
RJ and SA acknowledge support from the U.S. National Science Foundation under award \# 1931978.
All authors are grateful for computing infrastructure support from the Tufts High-Performance Computing cluster, partially funded by NSF under grant OAC CC* 2018149.

% Print Table of Contents
\section*{Appendix Contents} 
\printcontents[sections]{l}{1}{\setcounter{tocdepth}{2}}

%There are several things I'd like to plan to revise for next revision
%(i) App C (hyperparameters) needs to provide the concrete settings we recommend for each method, in addition to the ranges we searched over
%(ii) App B could have a bit more formal description of each method's definition of unlabeled loss.... it's a bit loose now
%(iii) we should have a code-review where we compare the actual code to our Alg B.1, to be sure we're actually doing what we say
%(iv) provide better justification why batch sizes and a few other settings vary a lot between methods

\section{Code and Data Resources for Reproducibility}
\label{app:code}

All code and data resources needed to reproduce our analysis, including information on exact splits we used for each of the 4 datasets ( TissueMNIST, PathMNIST, TMED-2, AIROGS) can be found in our github repo:

\url{https://github.com/tufts-ml/SSL-vs-SSL-benchmark}
% \url{https://anonymous.4open.science/r/SSL-vs-SSL-benchmark-48B0/README.md}

% The exact splits of the TissueMNIST, PathMNIST and AIROGS we used are available in this shareable folder: \url{https://tufts.box.com/s/wxmd8a1gbkmau57n05vquu3e5sqbpgls}. 

% For TMED2, we used the first split of the publicly available data, documented here: \url{https://tmed.cs.tufts.edu/tmed_v2.html}

\paragraph{Primer on our codebase.}
Our codebase builds upon the open-source PyTorch repo by~\citet{suzuki2020consistency}. 
\citeauthor{suzuki2020consistency}'s code was originally intended as a reimplementation in PyTorch of \citet{oliver2018realistic}'s benchmark of semi-supervised learning (while Oliver et al’s original repo was in Tensorflow, we prefer PyTorch).

We added many additional algorithms (we added MixMatch, FixMatch, FlexMatch, and CoMatch, as well as all 7 self-supervised methods) and customized the experiments, especially providing a runtime-budgeted hyperparameter tuning strategy as outlined in App.~\ref{App:Algorithms}.

In a way, this makes our repo a ``cousin'' of the codebase of \citet{su2021realistic}’s fine-grained classification benchmark, because their \href{https://github.com/cvl-umass/ssl-evaluation}{github repo} also credits Suzuki’s repo as an ancestor.

\clearpage

\section{Dataset Details}
\subsection{Example Images}
Below we show a few examples for each dataset. For full details, please refer to the original papers.

\setlength{\tabcolsep}{0.1cm}
% \begin{figure}[!h]
\begin{figure}[H]
\begin{tabular}{r c c c c c}
    \\
    {\rotatebox{90}{~~~~~~PathMNIST}}
   & 
    {\includegraphics[width=0.18\textwidth]{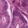}}
    &
    {\includegraphics[width=0.18\textwidth]{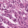}}
    &
    {\includegraphics[width=0.18\textwidth]{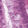}}
    &
    {\includegraphics[width=0.18\textwidth]{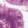}}
    &
    {\includegraphics[width=0.18\textwidth]{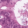}}
    
    \\
    {\rotatebox{90}{~~~~~~TissueMNIST}}
     & 
    {\includegraphics[width=0.18\textwidth]{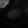}}
    &
    {\includegraphics[width=0.18\textwidth]{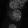}}
    &
    {\includegraphics[width=0.18\textwidth]{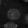}}
    &
    {\includegraphics[width=0.18\textwidth]{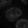}}
    &
    {\includegraphics[width=0.18\textwidth]{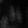}}
    
    \\
    {\rotatebox{90}{~~~~~~~~~~~TMED2}}
      & 
    {\includegraphics[width=0.18\textwidth]{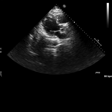}}
    &
    {\includegraphics[width=0.18\textwidth]{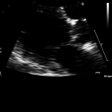}}
    &
    {\includegraphics[width=0.18\textwidth]{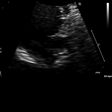}}
    &
    {\includegraphics[width=0.18\textwidth]{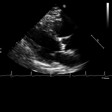}}
    &
    {\includegraphics[width=0.18\textwidth]{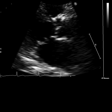}}
    
    \\
    {\rotatebox{90}{~~~~~~~~~AIROGS}}
     & 
    {\includegraphics[width=0.18\textwidth]{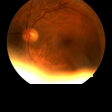}}
    &
    {\includegraphics[width=0.18\textwidth]{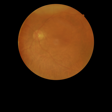}}
    &
    {\includegraphics[width=0.18\textwidth]{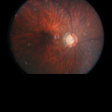}}
    &
    {\includegraphics[width=0.18\textwidth]{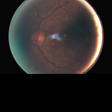}}
    &
    {\includegraphics[width=0.18\textwidth]{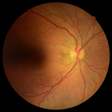}}
    
    \end{tabular}	
    \caption{Showing 5 random examples for each dataset.
     }
    \label{fig:top_attended_images}
\end{figure}

\subsection{Dataset Selection}
\label{app:dataset_selection}

We selected PathMNIST and TissueMNIST from 12 candidate datasets in the MedMNIST collections~\citep{yang2021medmnist,yang2023medmnist} by matching two criteria: (i) contains at least 5 imbalanced classes; (ii) can build a large unlabeled set (at least 50000 images). 
% \revision{It may be tempting to worry that 28x28 resolution is too small; however experiments by \citet{yang2023medmnist} suggest that on both Path and Tissue datasets that diagnostic accuracy given \emph{all labeled data} at 28x28 is quite good (AUROC above 0.93); using finer 224x224 images yields little improvement (${<}0.01$ gain in AUROC).}
Prior experiments from dataset creator \citet{yang2023medmnist} suggest 28x28 resolution is a reasonable choice. They report that a larger resolution (224x224) does not yield much more accurate classifiers for these two datasets.

\subsection{Classification Task Description}
\label{app:class_description}

\emph{TissueMNIST} contains 28x28 images of human kidney cortex cells. The dataset contains 8 classes. See ~\cite{yang2023medmnist} for details.

\begin{table}[H]
\begin{tabular}{c l l}
\toprule
Class ID & Abbreviation & Description \\
\midrule
0 & CD/CT & Collecting Duct, Connecting Tubule \\
1 & DCT & Distal Convoluted Tubule \\
2 & GE & Glomerular endothelial cells \\
3 & IE & Interstitial endothelial cells \\
4 & LEU & Leukocytes \\
5 & POD & Podocytes \\
6 & PT & Proximal Tubule Segments \\
7 & TAL & Thick Ascending Limb \\
\bottomrule
\end{tabular}
\end{table}

\noindent \emph{PathMNIST} contains 28x28 patches from colorectal cancer histology slides that comprise 9 tissue types. See~\cite{yang2023medmnist,kather2019predicting} for details.

\begin{table}[H]
\begin{tabular}{c l l}
\toprule
Class ID & Abbreviation & Description \\
\midrule
0 & ADI & adipose \\
1 & BACK & background \\
2 & DEB & debris \\
3 & LYM & lymphocytes \\
4 & MUC & mucus \\
5 & MUS & smooth muscle \\
6 & NORM & normal colon mucosa \\
7 & STR & cancer-associated stroma \\
8 & TUM & colorectal adenocarcinoma epithelium \\
\bottomrule
\end{tabular}
\end{table}

\noindent \emph{TMED-2} contains 112x112 2D grayscale images captured from routine echocardiogram scans (ultrasound images of the heart). In this study, we adopt the view classification task from ~\cite{huang2021new}. For more detail please see~\cite{huang2021new,huang2022tmed}

\begin{table}[H]
\begin{tabular}{c l l}
\toprule
Class ID & Abbreviation & Description \\
\midrule
0 & PLAX & parasternal long axis \\
1 & PSAX & parasternal short axis \\
2 & A2C & apical 2-chamber \\
3 & A4C & apical 4-chamber \\

\bottomrule
\end{tabular}
\end{table}

\noindent \emph{AIROGS} is a dataset of color fundus photographs of the retina. The binary classification task is to detect evidence of referable glaucoma~\citep{deventeAIROGSArtificialIntelligence2023}. We use 384x384 resolution, as suggested by several challenge participants.

\begin{table}[H]
\begin{tabular}{c l l}
\toprule
Class ID & Abbreviation & Description \\
\midrule
0 & No Glauc. & no referable glaucoma \\
1 & Glaucoma & referable glaucoma {\footnotesize (signs associated with
visual field defects on standard automated perimetry)} \\

\bottomrule
\end{tabular}
\end{table}

\clearpage

\section{Additional Results} 
\label{app:results}

\subsection{Impact of pretraining on accuracy-over-time profiles}

\begin{figure}[!h]
    \begin{subfigure}[b]{.48\textwidth}
         \includegraphics[width=.9\textwidth]{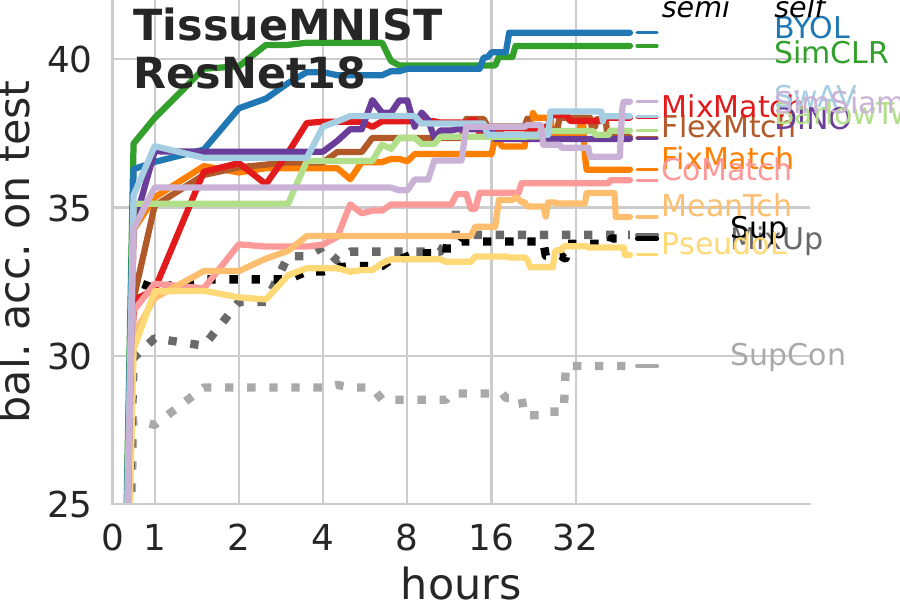}
        \caption{TissueMNIST : pretrained initial weights}
        % \label{fig:tissuemnist_performance_vs_time_pretrain}
    \end{subfigure}
\hspace{2mm}
    \begin{subfigure}[b]{.48\textwidth}
          \includegraphics[width=.9\textwidth]{CVPR_figures/mch/AccVsTime_Tissue.pdf}
        \caption{TissueMNIST : random initial weights}
        % \label{fig:tissuemnist_performance_vs_time_random}
    \end{subfigure}
\\
    \begin{subfigure}[b]{.48\textwidth}
        \includegraphics[width=.9\textwidth]{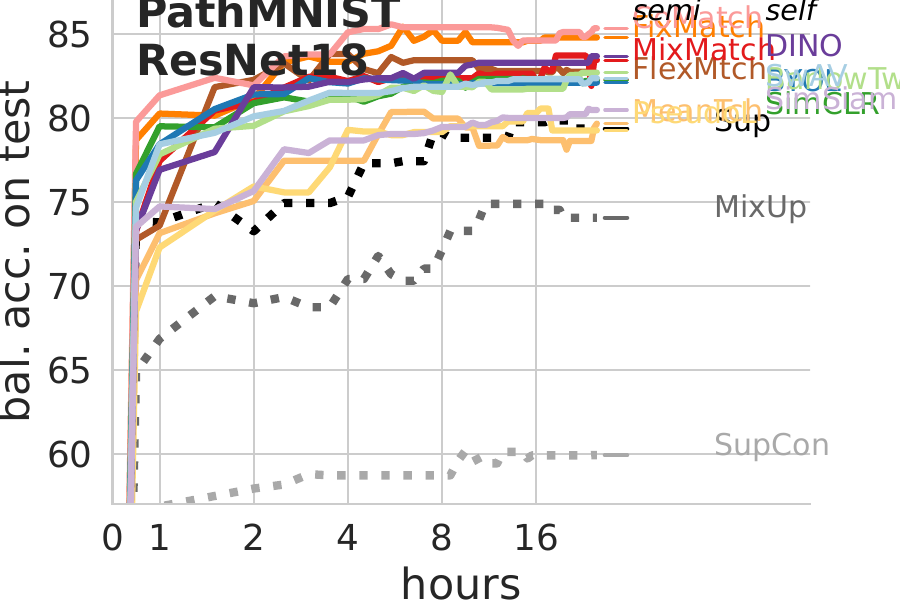}
        \caption{PathMNIST : pretrained initial weights}
        % \label{fig:path_performance_vs_time_pretrain}
    \end{subfigure}
\hspace{2mm}
    \begin{subfigure}[b]{.48\textwidth}
        \includegraphics[width=.9\textwidth]{CVPR_figures/mch/AccVsTime_Path.pdf}
        \caption{PathMNIST: random initial weights}
        % \label{fig:path_performance_vs_time_random}
    \end{subfigure}
\caption{
Balanced accuracy on test set over time for semi- and self-supervised methods, \textbf{with (left) and without (right) initial weight pretraining on ImageNet.}
Curves represent mean of each method at each time over 5 trials of Alg. ~\ref{alg:hyperparam_tuning}.}%endcaption
\label{fig:test_performance_vs_time_pretrain_vs_fromscratch}
\end{figure}

To study the impact of pretraining, we compare the accuracy-over-time profiles of TissueMNIST and PathMNIST based on the two different initialization strategy. Fig.~\ref{fig:test_performance_vs_time_pretrain_vs_fromscratch} shows balanced-accuracy-over-time profiles for initialization of neural net parameters to values pretrained on ImageNet (left column) and random initialization (right column).  Pretraining time on a source dataset is NOT counted to the runtime reported in x-axis.

On TissueMNIST (top row), SimCLR (green) and BYOL (blue) are the top two methods for both initialization types. Performance gains from pretraining are slight, BA for BYOL is around 42 with pretraining and 40 with random initialization.

On PathMNIST (bottom row), FixMatch and CoMatch are best in the pretraining case, with MixMatch and Flexmatch only a few points of balanced accuracy lower. MixMatch and CoMatch are best in the random initialization case.

Across both datasets, pretraining does not seem to impact the \textbf{top-performing methods}' ultimate accuracy by much, usually just a slight increase in BA of 0.5-3 points. 
One exception is FixMatch on PathMNIST, which improves by about 5 percentage points. We do not see the 10+ point gains reported by \citet{su2021realistic} in their Table 3.

Considering more limited time budgets (e.g. after only a few hours), we do see initialization from pretraining understandably tends to improve some methods.

% \begin{figure}[h!]
%   \centering
%   \begin{subfigure}[b]{0.4\textwidth}
%     \includegraphics[width=\textwidth]{figures/Pretrain/TissueMNIST_RAW_TEST_unified.pdf}
%     \caption{TissueMNIST}
%     \label{fig:sub1}
%   \end{subfigure}
%   \hfill
%   \begin{subfigure}[b]{0.4\textwidth}
%     \includegraphics[width=\textwidth]{figures/Pretrain/Path_RAW_TEST_unified.pdf}
%     \caption{PathMNIST}
%     \label{fig:sub2}
%   \end{subfigure}
%   \caption{Test performance}
%   \label{fig:image}
% \end{figure}

% \begin{figure}[h!]
%   \centering
%   \begin{subfigure}[b]{0.4\textwidth}
%     \includegraphics[width=\textwidth]{figures/Pretrain/TissueMNIST_RAW_VAL_unified.pdf}
%     \caption{TissueMNIST}
%     \label{fig:sub1}
%   \end{subfigure}
%   \hfill
%   \begin{subfigure}[b]{0.4\textwidth}
%     \includegraphics[width=\textwidth]{figures/Pretrain/Path_RAW_VAL_unified.pdf}
%     \caption{PathMNIST}
%     \label{fig:sub2}
%   \end{subfigure}
%   \caption{Val performance}
%   \label{fig:image}
% \end{figure}

\clearpage
\subsection{Validation-set profiles of accuracy-over-time}
\label{validation_result}

\begin{figure}[h!]
    \begin{subfigure}[b]{.48\textwidth}
         \includegraphics[width=.9\textwidth]{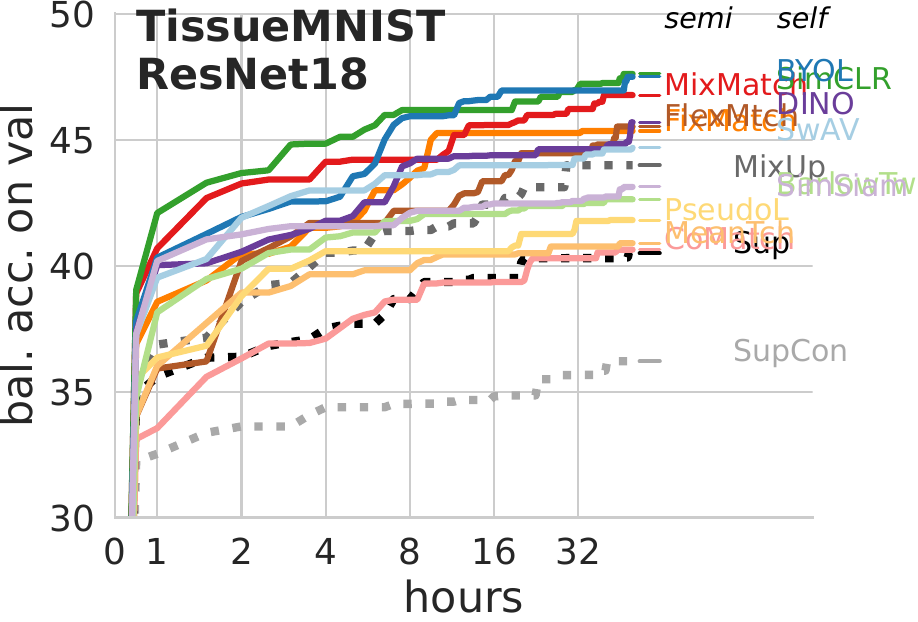}
        \caption{TissueMNIST}
    \end{subfigure}
\hspace{2mm}
    \begin{subfigure}[b]{.48\textwidth}
         \includegraphics[width=.9\textwidth]{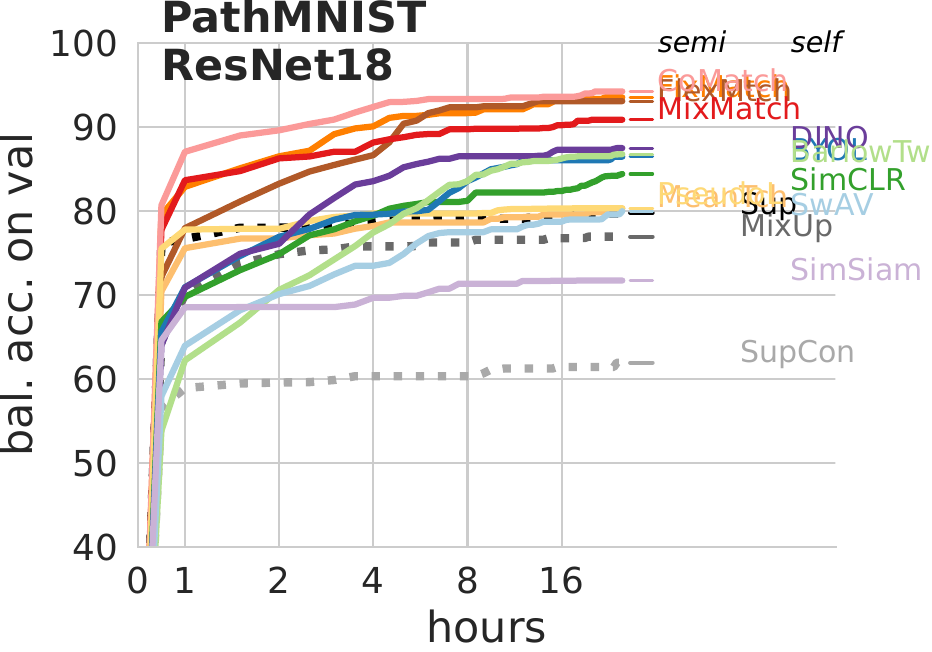}
        \caption{PathMNIST}
    \end{subfigure}
\\
    \begin{subfigure}[b]{.48\textwidth}
        \includegraphics[width=.9\textwidth]{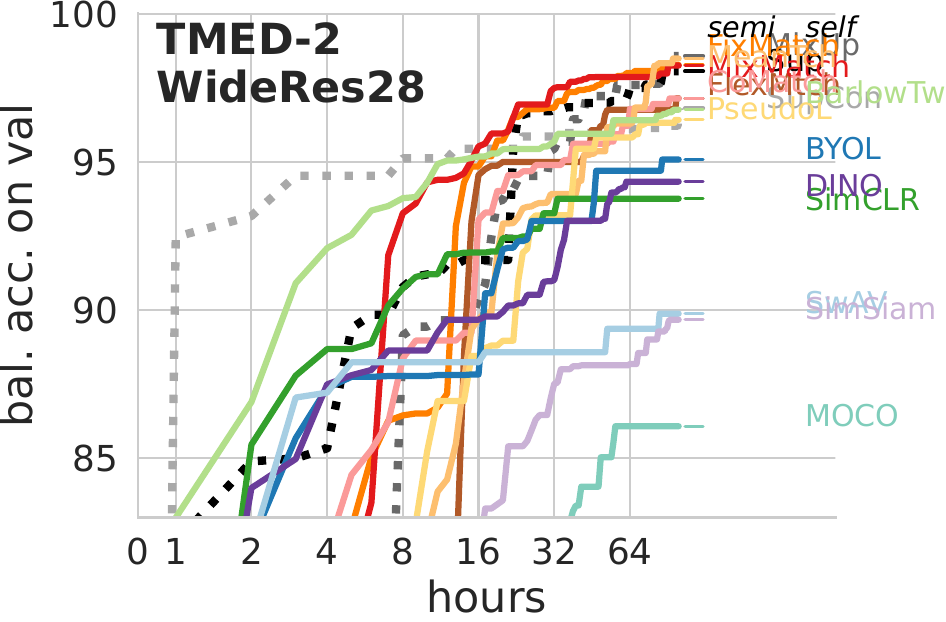}
        \caption{TMED2}
    \end{subfigure}
\hspace{2mm}
    \begin{subfigure}[b]{.48\textwidth}
        \includegraphics[width=.9\textwidth]{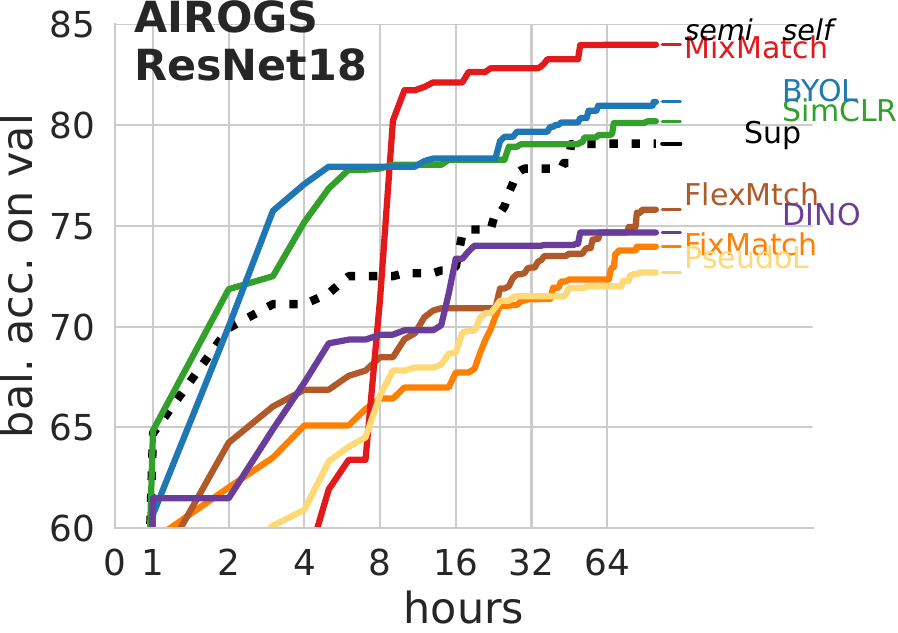}
        \caption{AIROGS}
    \end{subfigure}
% \raisebox{0.12\height}{%HACK fix vertical alignment
%     \begin{subfigure}[t]{.48\textwidth}
%         % \includegraphics[width=\textwidth]{figures_old/perf_vs_time_crop/legend_hz.pdf}
%         \includegraphics[width=\textwidth]{CVPR_figures/AIROGS/res18/AIROGS_RAW_VAL_Bacc_unified.pdf}
%         \caption{AIROGS}
%     \end{subfigure}
%     }%endraisebox
\caption{\textbf{Validation-set} accuracy over time profiles of semi- and self-supervised methods on 4 datasets (panels a-d). All curves here by definition must be monotonically increasing. The increasing profiles here on the validation set translate to similar trends in test set performance in Fig.~\ref{fig:test_performance_vs_time}, indicating successful generalization.
}
\label{fig:validation-acc-over-time}
\end{figure}

Fig.~\ref{fig:validation-acc-over-time} shows profiles of accuracy over time on the validation set, in contrast to the test set performance shown in the main paper's Fig.~\ref{fig:test_performance_vs_time}.

All curves here by definition must be monotonically increasing, because our unified algorithm selects new checkpoints only when they improve the validation-set balanced accuracy metric.
The important insight our work reveals is that the same model checkpoints selected here, based on validation-set accuracy, also tend to produce improved test-set accuracy over time (in Fig.~\ref{fig:test_performance_vs_time}).
This helps provide empirical confidence in using \textbf{\emph{realistically-sized}} validation sets.

\clearpage 
\subsection{Additional performance metrics: Profiles over time on AIROGS}
\label{More_metrics}

\begin{figure*}[h]
    \captionsetup[subfigure]{aboveskip=-1pt,belowskip=-1pt}
    \centering
    \begin{subfigure}[b]{0.23\textwidth}
        \includegraphics[width=\textwidth]{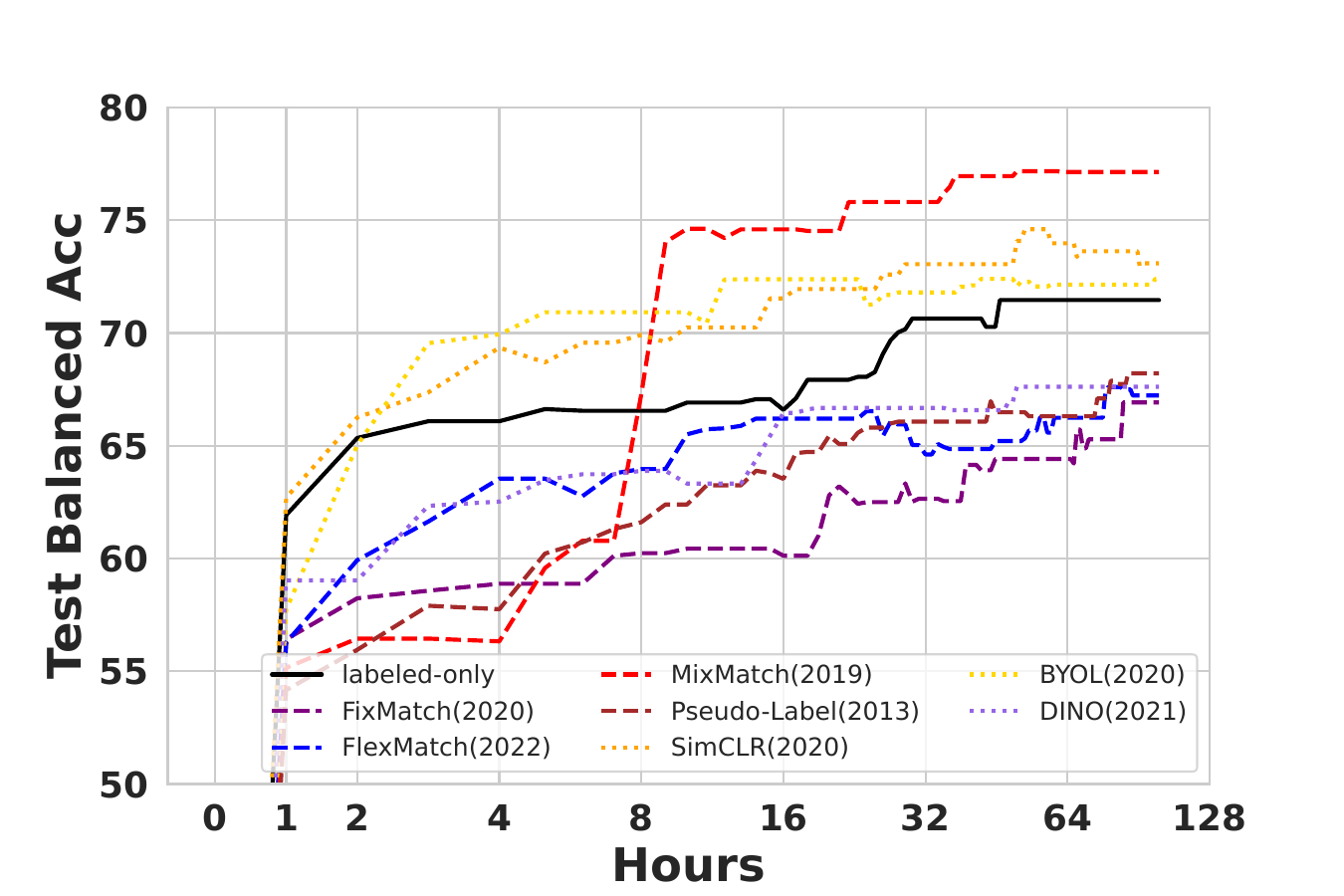}
        \caption{Balanced Acc}
    \end{subfigure}
    \hspace{1mm}
    \begin{subfigure}[b]{0.23\textwidth}
        \includegraphics[width=\textwidth]{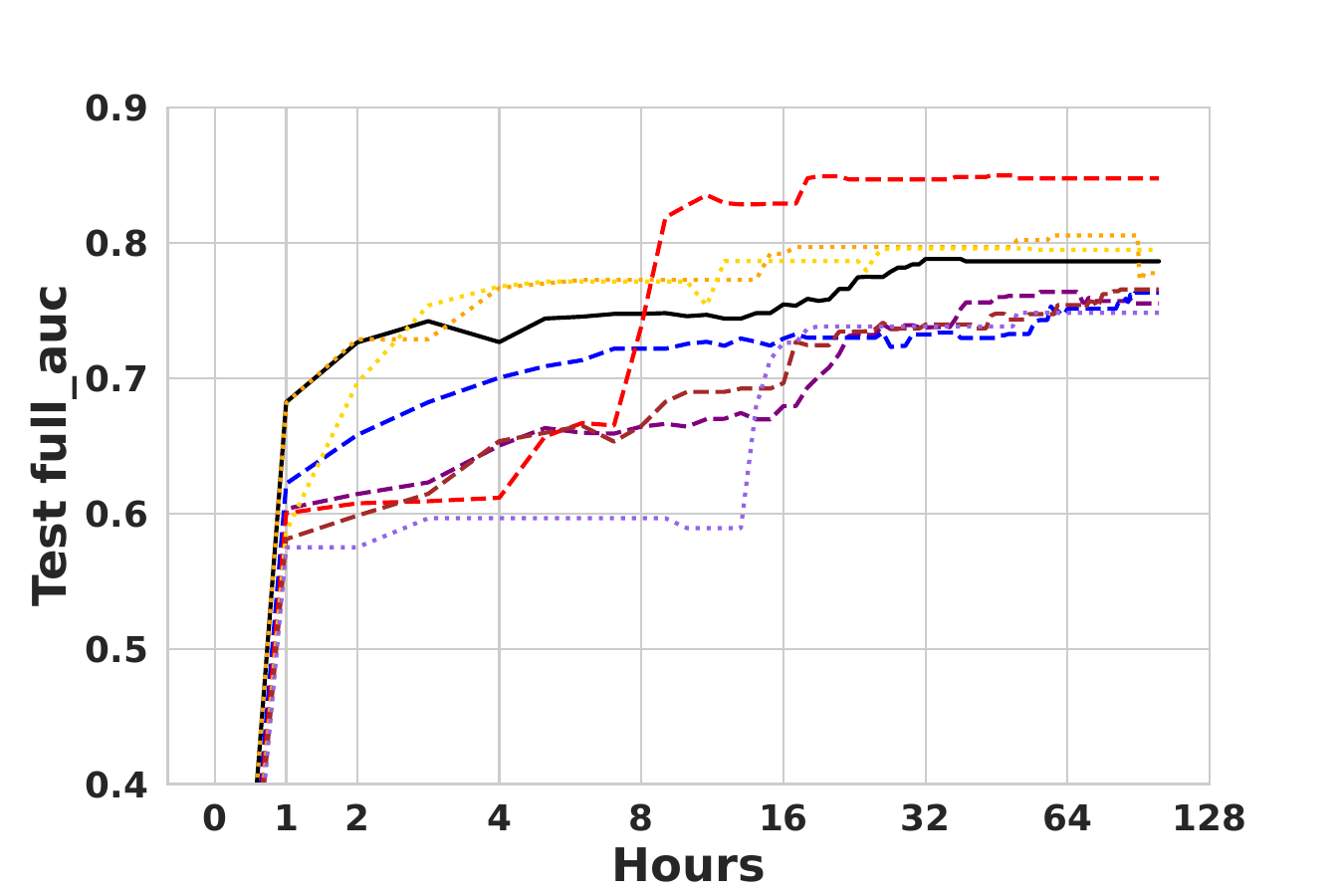}
        \caption{full auc}
        \label{fig:full_AUC}
    \end{subfigure}
    \hspace{1mm}
    \begin{subfigure}[b]{0.23\textwidth}
        \includegraphics[width=\textwidth]{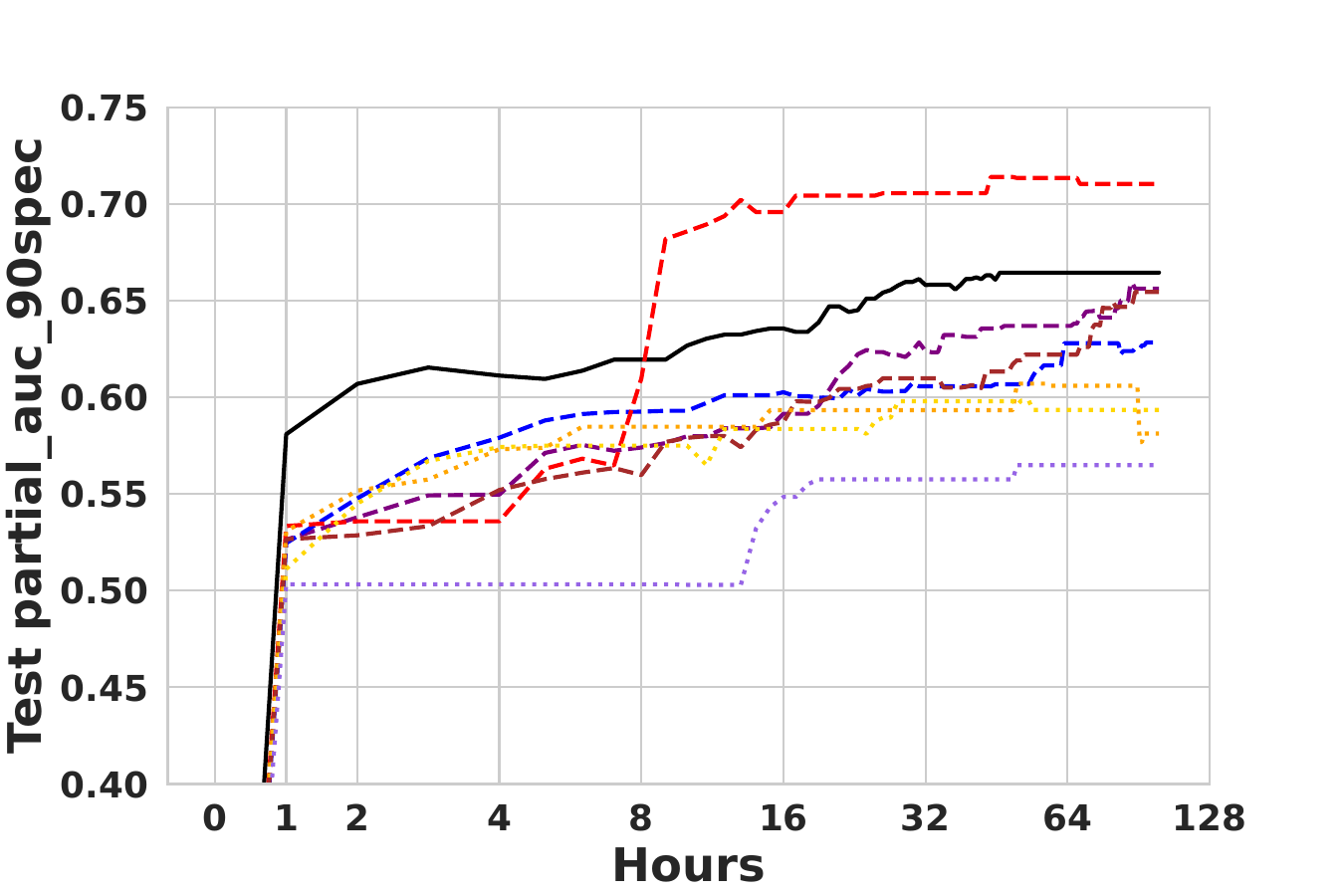}
        \caption{\small Partial AUC}
        \label{fig:partial_auc}
    \end{subfigure}
    \hspace{1mm}
    \begin{subfigure}[b]{0.23\textwidth}
        \includegraphics[width=\textwidth]{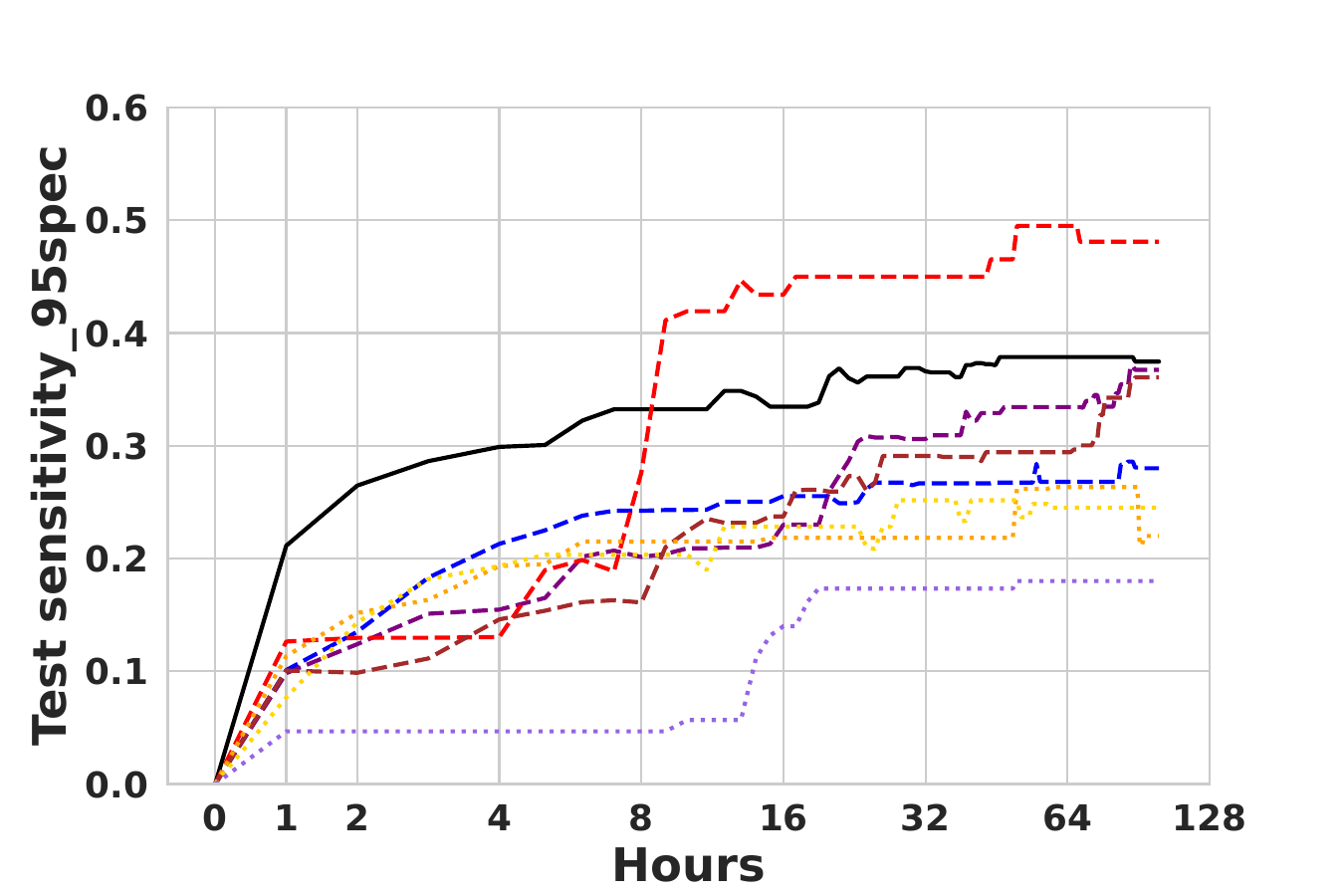}
        % \caption{Sensitivity at 95\% Specificity}
         \caption{Sensitivity}
        \label{fig:sensitivity}
    \end{subfigure}
    %\caption{\textbf{Res18. Test Performance over time profiles of semi- and self-supervised methods on AIROGS dataset across 3 metircs: Balanced Accuracy, Partial AUC for 90\% - 100\% specificity and Sensitivity at 95\% specificity (panels a-c).} At each time, we report mean of each method over \textcolor{red}{TODO: 5 (2 seeds shown in the plot, more to come soon)} trials of Alg.~\ref{alg:hyperparam_tuning}. \textcolor{red}{TODO: Self-supervised} results to come soon.}

    % HACK TO REDUCE WHITESPACE BETWEEN SUBFIG and SUBCAPTION
    \captionsetup[subfigure]{aboveskip=-1pt,belowskip=-1pt}
    \centering
    \begin{subfigure}[b]{0.23\textwidth}
        \includegraphics[width=\textwidth]{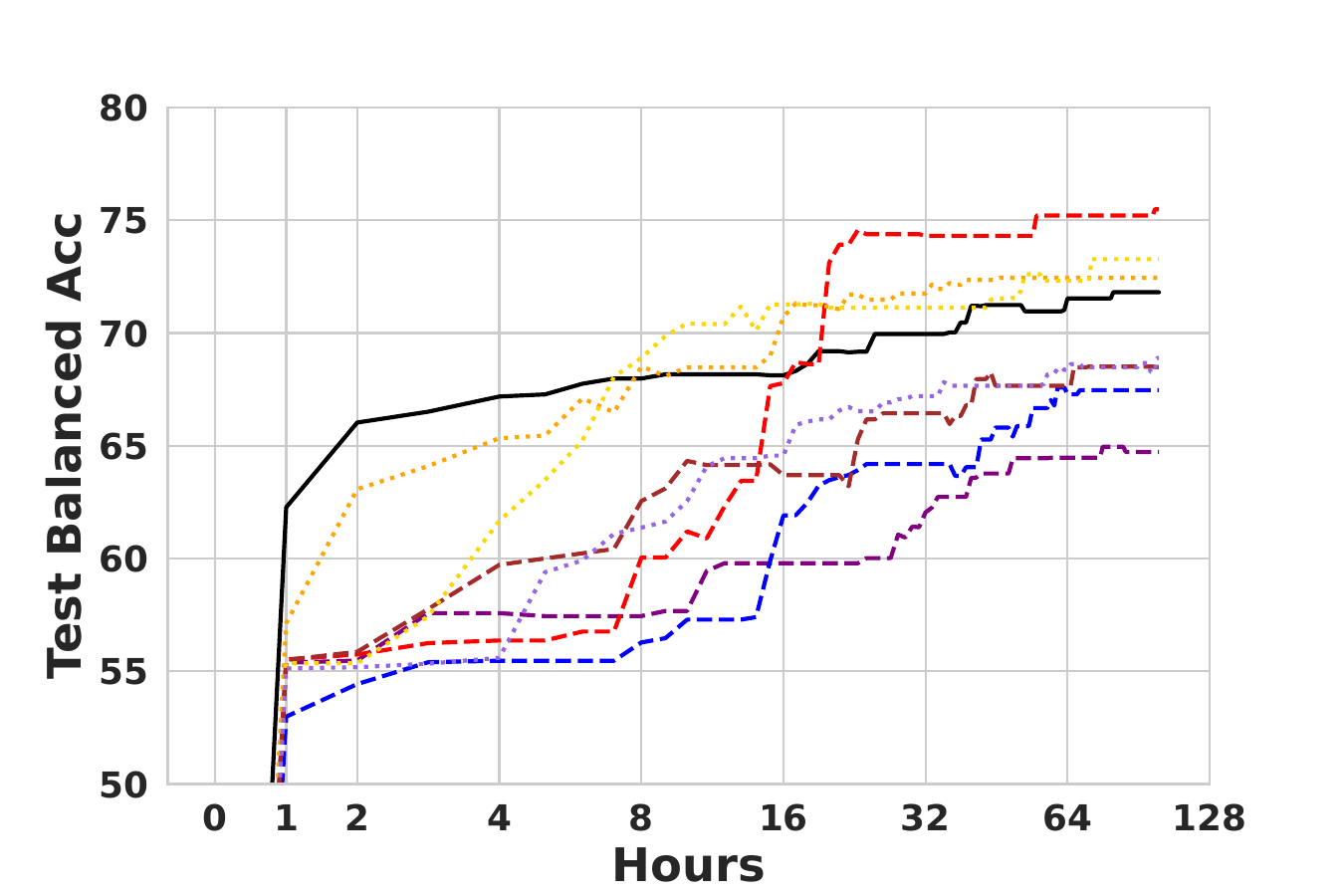}
        \caption{Balanced Acc}
    \end{subfigure}
    \hspace{1mm}
    \begin{subfigure}[b]{0.23\textwidth}
        \includegraphics[width=\textwidth]{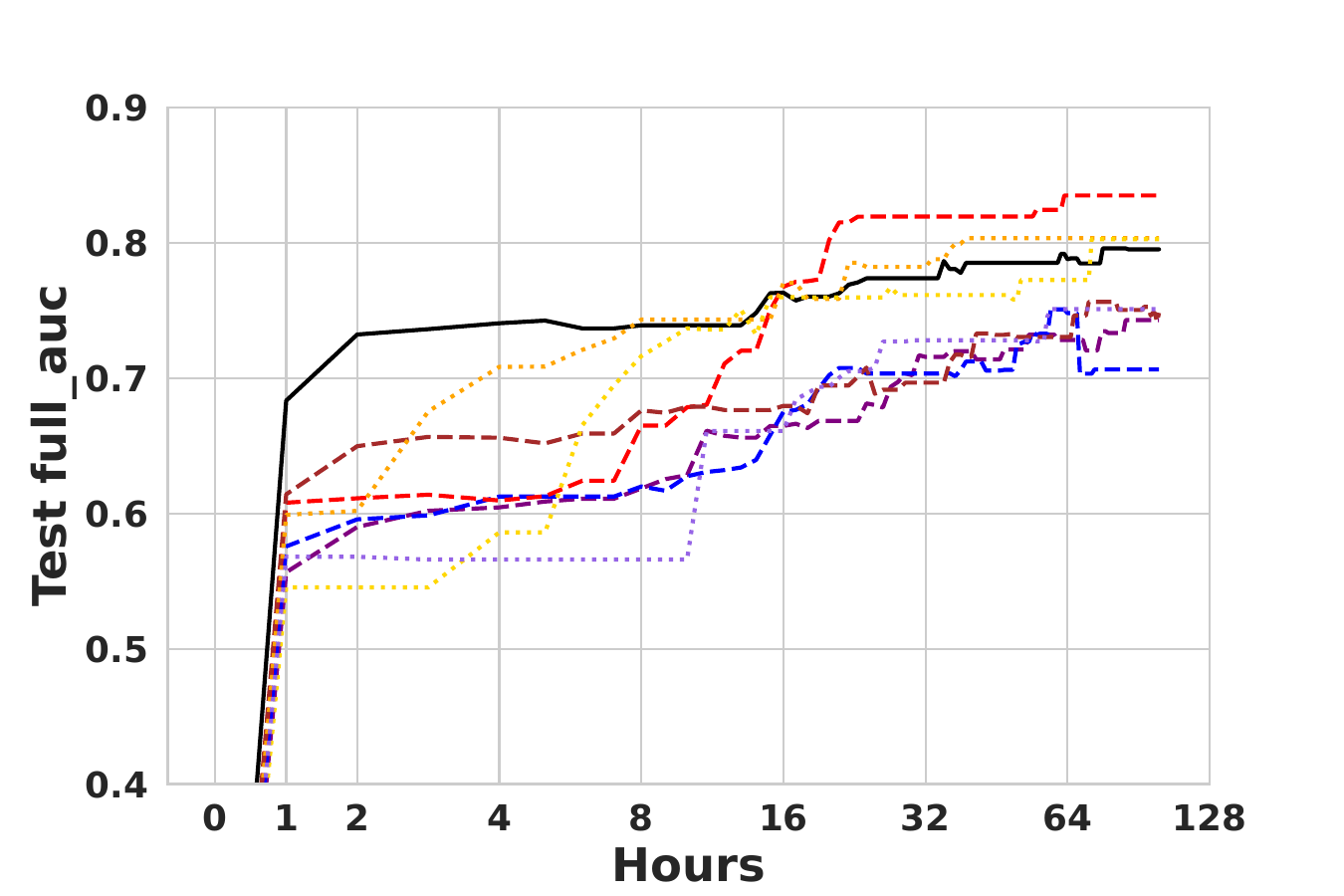}
        \caption{full auc}
    \end{subfigure}
    \hspace{1mm}
    \begin{subfigure}[b]{0.23\textwidth}
        \includegraphics[width=\textwidth]{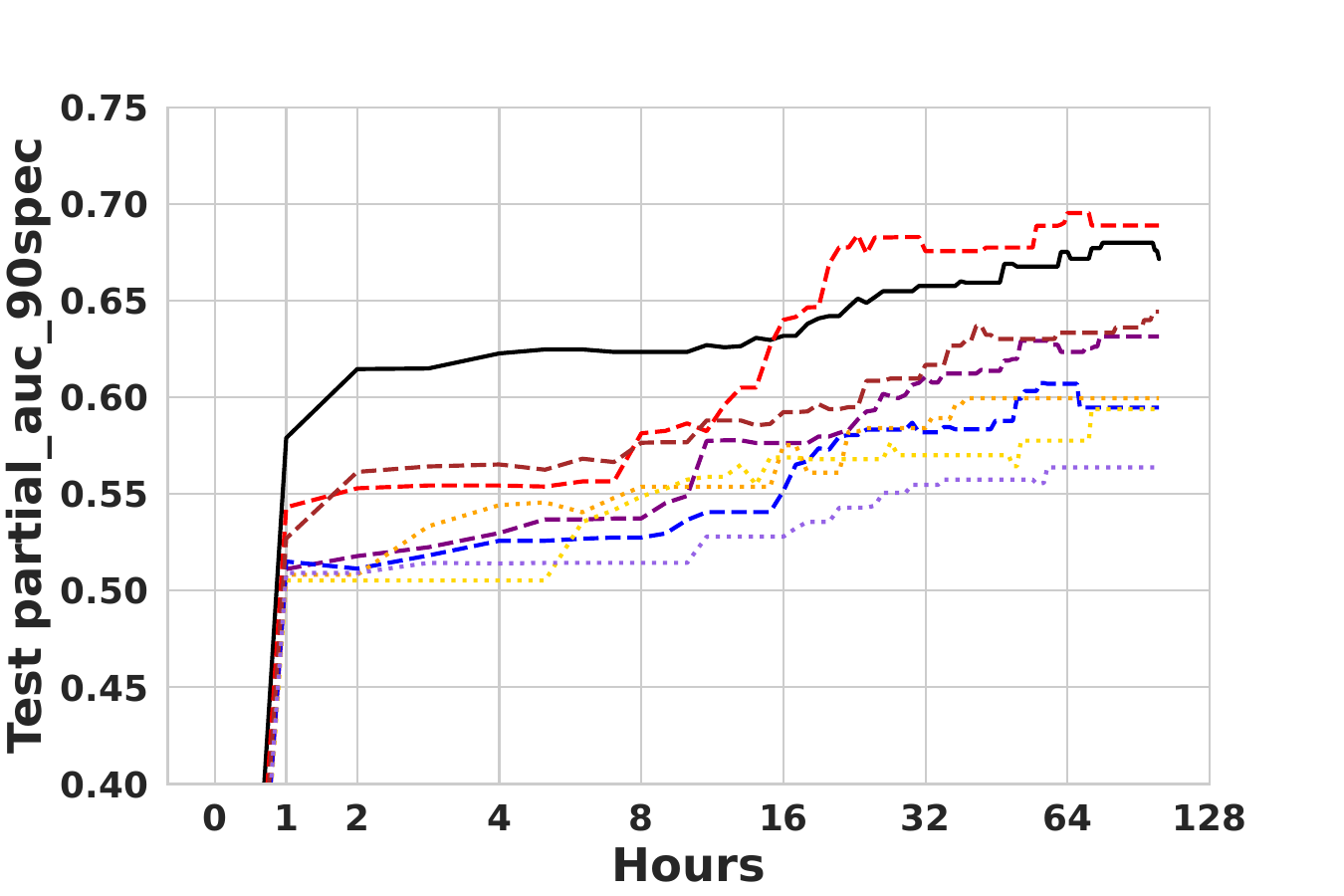}
        \caption{Partial AUC}
    \end{subfigure}
    \hspace{1mm}
    \begin{subfigure}[b]{0.23\textwidth}
        \includegraphics[width=\textwidth]{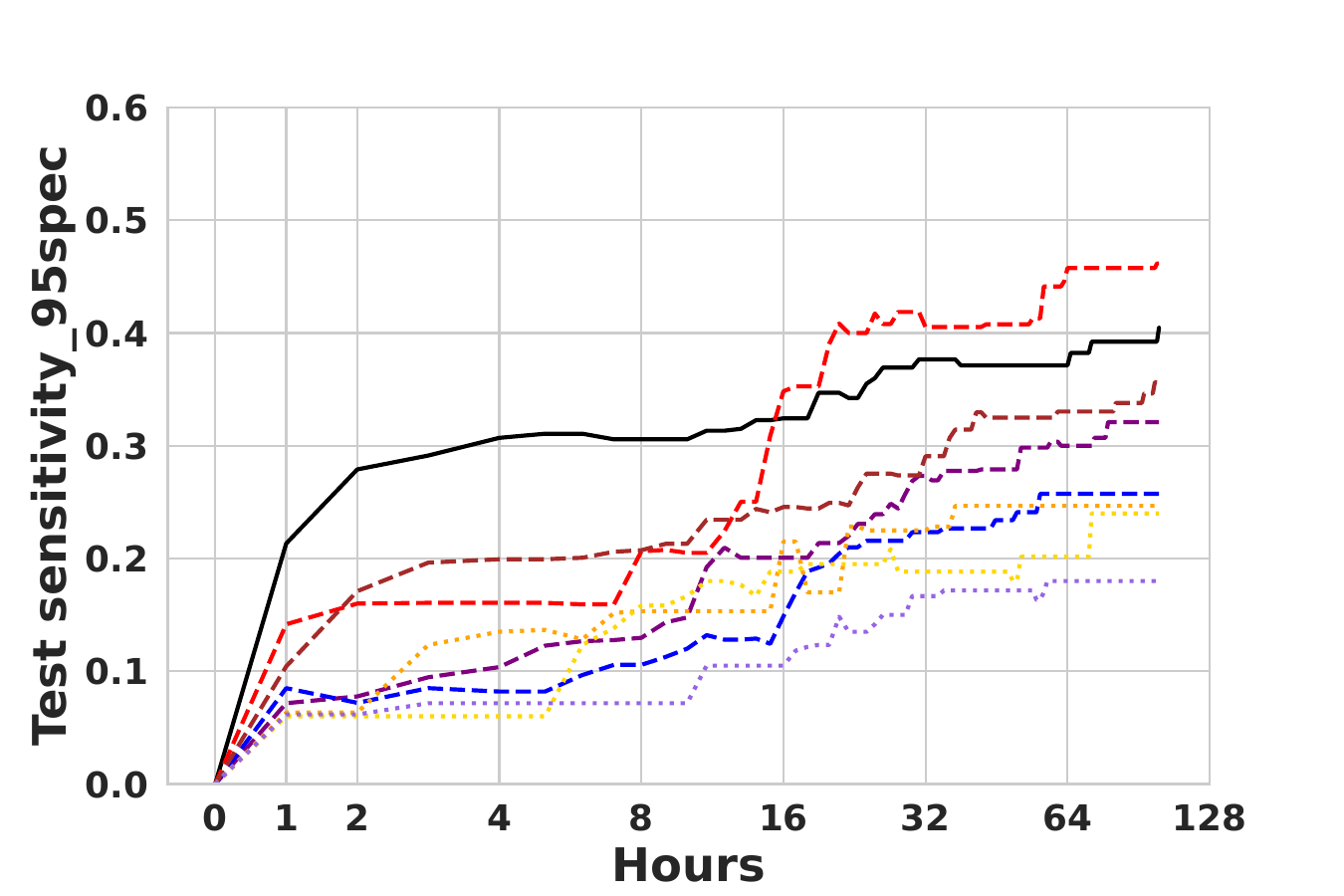}
        % \caption{Sensitivity@ 95\% Specificity}
        \caption{Sensitivity}
    \end{subfigure}
    \caption{\textbf{Profiles of several clinically-relevant performance metrics over time on the AIROGS test set.}
    \emph{Top row:} ResNet-18.
    \emph{Bottom row:} ResNet-50.
    \emph{Columns, left-to-right:}  Balanced Accuracy, AUROC, Partial AUROC focused on the 90\% - 100\% specificity regime, and sensitivity at 95\% specificity. At each time, we report mean of each method over 5 trials of Alg.~\ref{alg:hyperparam_tuning}.
    }%endcaption
    \label{fig:test_performance_vs_time_AIROGS}
\end{figure*}

In Fig.~\ref{fig:test_performance_vs_time_AIROGS}, we report the test performance over time on the AIROGS dataset across all 4 metrics of interest, including the partial AUROC and sensitivity at 95\% specificity metrics recommended by the AIROGS data creators as being particularly relevant for the glaucoma detection task. 

Broadly, our takeaway is that our proposed hyperparameter tuning method is viable for all these metrics, not just the BA metric covered in the main paper. Furthermore, this viability appears consistent across both ResNet-18 and ResNet-50 architectures.

%\todo{RJ Move the AIROGS res18 and res50 Other metrics figure here}

%Remove this section, as it does we didn't refer to it anywhere
% \subsection{Histograms of accuracy}

% \begin{figure}[h!]
%     %\hspace{-18mm}
%     \begin{subfigure}[b]{0.3\textwidth}
%         \includegraphics[width=\textwidth]{figures/Hyperparameter_space/TissueMNIST_RAW_TEST_unified.pdf}
%         %\caption{TissueMNIST}
%         \label{fig:tissuemnist}
%     \end{subfigure}
%     %\hspace{-7mm}
%     \hfill
%     \centering
%     \begin{subfigure}[b]{0.3\textwidth}
%         \includegraphics[width=\textwidth]{figures/Hyperparameter_space/PathMNIST_RAW_TEST_unified.pdf}
%         %\caption{PathMNIST}
%         \label{fig:pathmnist}
%     \end{subfigure}
%     %\hspace{-7mm}
%     \hfill
%     \begin{subfigure}[b]{0.3\textwidth}
%         \includegraphics[width=\textwidth]{figures/Hyperparameter_space/TMED2_RAW_TEST_unified.pdf}
%         %\caption{TMED2}
%         \label{fig:tmed2}
%     \end{subfigure}
%     \caption{Distribution of test accuracy that corresponded to the maximum validation accuracy. From left to right are TissueMNIST, PathMNIST and TMED2.}
%     \label{fig:testfigures}
% \end{figure}

\subsection{Variability in Performance Across Trials}

In Fig.~\ref{fig:tmedfigures_timepoint} on the next page, we explicitly visualize the variability in performance of each method across the 5 separate trials of Alg.~\ref{alg:hyperparam_tuning} (most other figures show the mean of these 5 trials for visual clarity).

\newcommand{\Wvar}{0.4}
\begin{figure}[t]
        \captionsetup[subfigure]{aboveskip=-1pt,belowskip=-3pt}
    \centering
    \begin{subfigure}[b]{\Wvar\textwidth}
        \includegraphics[width=\textwidth]{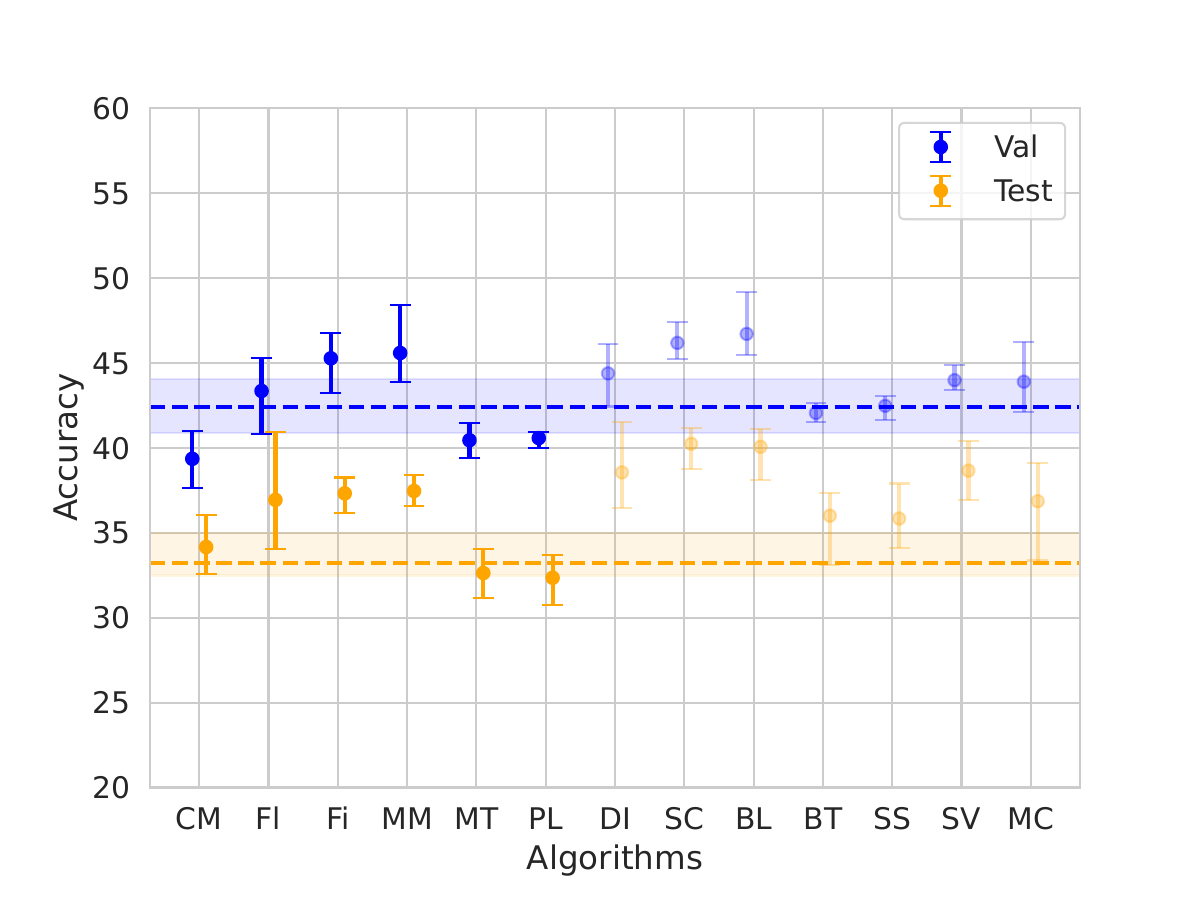}
        \caption{TissueMNIST after 16h}
    \end{subfigure}
    % \hfill
    \begin{subfigure}[b]{\Wvar\textwidth}
        \includegraphics[width=\textwidth]{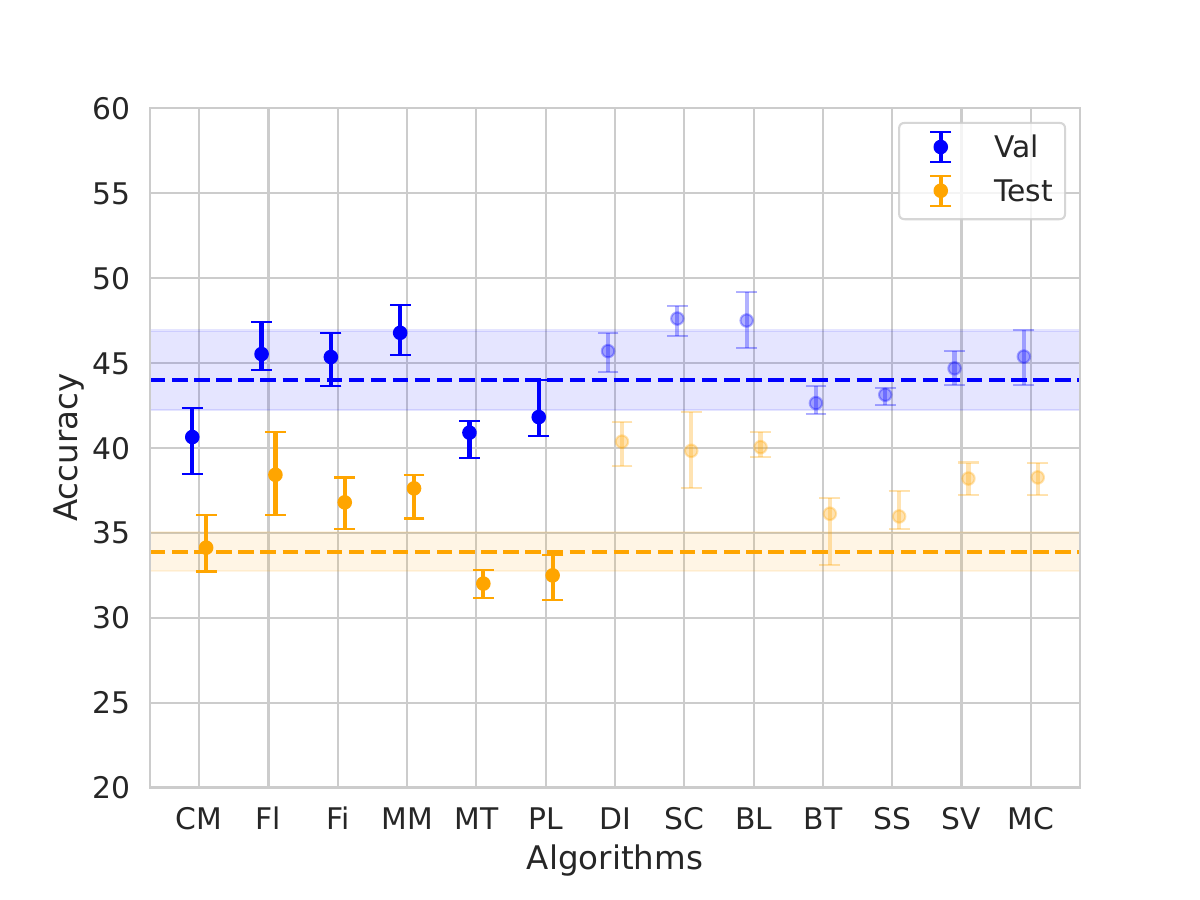}
        \caption{TissueMNIST after 50h}
    \end{subfigure}
    \\
    \vspace{-4mm}
    \begin{subfigure}[b]{\Wvar\textwidth}
        \includegraphics[width=\textwidth]{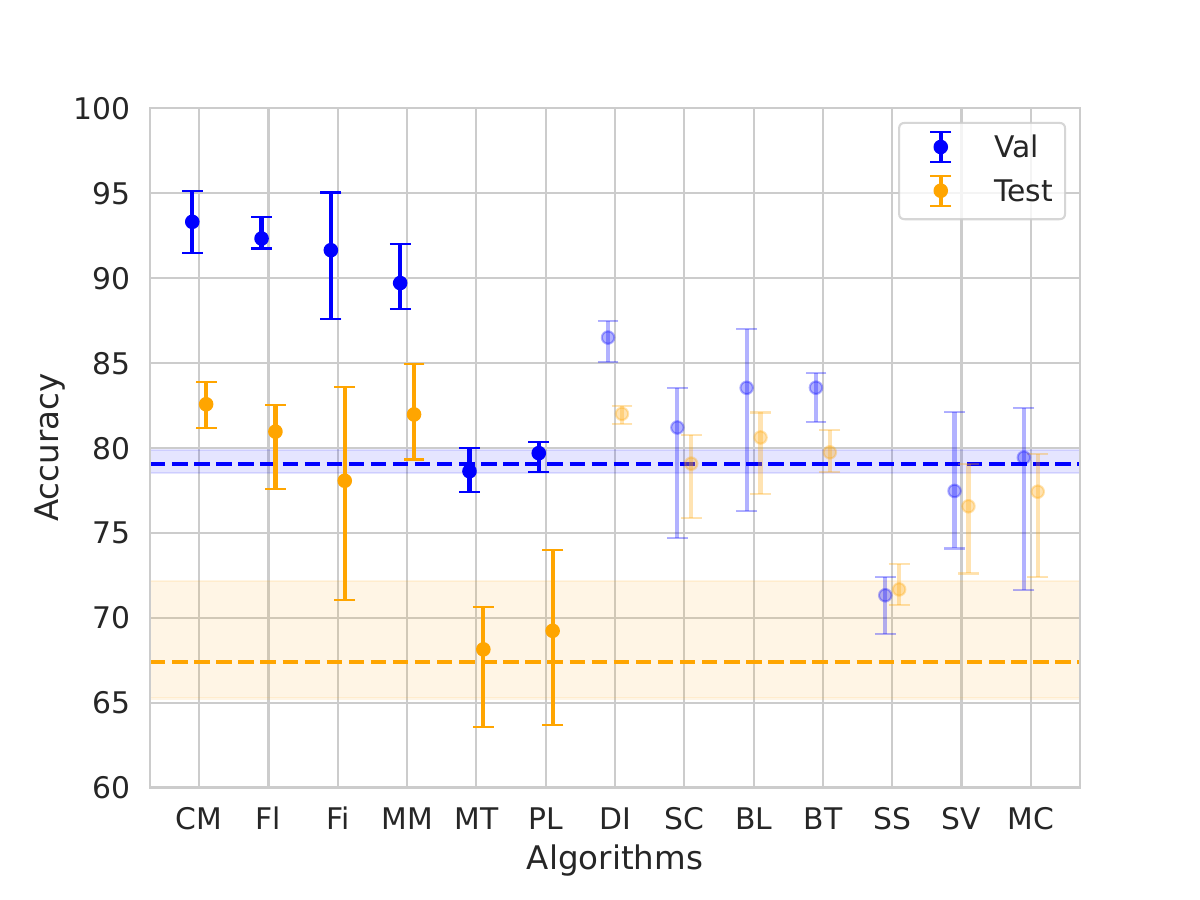}
        \caption{PathMNIST after 8h}
    \end{subfigure}
    % \hfill
    \begin{subfigure}[b]{\Wvar\textwidth}
        \includegraphics[width=\textwidth]{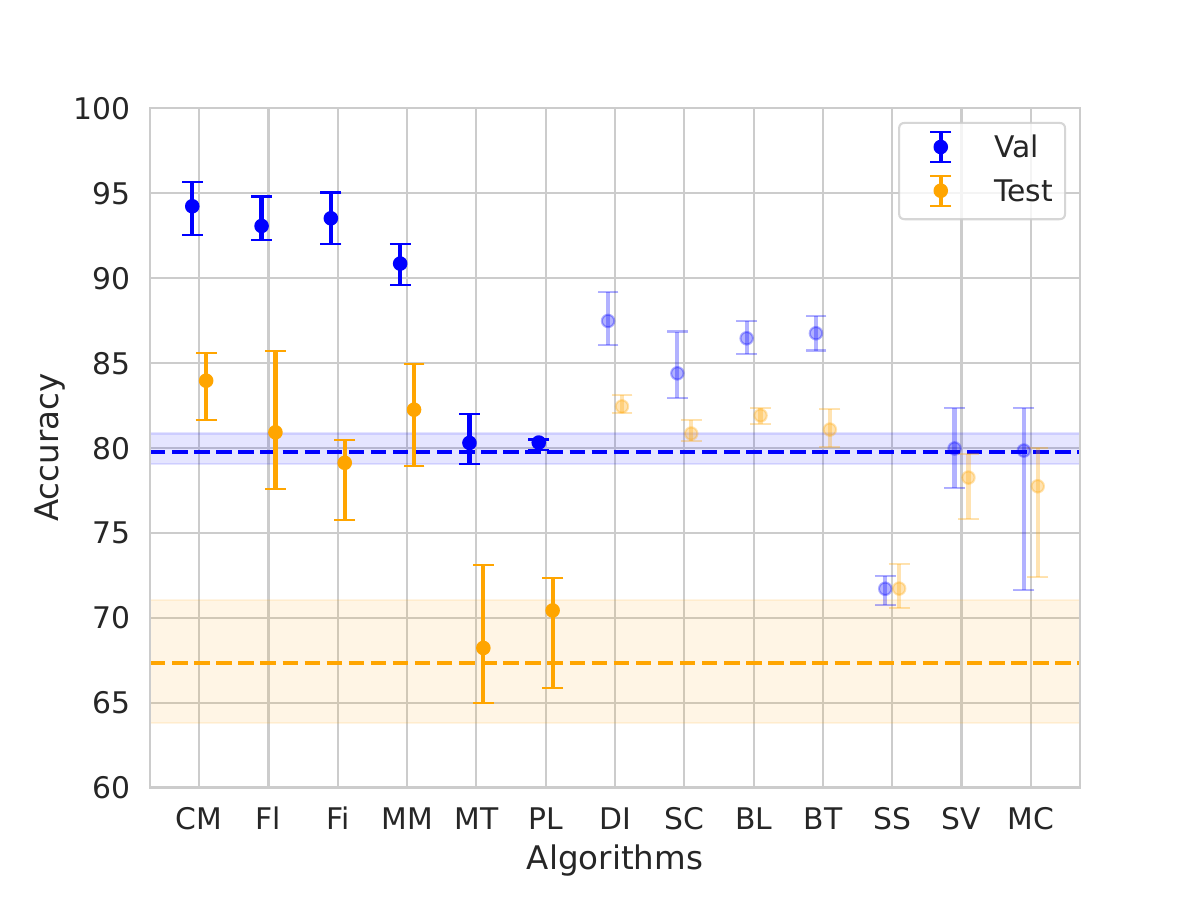}
        \caption{PathMNIST after 25h}
    \end{subfigure}
    \\
    \vspace{-4mm}
    \begin{subfigure}[b]{\Wvar\textwidth}
        \includegraphics[width=\textwidth]{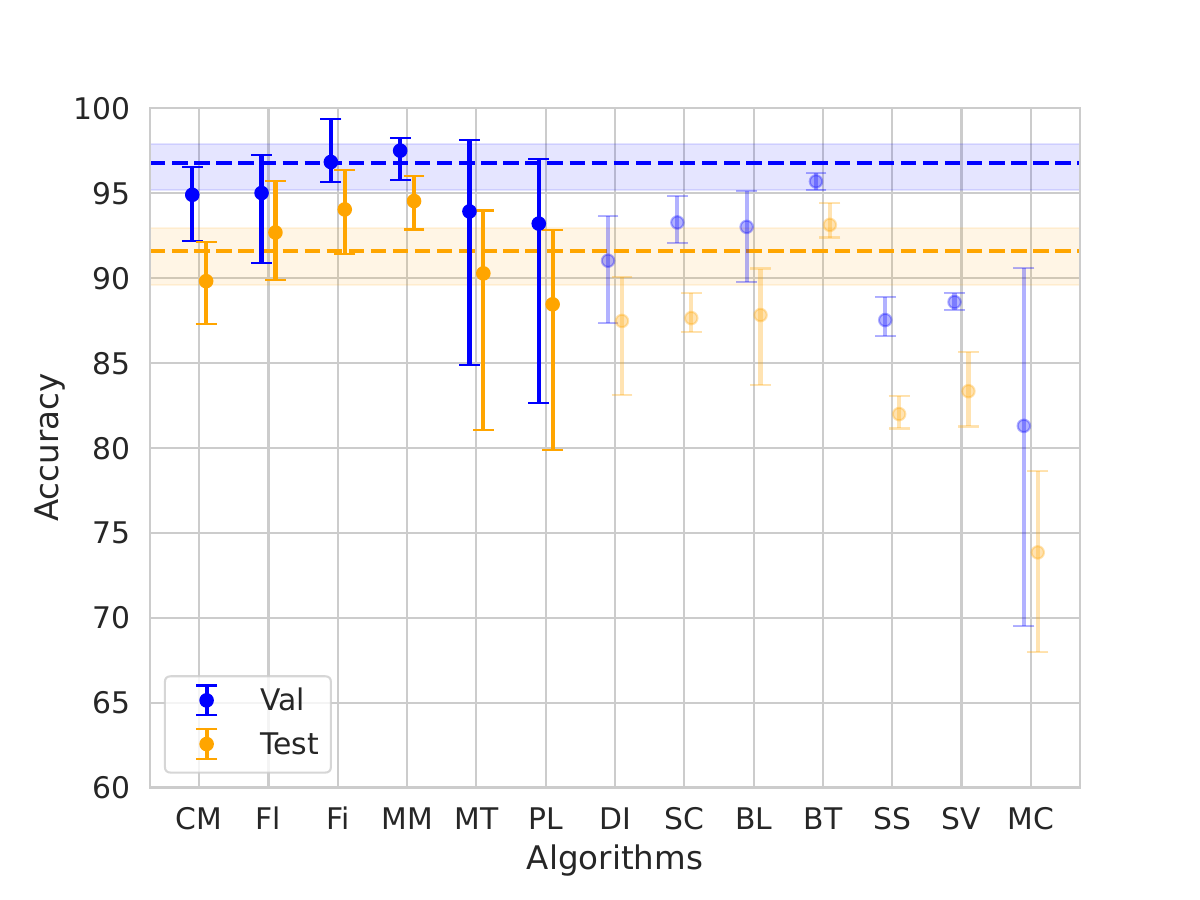}
        \caption{TMED-2 after 32h}
    \end{subfigure}
    % \hfill
    \begin{subfigure}[b]{\Wvar\textwidth}
        \includegraphics[width=\textwidth]{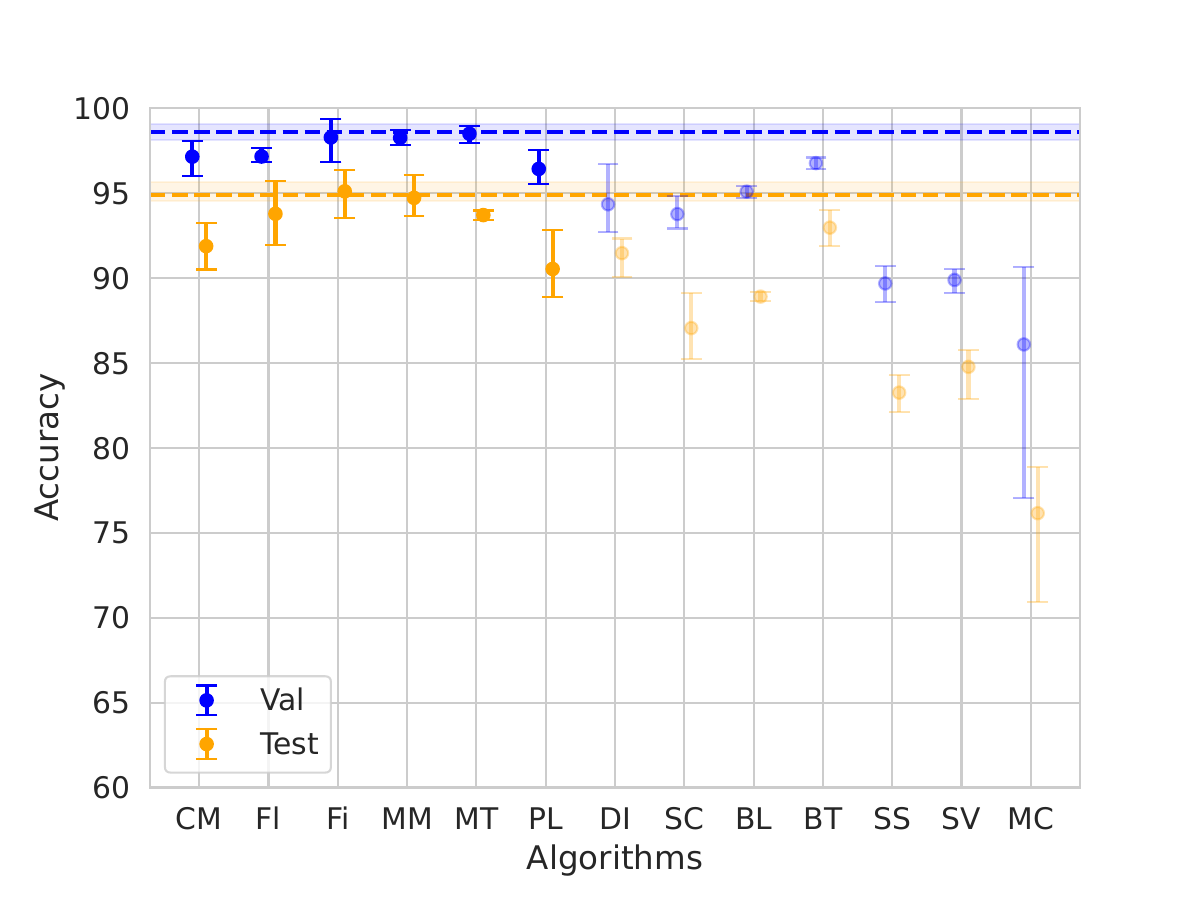}
        \caption{TMED-2 after 100h}
    \end{subfigure}
    \\
    \vspace{-4mm}
    \begin{subfigure}[b]{\Wvar\textwidth}
        \includegraphics[width=\textwidth]{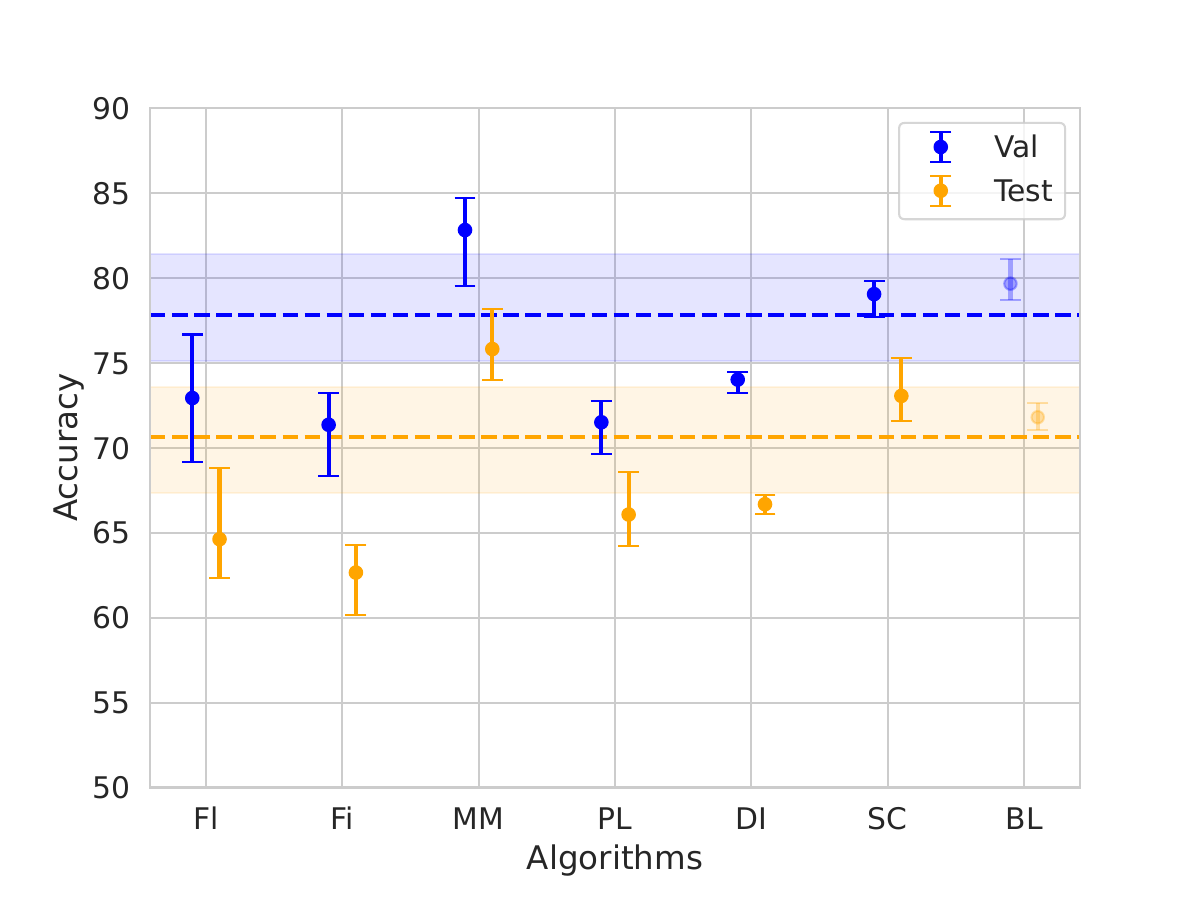}
        \caption{AIROGS(res18) after 32h}
    \end{subfigure}
    % \hfill
    \begin{subfigure}[b]{\Wvar\textwidth}
        \includegraphics[width=\textwidth]{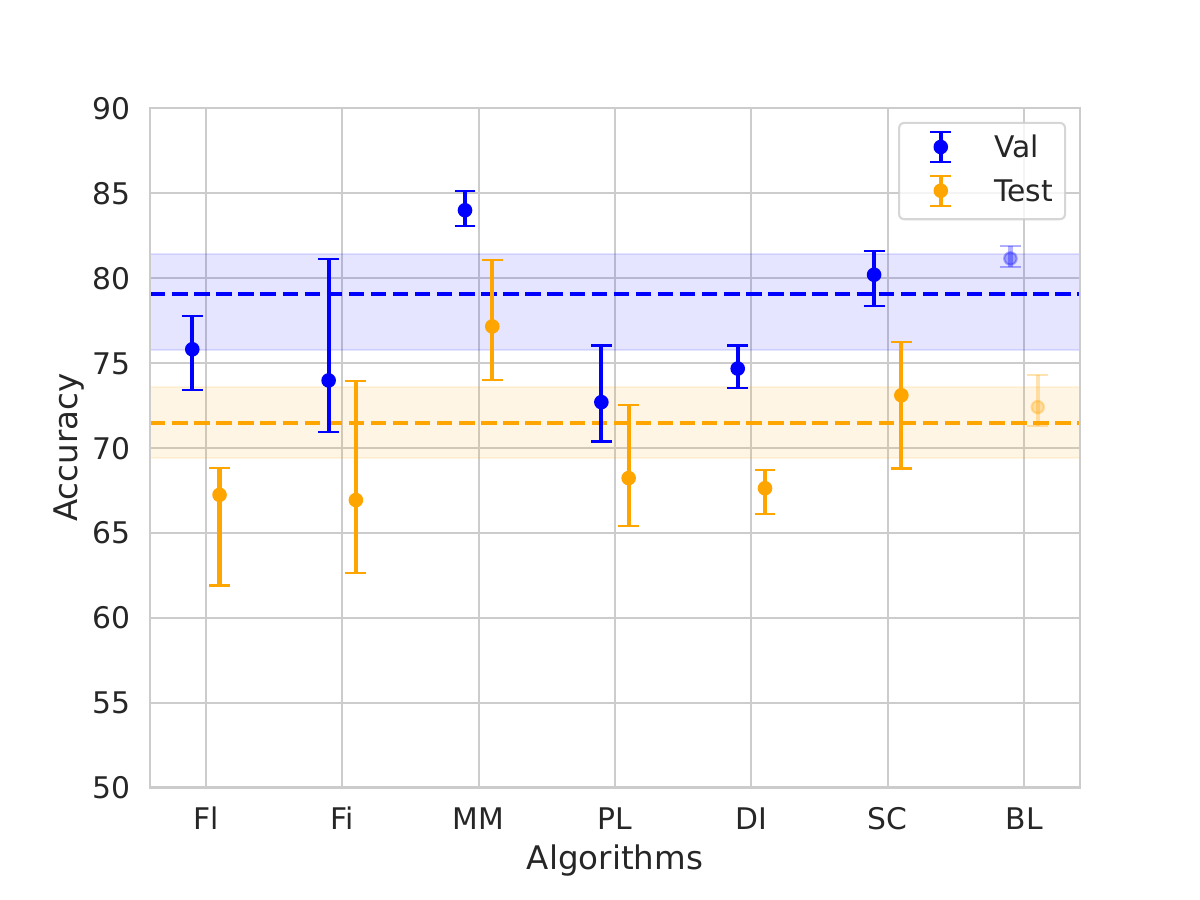}
        \caption{AIROGS(res18) after 100h}
    \end{subfigure}
    \vspace{-2mm}
    \caption{Balanced accuracy of different methods across 2 time budgets (columns) and four datasets (rows).
    For each method, the interval indicates the low and high performance of 5 separate trials of Alg.~\ref{alg:hyperparam_tuning}, while dot indicates the mean performance. Horizontal lines indicate the best labeled-set-only baseline at that time. Abbreviation: CM, Fl, Fi, MM, MT, PL, DI, SC, BL, BT, SS, SV, MC denote CoMatch, FlexMatch, FixMatch, MixMatch, Mean Teacher, Pseudo Label, DINO, SimCLR, BYOL, Barlow Twins, SimSiam, SwAV, MOCO (v2).
    }%endcaption
    \label{fig:tmedfigures_timepoint} % Keep this, main text needs this exact label
\end{figure}

\clearpage

\section{Method Details}
%\todo{RJ}
\label{App:Algorithms}

\subsection{Algorithm : Unified training and hyperparameter tuning via random search on a budget}
Algorithm~\ref{alg:hyperparam_tuning} outlines the hyperparameter tuning procedures used across all algorithms under comparison. 
The algorithm requires three sources of data:
a labeled training set $\mathcal{L} = \{X,Y\}$, an unlabeled set for training $\mathcal{U} = X^U$, and a separate realistically-sized labeled validation set $\{X^{val}, Y^{val}\}$.
We further require some budget restrictions: a common computational budget $T$ (maximum number of hours), and a maximum training epoch per hyperparameter configuration $E$.

We proceed as follows: We begin by randomly sampling a hyperparameter configuration from a defined range (see Appendix~\ref{App_Hyper_Details} for details). A model is then initialized and trained using the ADAM optimizer with the sampled hyperparameters. Each configuration is trained for a maximum of $E$ (200) epochs or stopped early if the validation performance does not improve for 20 consecutive epochs. The model's performance on the validation set is measured using balanced accuracy. Upon completion of training for a given hyperparameter configuration (either after reaching maximum epoch $E$ or after early stopping), a new configuration is sampled and the process repeats until the total compute budget $T$ is expended.

We track the best-so-far model performance every 30 minutes, and save the best-so-far model along with its validation and test performance. Semi-supervised algorithms simultaneously train the representation layers $v$ and classifier layer $w$, while self-supervised algorithms train the representation layers $v$ for each epoch and then fine-tune a linear classifier with weights $w$ anew at the end of each epoch using an sklearn logistic regression model~\citep{pedregosa2011scikit} with representation parameters $v$ frozen.
 
\begin{algorithm}[!h]
\caption{Unified Procedure for Training + Hyperparameter selection via random search}
\label{alg:hyperparam_tuning}
\begin{flushleft}
\textbf{Input}: 
\begin{itemize}
    \item Train set of features $\mathbf{X}$ paired with labels $\mathbf{Y}$, with extra unlabeled features $\mathbf{U}$
    \item Validation set of features $\mathbf{X}^{\text{val}}$ and labels $\mathbf{Y}^{\text{val}}$
    \item Runtime budget $T$, Max Epoch $E$
\end{itemize}
\textbf{Output}: Trained weights $\{v, w\}$, where $v$ is the representation module, $w$ is the classifier layer
\end{flushleft}
\begin{algorithmic}[1] %[1] enables line numbers
\While{time elapsed $<$ T}
    \State $\lambda \sim \textsc{DrawHypers}$ \Comment{Sample hyperparameters from pre-defined range (App.~\ref{App_Hyper_Details})}
    \State $\xi \gets \textsc{CreateOptim}(\lambda)$ \Comment{Initialize stateful optimizer e.g., ADAM}
    \State $\{v, w\} \sim \textsc{InitWeights}$ \Comment{Initialize model weights}
    \For{epoch $e$ in $1, 2, \ldots, E$}
        \If{self-supervised}
            \State $v \gets \textsc{TrainOneEpoch}(\mathbf{U}, v, \lambda, \xi)$
            \Comment{Optimize Eq.~\eqref{eq:objective_unified} with $\lambda^L=0$}
            \State $w \gets \textsc{TrainClassifier}(\mathbf{Y}, f_v(\mathbf{X}))$
        \ElsIf{semi-supervised}
            \State $v, w \gets \textsc{TrainOneEpoch}(\mathbf{X}, \mathbf{Y}, \mathbf{U}, v, w, \lambda, \xi)$
            \Comment{Optimize Eq.~\eqref{eq:objective_unified}}
        \Else
            \State $v, w \gets \textsc{TrainOneEpoch}(\mathbf{X}, \mathbf{Y}, v, w, \lambda, \xi)$
            \Comment{Optimize Eq.~\eqref{eq:objective_unified} with $\lambda^U=0$}
        \EndIf
        \State $m_e \gets \textsc{CalcPerf}(\mathbf{X}^{\text{val}}, \mathbf{Y}^{\text{val}}, v, w)$
        \Comment{Record performance metric on val.}
        \If{first try or $m_e > m_*$}
            \State $v_*, w_* \gets v, w$
            \State $\lambda_* \gets \lambda$
            \State $m_* \gets m_e$
            \Comment{Update best config found so far}
        \EndIf
        \If{\textsc{EarlyStop}($m_1, m_2, \ldots, m_e$) or time elapsed $> T$}
            \State \textbf{break}
        \EndIf
    \EndFor
\EndWhile
\State \textbf{return} $v_*, w_*, \lambda_*, m_*$
\end{algorithmic}
\end{algorithm}

\subsection{Semi-supervised method details}
\label{app:semi_supervised_method_details}

Semi-supervised learning trains on the labeled and unlabeled data simultaneously, usually with the total loss being a weighted sum of a labeled loss term and an unlabeled loss term. Different methods mainly differs in how unlabeled data is used to form training signals. Many approaches have been proposed and refined over the past decades. These include co-training, which involves training multiple classifiers on various views of the input data~\citep{blum1998combining,min2020mutually};
graph-structure-based models~\citep{zhu2003semi,iscen2019label}; generative models~\citep{kingma2014semi,kumar2017semi}; consistency regularization-based models that enforce consistent model outputs~\citep{laine2016temporal,tarvainen2017mean,berthelot2019mixmatch}; pseudo label-based models that impute labels for unlabeled data ~\citep{lee2013pseudo,cascante2021curriculum}; and hybrid models that combines several methods~\citep{sohn2020fixmatch}. Comprehensive reviews can be found in~\citet{zhu2005semi,chapelle2009semi,van2020survey}.

Among the deep classifier methods following Eq.~\eqref{eq:objective_unified}, below we describe each method we selected and how its specific unlabeled loss is constructed.

\textbf{Pseudo-Labeling} uses the current model to  assign class probabilities to each sample in the unlabeled batch. If, for an unlabeled sample, the maximum class probability $P(y_i)$ exceeds a certain threshold $\tau$, this sample contributes to the calculation of the unlabeled loss for the current batch. The cross-entropy loss is computed as if the true label of this sample is class $i$.

\textbf{Mean-Teacher} constructs the unlabeled loss by enforcing consistency between the model's output for a given sample and the output of the same sample from the Exponential Moving Average (EMA) model.

\textbf{MixMatch} uses the MixUp~\citep{zhang2017mixup} technique on both labeled data (features and labels) and unlabeled data (features and guessed labels) within each batch to produce transformed labeled and unlabeled data. The labeled and unlabeled losses are then calculated using these transformed samples. Specifically, the unlabeled loss is derived from the mean squared error between the model's output for the transformed unlabeled samples and their corresponding transformed guessed labels.

\textbf{FixMatch} generates two augmentation of an unlabeled sample, one with weak augmentation and the other using strong augmentations (e.g., RandAug~\citep{cubuk2020randaugment}). The unlabeled loss is then formulated by enforcing the model's output for the strongly augmented sample to closely resemble that of the weakly augmented sample using cross-entropy loss.

\textbf{FlexMatch} builds directly upon FixMatch by incorporating a class-specific threshold on the unlabeled samples during training. 

\textbf{CoMatch} marks the first introduction of contrastive learning into semi-supervised learning. The model is equipped with two distinct heads: a classification head, which outputs class probabilities for a given sample, and a projection head, which maps the sample into a low-dimensional embedding. These two components interact in a unique manner. The projection head-derived embeddings inform the similarities between different samples, which are then used to refine the pseudo-labels against which the classification head is trained. Subsequently, these pseudo-labels constitute a pseudo-label graph that trains the embedding graph produced by the projection head.

\subsection{Self-supervised method details}
\label{app:self_supervised_method_details}

In recent years, self-supervised learning algorithms have emerged rapidly and are known as one of the most popular field of machine learning. These include contrastive learning, which involves learning representations by maximizing agreement between differently augmented views of the same data~\citep{chen2020simple,he2020momentum}; predictive models that forecast future instances in the data sequence~\citep{oord2018representation}; generative models that learn to generate new data similar to the input~\citep{chen2020generative}; clustering-based approaches that learn representations by grouping similar instances~\citep{caron2018deep,caron2020unsupervised}; context-based models that predict a specific part of the data from other parts~\citep{devlin2018bert,brown2020language}; and hybrid models that combine various methods for more robust learning~\citep{chen2020big}. A more comprehensive review can be found in~\citep{jing2020self,zhuang2020comprehensive}.

Below, we provide for each selected self-supervised method a summary of its internal workings.

\textbf{SimCLR} generates two augmented versions of each image. Then feed these pairs of images into a base encoder network to generate image embeddings. This encoder is followed by a projection head, which is a multilayer neural network, to map these embeddings to a space where contrastive loss can be applied. Next, calculate the contrastive loss. The idea is to make the embeddings of augmented versions of the same image (positive pairs) as similar as possible and to push apart embeddings from different images (negative pairs). The loss function used is NCE loss.

\textbf{MOCO V2} creates two augmented versions of each image. These pairs are processed by two encoder networks: a query encoder, and a key encoder updated by a moving average of the query encoder. The contrastive loss is computed by comparing a positive pair (the query and corresponding key) against numerous negative pairs drawn from a large queue of keys. 

\emph{Note on runtime:}
We notice that the performance on MoCo can be increased when Shuffling BN across multiple GPUs. However, to ensure a fair comparison given our single-GPU setup, we refrained from employing any techniques to simulate multiple GPUs on one, as this would change the encoder's structure. 

\textbf{SwAV} begins by creating multiple augmented versions of each image. Then, these versions are input into a deep neural network to generate embeddings. Uses a clustering approach, called online stratified sampling, to predict assignments of each view's prototypes (or cluster centers) to others, encouraging the model to match the representations of different augmentations of the same image.

\emph{Note on runtime:}
We've observed that applying multiple augmentations can enhance the effectiveness of various methods. To prevent the results from being influenced by these augmentations, we've standardized the number of augmentations to two in SwAV, in line with the approach taken by other methods.

\textbf{BYOL} starts by creating two differently augmented versions of each image. These versions are processed through two identical neural networks, known as the target and online networks, which include a backbone and a projection head. The online network is updated through backpropagation, while the target network's weights are updated as a moving average of the online network's weights. The unique aspect of BYOL is that it learns representations without the need for negative samples.

\textbf{SimSiam} creates two differently augmented versions of each image. These versions are passed through two identical networks: one predictor network and one encoder network. The encoder network contains a backbone and a projection head.

\textbf{DINO} utilizes two differently augmented images, processed by a student and a teacher network. The teacher's weights evolve as a moving average of the student's. The key idea is self-distillation, where the student's outputs match the teacher's for one view but differ for the other, without traditional negative samples.

\textbf{Barlow Twins} processes two augmented views of an image through identical networks. The aim is to have similar representations between these networks while minimizing redundancy in their components, sidestepping the need for contrasting positive and negative pairs.

\section{Additional Analysis and Discussion}
\label{app:Additional_Analysis}

%\subsection{Hyperparameter Tuning and Model selection with realistic validation set}

\subsection{Effectiveness of Hyperparameter Tuning}
\label{app:Effectivness of Hyperparameter Tuning}

While \citet{oliver2018realistic} caution that extensive hyperparameter search may be futile with realistic validation set. Our experiments on the 4 dataset show that the validation set performance for each examined algorithm rise substantially over the course of hyperparameter tuning. This increase in validation set performance further translates to increased test set performance. 

% \todo{Moreover, we see that test set performance range for each algorithm over the 5 trial has little overlap for different time point in the hyperparameter tuning process (e.g., MixMatch on TissueMNIST~\ref{fig:tmedfigures_timepoint} after 4h VS after 50h)}

Given the trends we observed across 4 datasets, we think that for a chosen algorithm on a new dataset, following our hyperparameter tuning protocol (even with limited labeling budget and computation budget), we can likely expect better generalization (measured by test set performance) compared to not tuning hyperparameters at all.

\subsection{Differentiating Between Methods}

\citet{oliver2018realistic} offer both empirical and theoretical analysis of how well one can distinguish if one method is truly better than another on a limited labeled dataset. Below, we revisit each analysis for our specific experiments.

\subsubsection{Empirical Analysis of Differentiation}

\citet{oliver2018realistic} in their Fig 5 and 6 show that on SVHN, between 10 random samples of the validation set across several level of validation set size (1000, 500, 250, 100), the validation accuracy of the trained Pi-model, VAT, Pseudo-labeling and Mean Teacher model has substantial variability and overlap with each other. Thus, they caution that differentiating between models might be infeasible with realistic validation set size. 

In our present study, we employ a relaxed notion of ``realistic validation set'', by letting the validation set to be at most as large as the training set. Our experiments cover validation set size 235 (TMED), 400 (Tissue), 450 (Path), 600 (AIROGS); test set size 2019 (TMED2), 47280 (Tissue), 7180 (Path), 6000 (AIROGS). Our experiment shows that within the wide range of methods considered, differentiating between some models are possible. For example, in Fig.~\ref{fig:tmedfigures_timepoint} we can see that MixMatch is clearly better than Mean Teacher in TissueMNIST and PathMNIST, in both the validation set and test set, without overlap in the intervals. The field of semi-supervised learning has made significant advancements in recent years. It is crucial to reevaluate previous conclusions in light of the new developments.

\subsubsection{Theoretical Analysis of Differentiation}
\label{app:Theoretical Analysis}

Here, we show that the performance gain we observe on the test set are real. We perform the same theoretical analysis using the Hoeffding's inequality \citep{hoeffding1994probability} as in \citet{oliver2018realistic}. 
\begin{equation}
\mathbf{P}(|\bar{V} - \mathbb{E}[V]| < p) > 1-2\exp(-2np^2)
\end{equation}
where $\bar{V}$ is the empirical estimate of some model performance metric, $\mathbb{E}[V]$ is its hypothetical true value, $p$ is the desired maximum deviation between our estimate and the true value, and $n$ is the number of examples used.

On TissueMNIST, we have 47280 test samples, we will be more than 99.98\% confident that the test accuracy is within 1\% of its true value. On Path, we have 7180 test samples, we will be more than 99\% confident that the test accuracy is within 2\% its true value.

% On TMED2, we have 2019 test samples, we will be around 95\% confident that the test accuracy is within 3\% its true value.

% In Fig \ref{fig:test_performance_vs_time}, we see that after hyperparameter tuning, the final test accuracy of each algorithms improves much more than 1\% on TissueMNIST, 2\% on PathMNIST, and than 3\% on TMED-2, showing the efficacy of hyperparameter tuning.

% Similarly, we can see that the difference between top-performing algorithms (e.g., MixMatch) and worst-performaning alogrithm (e.g., Mean Teacher) is clearly larger then 1\% on TissueMNIST, 2\% on PathMNIST, and than 3\% on TMED-2. 
% Thus we can argue that differentiation between certain methods are viable.

In Fig \ref{fig:test_performance_vs_time}, we see that after hyperparameter tuning, the final test accuracy of each algorithms improves much more than 1\% on TissueMNIST and 2\% on PathMNIST showing the efficacy of hyperparameter tuning.

Similarly, we can see that the difference between top-performing algorithms (e.g., MixMatch) and worst-performaning alogrithm (e.g., Mean Teacher) is clearly larger then 1\% on TissueMNIST, 2\% on PathMNIST. Thus we can argue that differentiation between certain methods are viable. The same analysis can also be applied to TMED-2 and AIROGS.

% \subsection{Transfer Learning}
% \label{app:PracticalTransferLearning}

% Pretraining is a widely adopted approach, however, the success of pretraining is not always guaranteed. For example, \citet{he2019rethinking} show that pretraining does not necessarily yield performance improvements, even between tasks that are seemingly 'similar'. \citet{raghu2019transfusion} find that the effectiveness of transferring from natural image to medical image might be limited, possibly due to various reasons like domain mismatch. On the other hand, transferring from a similar domain are more likely to succeed. For example, within the medical domain \citet{alzubaidi2020towards,liang2020transfer,heker2020joint}. Similarly, table 2 of \citet{su2021realistic} looks at pretraining on their Semi-Aves dataset of fine-grained bird species classification. Pretraining a labeled-set only baseline on the iNaturalist dataset (which contains overlapping bird categories of images with the target Semi-Aves task) reaches 65\% top-1 accuracy, outperforming pretraining on all of ImageNet (52.7\%).

\subsection{Answers to Common Questions from Reviewers}

Here we answer a few questions that were common to several reviewers of our paper. 

\subsubsection{For a medical image application, would a larger labeled dataset be more important than than developing semi-supervised or self-supervised methods?}
Yes, in general, is is preferable to collect as large of a labeled dataset as possible, at least up to the point of performance saturation.
Investing in data collection likely has a larger payoff than investing in SSL. However, extensive collection of labeled examples is \textbf{not practical} for many real-world clinical tasks due to reasons like financial cost, logistics, privacy and legal issues (see \citet{oliver2018realistic},\citet{berthelot2019mixmatch}, \citet{shekoofeh2021big}).

For this reason, methods for overcoming limited labeled data, such as semi-SL and self-SL, are \textbf{important topics in medical imaging applications}. The clinical use case of SSL motivates several recent methodological works, such ~\citet{zhang2022boostmis}, \citet{aziziRobustDataefficientGeneralization2023}, and \citet{shekoofeh2021big}.

\subsubsection{Isn't it already well-known that hyperparameter tuning  with a realistic-sized validation set is viable?}
When labeled data is abundant, as in common \emph{supervised} learning settings, hyperparameter tuning is widely known as effective. 
However, our work focuses on the \emph{semi/self-SL} setting, where labels are limited.
We carefully reviewed semi/self-SL literature and argue that the viability of tuning on \emph{realistic-sized} validation sets is \textbf{not well-known} in this setting. 
As our paper's Table~\ref{tab:related_work_comparison} shows, existing SSL benchmarks often use validation sets larger than the training set! Seminal work by \citet{oliver2018realistic}
cautions that ``Extensive hyperparameter tuning may be somewhat futile due to an excessively small collection of held-out data ...”. 
\citet{su2021realistic} use a similar claim to justify not doing \emph{any tuning} on their Semi-Fungi dataset experiments.

\subsubsection{Does MixMatch outperform Flex/Fix/CoMatch because RandAug not suitable for medical imaging?}
In general, RandAug-type augmentation can be successful for medical imaging tasks \citep{zhang2022boostmis,chen2020realistic}, though we agree that it might not be ``optimal''.
Instead, we hypothesize that MixMatch's primary advantage is lower runtime cost per iteration compared to FixMatch and successors.
In our AIROGS ResNet-18 experiments (Fig.~\ref{fig:test_performance_vs_time}), MixMatch explores at least 80\% more hyperparameter combinations than its counterparts (111 vs. 59 for FixMatch).

\clearpage

\section{Hyperparameter Details} 

% We further publish the chosen hyperparameter setting from the search in \url{https://github.com/tufts-ml/SSL-vs-SSL-benchmark}
\subsection{Hyperparameter Tuning Strategy: Random Search Details}

Below, in a specific table for each of the 16 methods (supervised, semi-, or self-), we provide a method-specific table showing the random sampling distribution used for each hyperparameter for the random search of Alg.~\ref{alg:hyperparam_tuning}.

\label{App_Hyper_Details}.

\newcommand{\MyTabWidth}{.48\textwidth}

\begin{tabular}{l r}
\multicolumn{2}{c}{Settings Common to All Methods} \\
\hline
Optimizer & Adam \\ 
\hline
Learning rate schedule & Cosine
\end{tabular}

\subsubsection{Supervised Baselines}

\begin{multicols}{2}

\resizebox{\MyTabWidth}{!}{
\begin{tabular}{l r}
\multicolumn{2}{c}{Labeled only} \\
\hline
Batch size & 64 \\ 
\hline
Learning rate & $3 \times 10^x, X \sim \text{Unif}(-5, -2)$ \\ 
\hline
Weight decay & $4 \times 10^x, X \sim \text{Unif}(-6, -3)$
\end{tabular}
}

\resizebox{\MyTabWidth}{!}{
\begin{tabular}{l r}
\multicolumn{2}{c}{MixUp} \\
\hline
Batch size & 64 \\ 
\hline
Learning rate & $3 \times 10^x, X \sim \text{Unif}(-5, -2)$ \\ 
\hline
Weight decay & $4 \times 10^x, X \sim \text{Unif}(-6, -3)$\\
\hline
Beta shape $\alpha$ & $x, X \sim \text{Unif}(0.1, 10)$ \footnotemark[1]
\end{tabular}
}

\resizebox{\MyTabWidth}{!}{
\begin{tabular}{l r}
\multicolumn{2}{c}{Sup Contrast} \\
\hline
Batch size & 256 \\ 
\hline
Learning rate & $3 \times 10^x, X \sim \text{Unif}(-5.5, -1.5)$ \\ 
\hline
Weight decay & $4 \times 10^x, X \sim \text{Unif}(-7.5, -3.5)$\\
\hline
Temperature & $x, X \sim \text{Unif}(0.05, 0.15)$
\end{tabular}
}
\end{multicols}

\subsubsection{Semi-Supervised Methods}

\begin{multicols}{2} %%% Semi-SL methods
\resizebox{\MyTabWidth}{!}{
\begin{tabular}{l r}
\multicolumn{2}{c}{FlexMatch} \\
\hline
Labeled batch size & 64 \\ 
\hline
Unlabeled batch size & 448 \footnotemark[2]\\ 
\hline
Learning rate & $3 \times 10^x, X \sim \text{Unif}(-5, -2)$ \\ 
\hline
Weight decay & $4 \times 10^x, X \sim \text{Unif}(-6, -3)$\\
\hline
Unlabeled loss coefficient  & $10^x, X \sim \text{Unif}(-1, 1)$\\
\hline
Unlabeled loss warmup schedule & No warmup\\
\hline
Pseudo-label threshold & 0.95\\
\hline
Sharpening temperature & 1.0
\end{tabular}
}

\resizebox{\MyTabWidth}{!}{
\begin{tabular}{l r}
\multicolumn{2}{c}{FixMatch} \\
\hline
Labeled batch size & 64 \\ 
\hline
Unlabeled batch size & 448 \footnotemark[2]\\ 
\hline
Learning rate & $3 \times 10^x, X \sim \text{Unif}(-5, -2)$ \\ 
\hline
Weight decay & $4 \times 10^x, X \sim \text{Unif}(-6, -3)$\\
\hline
Unlabeled loss coefficient  & $10^x, X \sim \text{Unif}(-1, 1)$\\
\hline
Unlabeled loss warmup schedule & No warmup\\
\hline
Pseudo-label threshold & 0.95\\
\hline
Sharpening temperature & 1.0
\end{tabular}
\footnotetext[2]{For TMED2, unlabeled batch size is set to 320 to reduce GPU memory usage}
}

\resizebox{\MyTabWidth}{!}{
\begin{tabular}{l r}
\multicolumn{2}{c}{CoMatch} \\
\hline
Labeled batch size & 64 \\ 
\hline
Unlabeled batch size & 448 \footnotemark[2]\\ 
\hline
Learning rate & $3 \times 10^x, X \sim \text{Unif}(-5, -2)$ \\ 
\hline
Weight decay & $4 \times 10^x, X \sim \text{Unif}(-6, -3)$\\
\hline
Unlabeled loss coefficient  & $10^x, X \sim \text{Unif}(-1, 1)$\\
\hline
Unlabeled loss warmup schedule & No warmup\\
\hline
Contrastive loss coefficient  & $5 \times 10^x, X \sim \text{Unif}(-1, 1)$\\
\hline
Pseudo-label threshold & 0.95\\
\hline
Sharpening temperature & 0.2
\end{tabular}
}

\resizebox{\MyTabWidth}{!}{
\begin{tabular}{l r}
\multicolumn{2}{c}{MixMatch} \\
\hline
Labeled batch size & 64 \\ 
\hline
Unlabeled batch size & 64 \\ 
\hline
Learning rate & $3 \times 10^x, X \sim \text{Unif}(-5, -2)$ \\ 
\hline
Weight decay & $4 \times 10^x, X \sim \text{Unif}(-6, -3)$\\
\hline
Beta shape $\alpha$ & $x, X \sim \text{Unif}(0.1, 1)$ \footnotemark[1]\\
\hline
Unlabeled loss coefficient  & $7.5\times 10^x, X \sim \text{Unif}(0, 2)$\\
\hline
Unlabeled loss warmup schedule & linear\\
\hline
Sharpening temperature & 0.5
\end{tabular}
\footnotetext[1]{In practice, we round each sampled $\alpha$ value to the nearest tenth decimal place}
}

\resizebox{\MyTabWidth}{!}{
\begin{tabular}{l r}
\multicolumn{2}{c}{Mean Teacher} \\
\hline
Labeled batch size & 64 \\ 
\hline
Unlabeled batch size & 64 \\ 
\hline
Learning rate & $3 \times 10^x, X \sim \text{Unif}(-5, -2)$ \\ 
\hline
Weight decay & $4 \times 10^x, X \sim \text{Unif}(-6, -3)$\\
\hline
Unlabeled loss coefficient  & $8\times 10^x, X \sim \text{Unif}(-1, 1)$\\
\hline
Unlabeled loss warmup schedule & linear\\
\end{tabular}
}

\resizebox{\MyTabWidth}{!}{
\begin{tabular}{l r}
\multicolumn{2}{c}{Pseudo-label} \\
\hline
Labeled batch size & 64 \\ 
\hline
Unlabeled batch size & 64 \\ 
\hline
Learning rate & $3 \times 10^x, X \sim \text{Unif}(-5, -2)$ \\ 
\hline
Weight decay & $4 \times 10^x, X \sim \text{Unif}(-6, -3)$\\
\hline
Unlabeled loss coefficient  & $10^x, X \sim \text{Unif}(-1, 1)$\\
\hline
Unlabeled loss warmup schedule & Linear\\
\hline
Pseudo-label threshold & 0.95
\end{tabular}
}
\end{multicols}

%%%%%%%%%%%%%%%%%%%%%%%%%%%%%%%%%%%%%%%%%%%%%%%%%%%%%%%%%%%%%%%%%%

\subsubsection{Self-supervised Methods}
\begin{multicols}{2} %%% Semi-SL methods

\resizebox{\MyTabWidth}{!}{
\begin{tabular}{l r}
\multicolumn{2}{c}{SwAV} \\
\hline
Batch size & 256 \\ 
\hline
Learning rate & $1 \times 10^x, X \sim \text{Unif}(-4.5, -1.5)$ \\ 
\hline
Weight decay & $1 \times 10^x, X \sim \text{Unif}(-6.5, -3.5)$\\
\hline
Temperature & $x, X \sim \text{Unif}(0.07, 0.12)$\\
\hline
Num. prototypes &  $1 \times 10^x, X \sim \text{Unif}(1, 3)$ \\
\end{tabular}
}

\resizebox{\MyTabWidth}{!}{
\begin{tabular}{l r}
\multicolumn{2}{c}{MoCo} \\
\hline
Batch size & 256 \\
\hline
Learning rate & $1 \times 10^x, X \sim \text{Unif}(-4.5, -1.5)$ \\ 
\hline
Weight decay & $1 \times 10^x, X \sim \text{Unif}(-6.5, -3.5)$\\
\hline
Temperature & $x, X \sim \text{Unif}(0.07, 0.12)$\\
\hline
Momentum & $x, X \sim \text{Unif}(0.99, 0.9999)$
\end{tabular}
}

\resizebox{\MyTabWidth}{!}{
\begin{tabular}{l r}
\multicolumn{2}{c}{SimCLR} \\
\hline
Batch size & 256 \\
\hline
Learning rate & $1 \times 10^x, X \sim \text{Unif}(-4.5, -1.5)$ \\ 
\hline
Weight decay & $1 \times 10^x, X \sim \text{Unif}(-6.5, -3.5)$\\
\hline
Temperature & $x, X \sim \text{Unif}(0.07, 0.12)$\\
\end{tabular}
}

\resizebox{\MyTabWidth}{!}{
\begin{tabular}{l r}
\multicolumn{2}{c}{SimSiam} \\
\hline
Batch size & 256 \\
\hline
Learning rate & $1 \times 10^x, X \sim \text{Unif}(-4.5, -1.5)$ \\ 
\hline
Weight decay & $1 \times 10^x, X \sim \text{Unif}(-6.5, -3.5)$\\
\end{tabular}
}

\resizebox{\MyTabWidth}{!}{
\begin{tabular}{l r}
\multicolumn{2}{c}{BYOL} \\
\hline
Batch size & 256 \\
\hline
Learning rate & $1 \times 10^x, X \sim \text{Unif}(-4.5, -1.5)$ \\ 
\hline
Weight decay & $1 \times 10^x, X \sim \text{Unif}(-6.5, -3.5)$\\
\hline
Temperature & $x, X \sim \text{Unif}(0.07, 0.12)$\\
\hline
Momentum & $x, X \sim \text{Unif}(0.99, 0.9999)$
\end{tabular}
}

\resizebox{\MyTabWidth}{!}{
\begin{tabular}{l r}
\multicolumn{2}{c}{DINO} \\
\hline
Batch size & 256 \\
\hline
Learning rate & $1 \times 10^x, X \sim \text{Unif}(-4.5, -1.5)$ \\ 
\hline
Weight decay & $1 \times 10^x, X \sim \text{Unif}(-6.5, -3.5)$\\
\hline
Temperature & $x, X \sim \text{Unif}(0.07, 0.12)$\\
\hline
Momentum & $x, X \sim \text{Unif}(0.99, 0.9999)$
\end{tabular}
}

\resizebox{\MyTabWidth}{!}{
\begin{tabular}{l r}
\multicolumn{2}{c}{Barlow Twins} \\
\hline
Batch size & 256 \\
\hline
Learning rate & $1 \times 10^x, X \sim \text{Unif}(-4.5, -1.5)$ \\ 
\hline
Weight decay & $1 \times 10^x, X \sim \text{Unif}(-6.5, -3.5)$\\
\hline
Temperature & $x, X \sim \text{Unif}(0.07, 0.12)$\\
\hline
Momentum & $x, X \sim \text{Unif}(0.99, 0.9999)$
\end{tabular}
}

\end{multicols}

\subsection{Hyperparameter transfer strategy}
\label{App:Hyper_Transfer_strategy}
To make the most of limited labeled data, one potential strategy recommended by \citet{su2021realistic} is to use the entire labeled set for training, reserving no validation set at all. This relies on pre-established hyperparameters from other dataset/experiments. In this study, we experiment with two scenarios: using pre-determined hyperparameters tuned for  CIFAR-10, or using hyperparameters tuned for TissueMNIST. 

The CIFAR-10 hyperparameters are sourced from repositories published by each method's original authors, as this is a common benchmark in the SSL literature. We ensure that each hyperparameter choice, when applied using the re-implented code for each method in our codebase, matches previously reported results on CIFAR-10. 

The TissueMNIST hyperparameters originate from our experiments as depicted in Figure~\ref{fig:validation-acc-over-time} ($a$). For exact values, see App.~\ref{app:hyperparameters_tuned_to_Tissue}.

For each method using the transfer strategy, we perform training on the combined train+validation set, setting the maximum number of epochs to 100 for PathMNIST and AIROGS (80 epochs for TMED2).
Training is terminated early if the train loss does not improve over 20 consecutive epochs. Empirically, we observe that all models which did not trigger early stopping reached a plateau in training loss.

\subsubsection{Best Hyperparameters on TissueMNIST for Semi-Supervised Methods}
\label{app:hyperparameters_tuned_to_Tissue}

Below we report the chosen hyperparameters on TissueMNIST for each semi-supervised method, as used in the hyperparameter transfer experiments.

\begin{multicols}{2}

\resizebox{\MyTabWidth}{!}{
\begin{tabular}{l r r r r r}
\multicolumn{5}{c}{FlexMatch} \\
\hline
  & seed0 & seed1 & seed2 & seed3 & seed4 \\ 
\hline
% Labeled batch size & 64 & 64 & 64 & 64 & 64 \\ 
% % \hline
% Unlabeled batch size & 448 & 448 & 448 & 448 & 448 \\ 
% % \hline
Learning rate & 0.00036 & 0.00016 & 0.00016 & 0.00068 & 0.00006 \\ 
% \hline
Weight decay & 0.00259 & 0.00001 & 0.00371 & 0.00023 & 0.002103\\
% \hline
Unlabeled loss coefficient  & 2.22 & 0.82 & 5.00 & 1.94 & 6.09\\
\end{tabular}
}

\resizebox{\MyTabWidth}{!}{
\begin{tabular}{l r r r r r}
\multicolumn{5}{c}{FixMatch} \\
\hline
  & seed0 & seed1 & seed2 & seed3 & seed4 \\ 
\hline
% Labeled batch size & 64 & 64 & 64 & 64 & 64 \\ 
% % \hline
% Unlabeled batch size & 448 & 448 & 448 & 448 & 448 \\ 
% % \hline
Learning rate & 0.00074 & 0.00034 & 0.00392 & 0.00102 & 0.00037 \\ 
% \hline
Weight decay & 0.00045 & 0.00315 & 0.00001 & 0.00005 & 0.00058\\
% \hline
Unlabeled loss coefficient  & 3.08 & 6.70 & 1.85 & 1.46 & 0.47\\
\end{tabular}
}

\resizebox{\MyTabWidth}{!}{
\begin{tabular}{l r r r r r}
\multicolumn{5}{c}{CoMatch} \\
\hline
  & seed0 & seed1 & seed2 & seed3 & seed4 \\ 
\hline
% Labeled batch size & 64 & 64 & 64 & 64 & 64 \\ 
% % \hline
% Unlabeled batch size & 448 & 448 & 448 & 448 & 448 \\ 
% % \hline
Learning rate & 0.00124 & 0.00145 & 0.00061 & 0.00026 & 0.00113 \\ 
% \hline
Weight decay & 0.00042 & 0.00009 & 0.00005 & 0.00009 & 0.00017\\
% \hline
Unlabeled loss coefficient  & 0.30 & 1.71 & 1.26 & 2.74 & 0.46\\
Contrastive loss coefficient  & 1.26 & 2.21 & 3.71 & 0.56 & 1.37\\
\end{tabular}
}

\resizebox{\MyTabWidth}{!}{
\begin{tabular}{l r r r r r}
\multicolumn{5}{c}{MixMatch} \\
\hline
  & seed0 & seed1 & seed2 & seed3 & seed4 \\ 
\hline
% Labeled batch size & 64 & 64 & 64 & 64 & 64 \\ 
% % \hline
% Unlabeled batch size & 448 & 448 & 448 & 448 & 448 \\ 
% % \hline
Learning rate & 0.00028 & 0.00003 & 0.00018 & 0.00009 & 0.00005 \\ 
% \hline
Weight decay & 0.000005 & 0.00195 & 0.00005 & 0.00085 & 0.00082\\
Beta shape $\alpha$ & 0.2 & 0.9 & 0.9 & 0.8 & 0.7\\
% \hline
Unlabeled loss coefficient  & 9.13 & 37.96 & 8.06 & 25.16 & 11.17\\
\end{tabular}
}

\resizebox{\MyTabWidth}{!}{
\begin{tabular}{l r r r r r}
\multicolumn{5}{c}{Mean Teacher} \\
\hline
  & seed0 & seed1 & seed2 & seed3 & seed4 \\ 
\hline
% Labeled batch size & 64 & 64 & 64 & 64 & 64 \\ 
% % \hline
% Unlabeled batch size & 448 & 448 & 448 & 448 & 448 \\ 
% % \hline
Learning rate & 0.00062 & 0.00022 & 0.00005 & 0.00128 & 0.00125 \\ 
% \hline
Weight decay & 0.00189 & 0.00001 & 0.00008 & 0.00001 & 0.00001\\
% \hline
Unlabeled loss coefficient  & 67.67 & 0.87 & 1.25 & 7.60 & 13.56\\
\end{tabular}
}

\resizebox{\MyTabWidth}{!}{
\begin{tabular}{l r r r r r}
\multicolumn{5}{c}{Pseudo-label} \\
\hline
  & seed0 & seed1 & seed2 & seed3 & seed4 \\ 
\hline
% Labeled batch size & 64 & 64 & 64 & 64 & 64 \\ 
% % \hline
% Unlabeled batch size & 448 & 448 & 448 & 448 & 448 \\ 
% % \hline
Learning rate & 0.00007 & 0.00021 & 0.00005 & 0.00063 & 0.00060 \\ 
% \hline
Weight decay & 0.00033 & 0.00093 & 0.00383 & 0.00005 & 0.00087\\
% \hline
Unlabeled loss coefficient  & 0.19 & 0.16 & 8.73 & 0.82 & 0.25\\
\end{tabular}
}
\end{multicols}

\subsubsection{Best Hyperparameters on TissueMNIST for Self-Supervised Methods}

Below we report the chosen hyperparameters on TissueMNIST for each self-supervised method, as used in the hyperparameter transfer experiments.

\begin{multicols}{2}

\resizebox{\MyTabWidth}{!}{
\begin{tabular}{l r r r r r}
\multicolumn{5}{c}{SwAV} \\
\hline
  & seed0 & seed1 & seed2 & seed3 & seed4 \\ 
\hline
% Labeled batch size & 64 & 64 & 64 & 64 & 64 \\ 
% % \hline
% Unlabeled batch size & 448 & 448 & 448 & 448 & 448 \\ 
% % \hline
Learning rate & 0.00065 & 0.00325 & 0.00012 & 0.00086 & 0.00196 \\ 
% \hline
Weight decay & 0.0001497 & 0.0000056 & 0.0000006 & 0.0000021 & 0.0000003\\
% \hline
Num. prototypes & 845 & 131 & 36 & 201 & 59\\
\end{tabular}
}

\resizebox{\MyTabWidth}{!}{
\begin{tabular}{l r r r r r}
\multicolumn{5}{c}{MoCo} \\
\hline
  & seed0 & seed1 & seed2 & seed3 & seed4 \\ 
\hline
% Labeled batch size & 64 & 64 & 64 & 64 & 64 \\ 
% % \hline
% Unlabeled batch size & 448 & 448 & 448 & 448 & 448 \\ 
% % \hline
Learning rate & 0.00288 & 0.00023 & 0.00043 & 0.00005 & 0.02629 \\ 
% \hline
Weight decay & 0.000002 & 0.0000008 & 0.0000003 & 0.0000005 & 0.0000004\\
% \hline
temperature  & 0.09331 & 0.07097 & 0.10987 & 0.07414 & 0.07080\\
Momentum & 0.99242 & 0.99672 & 0.99267 & 0.99950 & 0.99538\\
\end{tabular}
}

\resizebox{\MyTabWidth}{!}{
\begin{tabular}{l r r r r r}
\multicolumn{5}{c}{SimCLR} \\
\hline
  & seed0 & seed1 & seed2 & seed3 & seed4 \\ 
\hline
% Labeled batch size & 64 & 64 & 64 & 64 & 64 \\ 
% % \hline
% Unlabeled batch size & 448 & 448 & 448 & 448 & 448 \\ 
% % \hline
Learning rate & 0.00217 & 0.00131 & 0.000640 & 0.00380 & 0.00136 \\ 
% \hline
Weight decay & 0.00002 & 0.00001 & 0.00001 & 0.00001 & 0.00001\\
% \hline
temperature  & 0.11719 & 0.10426 & 0.08652 & 0.07784 & 0.11478\\
\end{tabular}
}

\resizebox{\MyTabWidth}{!}{
\begin{tabular}{l r r r r r}
\multicolumn{5}{c}{SimSiam} \\
\hline
  & seed0 & seed1 & seed2 & seed3 & seed4 \\ 
\hline
% Labeled batch size & 64 & 64 & 64 & 64 & 64 \\ 
% % \hline
% Unlabeled batch size & 448 & 448 & 448 & 448 & 448 \\ 
% % \hline
Learning rate & 0.0002 & 0.00056 & 0.00013 & 0.00338 & 0.00098 \\ 
% \hline
Weight decay & 0.000066 & 0.000046 & 0.000023 & 0.000001 & 0.000001\\
\end{tabular}
}

\resizebox{\MyTabWidth}{!}{
\begin{tabular}{l r r r r r}
\multicolumn{5}{c}{BYOL} \\
\hline
  & seed0 & seed1 & seed2 & seed3 & seed4 \\ 
\hline
% Labeled batch size & 64 & 64 & 64 & 64 & 64 \\ 
% % \hline
% Unlabeled batch size & 448 & 448 & 448 & 448 & 448 \\ 
% % \hline
Learning rate & 0.000245 & 0.001308 & 0.000371 & 0.001653 & 0.001959 \\ 
% \hline
Weight decay & 0.0000007 & 0.0000057 & 0.0000004 & 0.000003 & 0.000001\\
% \hline
Momentum & 0.9928618 & 0.996167 & 0.9988484 & 0.9940063 & 0.9934791\\
\end{tabular}
}

\resizebox{\MyTabWidth}{!}{
\begin{tabular}{l r r r r r}
\multicolumn{5}{c}{DINO} \\
\hline
  & seed0 & seed1 & seed2 & seed3 & seed4 \\ 
\hline
% Labeled batch size & 64 & 64 & 64 & 64 & 64 \\ 
% % \hline
% Unlabeled batch size & 448 & 448 & 448 & 448 & 448 \\ 
% % \hline
Learning rate & 0.000245 & 0.001308 & 0.000371 & 0.001653 & 0.001959 \\ 
% \hline
Weight decay & 0.0000007 & 0.0000057 & 0.0000004 & 0.000003 & 0.000001\\
% \hline
Momentum & 0.9928618 & 0.996167 & 0.9988484 & 0.9940063 & 0.9934791\\
\end{tabular}
}

\resizebox{\MyTabWidth}{!}{
\begin{tabular}{l r r r r r}
\multicolumn{5}{c}{Barlow Twins} \\
\hline
  & seed0 & seed1 & seed2 & seed3 & seed4 \\ 
\hline
% Labeled batch size & 64 & 64 & 64 & 64 & 64 \\ 
% % \hline
% Unlabeled batch size & 448 & 448 & 448 & 448 & 448 \\ 
% % \hline
Learning rate & 0.000245 & 0.001308 & 0.000371 & 0.001653 & 0.001959 \\ 
% \hline
Weight decay & 0.0000007 & 0.0000057 & 0.0000004 & 0.000003 & 0.000001\\
% \hline
Momentum & 0.9928618 & 0.996167 & 0.9988484 & 0.9940063 & 0.9934791\\
\end{tabular}
}
\end{multicols}

%\section{Related Work: Additional discussion}
%\input{App_related_work}
%\label{App_related_work}

%%%%%%%%%%%%%%%%%%%%%%%%%%%%%%%%%%%%%%%%%%%%%%%%%%%%%%%%%%%%

\end{document}